\documentclass[11pt]{article}
\usepackage[a4paper, margin=2cm]{geometry}
\usepackage[utf8]{inputenc}
\usepackage[T1]{fontenc}
\usepackage{authblk,natbib,amssymb,graphicx}
\usepackage{amsmath,amsthm,tikz,stmaryrd,float}
\usepackage{tgtermes}
\usepackage[tworuled,vlined]{algorithm2e}
\usepackage{listings}
\usepackage{bookmark}
\bookmarksetup{depth=2}

\usepackage{url}

\definecolor{amber}{rgb}{1.0, 0.75, 0.0}
\definecolor{fire}{rgb}{0.99215686, 0.64705882, 0.05882353}
\definecolor{amaranth}{rgb}{0.9, 0.17, 0.31}
\definecolor{azure}{rgb}{0.0, 0.5, 1.0}
\hypersetup{colorlinks,linkcolor={fire},citecolor={azure},urlcolor={amaranth}}

\definecolor{codegreen}{rgb}{0,0.6,0}
\definecolor{codegray}{rgb}{0.5,0.5,0.5}
\definecolor{codepurple}{rgb}{0.58,0,0.82}
\definecolor{backcolour}{rgb}{0.94,0.94,0.96}
\lstdefinestyle{mystyle}{
    backgroundcolor=\color{backcolour},   
    stringstyle=\color{codepurple},
    commentstyle=\color{codegreen},
    numberstyle=\tiny\color{codegray},
    keywordstyle=\bfseries\color{red!40!black},
    basicstyle=\ttfamily\footnotesize,
    numbers=left,      
}
\usetikzlibrary{bayesnet}

\DeclareMathOperator*{\argmin}{arg\,min}
\usepackage{xspace}

\def \({\left(}
\def \){\right)}
\def \[{\left[}
\def \]{\right]}

\DeclareMathOperator*{\extr}{extr}
\newcommand{\mF}{\mathcal{F}}

\newcommand{\mM}{\mathcal{M}}
\newcommand{\mN}{\mathcal{N}}
\newcommand{\KL}{\textrm{KL}}
\newcommand{\EE}{\mathbb{E}}
\newcommand{\bbR}{\mathbb{R}}

\DeclareMathOperator{\Tr}{Tr}
\newcommand{\iid}[0]{\rm iid}

\newcommand{\tramp}{\textsf{Tree-AMP}\xspace}

\newcommand{\pymc}{\textsf{PyMC3}\xspace}
\newcommand{\scikit}{\textsf{Scikit-Learn}\xspace}

\newtheorem{example}{Example} 
\newtheorem{theorem}{Theorem}
\newtheorem{proposition}[theorem]{Proposition} 
\newtheorem{remark}[theorem]{Remark}
\long\def\acks#1{\vskip 0.3in\noindent{\large\bf Acknowledgments}\vskip 0.2in
\noindent #1}

\begin{document}

\title{Tree-AMP: Compositional Inference with Tree Approximate Message Passing}

\author[1]{Antoine Baker}
\author[2]{Benjamin Aubin}
\author[1]{Florent Krzakala}
\author[2]{Lenka Zdeborov\'a}

\affil[1]{Laboratoire de Physique, CNRS, École Normale Supérieure, PSL University, Paris, France}
\affil[2]{Institut de Physique Théorique, CNRS, CEA, Université Paris-Saclay, Saclay, France}

\maketitle

\begin{abstract}%   <- trailing '%' for backward compatibility of .sty file
We introduce \tramp, standing for \emph{Tree Approximate Message Passing}, a python package for compositional inference in
high-dimensional tree-structured models. The package provides a unifying framework to study several 
approximate message passing algorithms previously derived for a variety of machine learning tasks such as generalized linear models, inference in multi-layer networks, matrix factorization, and reconstruction using non-separable penalties. For some models, the asymptotic performance of the algorithm can be theoretically predicted by the state evolution, and the measurements entropy estimated by the free entropy formalism. The implementation is modular by design: each module, which implements a factor, can be composed at will with other modules to solve complex inference tasks. The user only needs to declare the factor graph of the model: the inference algorithm, state evolution and entropy estimation are fully automated. The source code is publicly available at \href{https://github.com/sphinxteam/tramp}{https://github.com/sphinxteam/tramp} and the documentation at \href{https://sphinxteam.github.io/tramp.docs/}{https://sphinxteam.github.io/tramp.docs}. 
\end{abstract}

\newpage
\tableofcontents

\newpage
\section{Introduction}
\label{sec:intro}

%\paragraph{Probabilistic programming}
Probabilistic models have been used in many applications, as diverse as scientific data analysis, coding, natural language and signal processing. They also offer a powerful framework \citep{Bishop2013} to several challenges in machine learning: dealing with uncertainty, choosing hyper-parameters, causal reasoning and model selection. However, the difficulty of deriving and implementing approximate inference algorithms for each new model may have hindered the wider adoption of Bayesian methods. The probabilistic programming approach seeks to make Bayesian inference as user friendly and streamlined as  possible: ideally the user would only need to declare the probabilistic model and run an inference engine. Several probabilistic programming frameworks have been proposed, well suited for different contexts and leveraging  variational inference or sampling methods to automate inference. To give a few examples, \textsf{pomegranate} \citep{Schreiber2018} fits probabilistic models using maximum likelihood. \textsf{Church} \citep{Goodman2008} and successors are universal languages for representing generative models. \textsf{Infer.NET} \citep{InferNET18} implements several message passing algorithms such as Expectation Propagation. \textsf{Stan} \citep{Carpenter2017} uses Hamiltonian Monte Carlo, while \textsf{Anglican} \citep{Wood2014} uses particle MCMC as the sampling method. \textsf{Turing} \citep{Ge2018} offers a Julia implementation. Recently,  \textsf{Edward} \citep{Tran2016} and \textsf{Pyro}  \citep{Bingham2018pyro} tackle deep probabilistic problems, scaling inference up to large data and complex models.

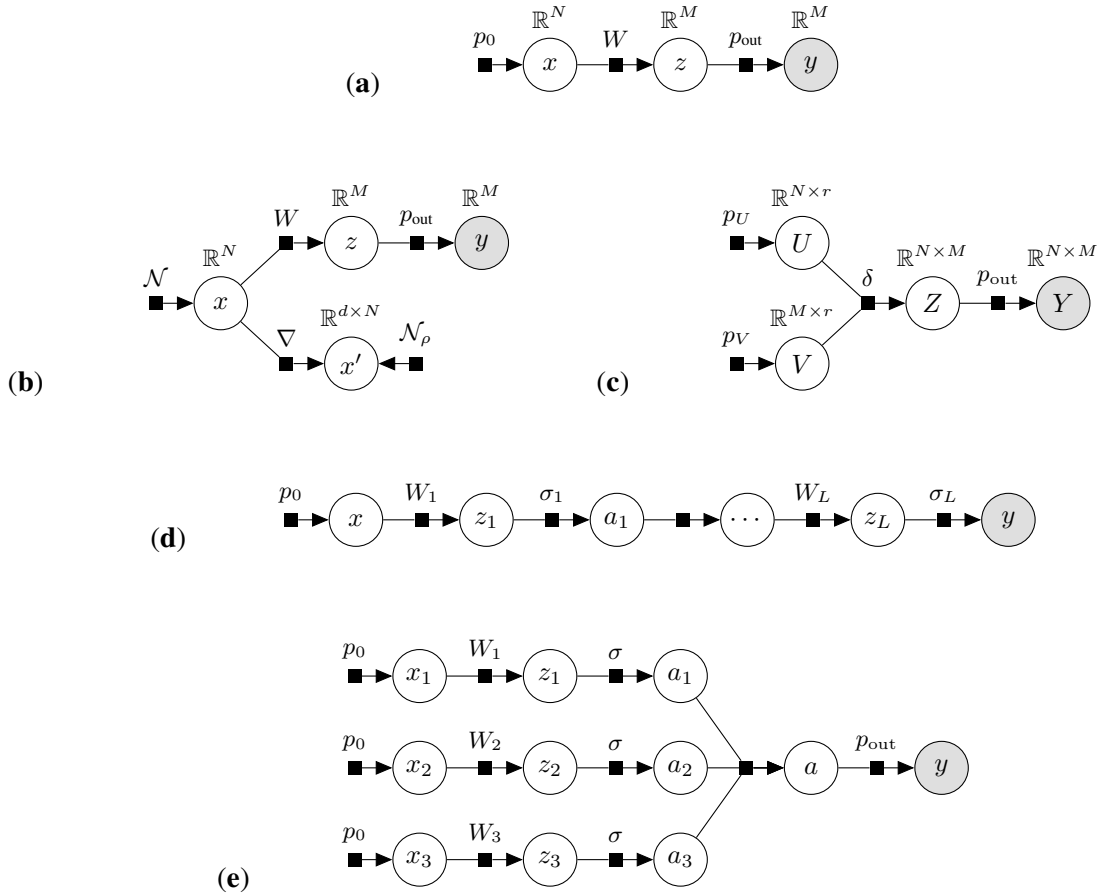
\begin{figure}[t]
  \centering
  (\textbf{a})\hspace{1cm}
  \begin{tikzpicture}
    \node[latent, label=$\mathbb{R}^N$]                             (x)  {$x$} ;
    \node[latent, label=$\mathbb{R}^M$, right=of x]                 (z)  {$z$} ;
    \node[obs, label=$\mathbb{R}^M$, right=of z]  (y)  {$y$} ;

    \factor[left=of x] {p} {$p_0$} {} {x} ;
    \factor[left=of z] {p} {$W$} {x} {z} ;
    \factor[left=of y] {out} {$p_\text{out}$} {z} {y} ;
  \end{tikzpicture}
  
  \vspace{1cm}
 \begin{minipage}[c]{0.45\linewidth}
    (\textbf{b}) \hspace{1cm}
    \begin{tikzpicture}
    \node[latent, label=$\mathbb{R}^N$] (x) {$x$} ;
    \node[latent, label=$\mathbb{R}^M$, right=of x, yshift= 0.8cm] (z) {$z$} ;
    \node[obs, label=$\mathbb{R}^M$, right=of z]  (y)  {$y$} ;
    \node[latent, label=$\mathbb{R}^{d \times N}$, right=of x, yshift=-0.8cm] (dx) {$x'$} ;

    \factor[left=of x] {p} {$\mathcal{N}$} {} {x} ;
    \factor[left=of z] {p} {$W$} {x} {z} ;
    \factor[left=of y] {out} {$p_\text{out}$} {z} {y} ;
    \factor[left=of dx] {grad} {$\nabla$} {x} {dx} ;
    \factor[right=of dx] {pdx} {$\mathcal{N}_\rho$} {} {dx} ;
    \end{tikzpicture}
    \end{minipage}
    \begin{minipage}[c]{0.45\linewidth}
    (\textbf{c})\hspace{1cm}
    \begin{tikzpicture}
    \node[latent, label=$\mathbb{R}^{N\times M}$]           (z)  {$Z$} ;
    \node[latent, label=$\mathbb{R}^{N\times r}$, left=of z, yshift=0.8cm] (u)  {$U$} ;
    \node[latent, label=$\mathbb{R}^{M\times r}$, left=of z, yshift=-0.8cm] (v)  {$V$} ;
    \node[obs, right=of z, label=$\mathbb{R}^{N\times M}$]  (y)  {$Y$} ;
    
    \factor[left=of u] {p} {$p_U$} {} {u} ;
    \factor[left=of v] {p} {$p_V$} {} {v} ;
    \factor[left=of z] {p} {$\delta$} {u, v} {z} ;
    \factor[left=of y] {p} {$p_{\rm out}$} {z} {y} ;
    
  \end{tikzpicture}
\end{minipage}

  \vspace{1cm}
  (\textbf{d})\hspace{1cm}
    \begin{tikzpicture}
    \node[latent]                            (x)  {$x$} ;
    \node[latent, right=of x]               (z1)  {$z_1$} ;
    \node[latent, right=of z1]                 (a1)  {$a_1$} ;
    \node[latent, right=of a1]                 (d)  {$\cdots$} ;
    \node[latent, right=of d]                 (L)  {$z_L$} ;
    \node[obs, right=of L]  (y)  {$y$} ;

    \factor[left=of x] {p} {$p_0$} {} {x} ;
    \factor[left=of z1] {p} {$W_1$} {x} {z1} ;
    \factor[left=of a1] {p} {$\sigma_1$} {z1} {a1} ;
    \factor[left=of d] {p} {} {a1} {d} ;
    \factor[left=of L] {p} {$W_L$} {d} {L} ;
    \factor[left=of y] {out} {$\sigma_L$} {L} {y} ;
  \end{tikzpicture}
  
  \vspace{1cm}
 (\textbf{e})\hspace{1cm}
    \begin{tikzpicture}
    \node[latent]                            (x2)  {$x_2$} ;
    \node[latent, right=of x2]               (z2)  {$z_2$} ;
    \node[latent, right=of z2]                 (a2)  {$a_2$} ;
    \node[latent, right=of a2]                 (a)  {$a$} ;
    \node[obs, right=of a]  (y)  {$y$} ;
    \node[latent, above=of a2, yshift= -0.5cm]                 (a1)  {$a_1$} ;
    \node[latent, left=of a1]               (z1)  {$z_1$} ;
    \node[latent, left=of z1]                   (x1)  {$x_1$} ;
    \node[latent, below=of a2, yshift= 0.5cm]                 (a3)  {$a_3$} ;
    \node[latent, left=of a3]               (z3)  {$z_3$} ;
    \node[latent, left=of z3]                   (x3)  {$x_3$} ;

    \factor[left=of x2] {p} {$p_0$}  {} {x2} ;
    \factor[left=of z2] {p} {$W_2$} {x2} {z2} ;
    \factor[left=of a2] {p} {$\sigma$} {z2} {a2} ;
    \factor[left=of a] {p} {} {a2} {a} ;
    \factor[left=of x1] {p} {$p_0$} {} {x1}  ;
    \factor[left=of z1] {p} {$W_1$} {x1} {z1} ;
    \factor[left=of a1] {p} {$\sigma$} {z1} {a1} ;
    \factor[left=of a] {p} {} {a1} {a} ;
    \factor[left=of x3] {p} {$p_0$}  {} {x3} ;
    \factor[left=of z3] {p} {$W_3$} {x3} {z3} ;
    \factor[left=of a3] {p} {$\sigma$} {z3} {a3} ;
    \factor[left=of a] {p} {} {a3} {a} ;
    \factor[left=of y] {p} {$p_{\rm out}$} {a} {y} ;
  \end{tikzpicture}
   \caption{
    Tree-structured models.
    (\textbf{a}) Generalized linear model with separable prior $p_0(x)$, separable likelihood $p_\text{out}(y |z)$ and linear channel $z = Wx$.
    (\textbf{b}) Reconstruction using a sparse gradient prior where sparsity is enforced by the Gauss-Bernoulli prior $\mathcal{N}_\rho = [1-\rho] \delta + \rho \mathcal{N}$.
    (\textbf{c}) Low-rank matrix factorization with separable priors $p_U$ and $p_V$, separable likelihood $p_\text{out}(Y |Z)$ and factorization $Z = \frac{UV^\intercal}{\sqrt{N}}$.
    (\textbf{d}) Multi-layer network with activation functions $a_l = \sigma_l(z_l)$ and linear channels $z_l = W_l a_{l-1}$.
    (\textbf{e}) Committee machine with three experts.
  }
  \label{fig:tree_models}
\end{figure}

In this paper, we present \tramp (Tree Approximate Message Passing).
%\paragraph{AMP algorithms}
In the current rich software ecosystem, \tramp aims to fill a particular niche: using message passing algorithms with theoretical guarantees of performance in specific asymptotic settings. 
% The EP algorithm comes in many different forms, due to the various choices possible for approximation scheme, the factorization of the target distribution and the iterative scheme \citep{Vehtari2014}.  As will be detailed in Section~\ref{sec:EP}, \tramp uses a specific flavor of EP as its inference engine, which is closely related to AMP algorithms.
As will be detailed in Section~\ref{sec:EP}, \tramp uses the Expectation Propagation (EP) algorithm \citep{Minka_EP_2001} as its inference engine, which is also implemented by \textsf{Infer.NET}.
Application-wise, for inference in statistical models or Bayesian machine learning tasks, the scope of \tramp is very limited compared to the \textsf{Infer.NET} package. 
Indeed \tramp is restricted to models like the ones presented in Figure~\ref{fig:tree_models}, that is tree-structured factor graphs connecting high-dimensional variables, while \textsf{Infer.NET} can be applied to generic factor graphs with variables of arbitrary dimensions and types.
However for the models considered in Figure~\ref{fig:tree_models}, under specific asymptotic settings, \tramp offers an in-depth theoretical analysis of its performance as will be detailed in Section~\ref{sec:SE}: its errors can be predicted using the state evolution formalism, the free entropy formalism further predicts when the algorithm achieves or not the Bayes-optimal performance and also allows to estimate information theoretic quantities. For this reason, we believe the \tramp package should be of special interest to theoretical researchers seeking a better understanding of EP/AMP algorithms. As an alternative to \tramp let us mention the \href{https://github.com/GAMPTeam/vampyre}{\sf Vampyre} package that allows inference in multi-layer networks \citep{Fletcher2018}.

There is a long history behind message passing \citep{Yedidia2001,mezard2009information}, approximate message passing (AMP) \citep{Donoho2009} and vector approximate message passing (VAMP) \citep{Schniter2016}, that we shall discuss later on. As  exemplified in the context of compressed sensing, the AMP algorithm has a fundamental property: its performance on random instances in the high-dimensional limit, measured by the mean squared error on the signals, can be rigorously predicted by the so-called state evolution \citep{Donoho2009,Bayati2011}, a rigorous version of the physicists "cavity method" \citep{mezard1987spin}. These performances can be shown, in some cases, to reach the Bayes optimal one in polynomial time \citep{barbier2016mutual,Reeves2016}, quite a remarkable feat! More recently, variant of the AMP approach has been developed with (some) correlated data and matrices \citep{Schniter2016,ma2017orthogonal}, again with guarantees of optimally in some cases \citep{barbier2018mutual,gerbelot2020asymptotic}. These approaches are intimately linked with the Expectation Propagation algorithm \citep{Minka_EP_2001} and the Expectation Consistency framework \citep{Opper2005JMLR}. 

The state evolution and the Bayes optimal guarantees were  extended to a wide variety of models including for instance generalized linear models (GLM) \citep{rangan2011generalized,Barbier2019}, matrix factorization \citep{rangan2012iterative,deshpande2014information,dia2016mutual,Lesieur2017}, committee machines \citep{Aubin2018}, optimization with non separable penalties (such as total variation) \citep{som2012compressive,metzler2015optimal,tan2015compressive,Manoel2018}, inference in multi-layer networks \citep{Manoel2017, Fletcher2018,Gabrie2018} and even arbitrary trees of GLMs \citep{Reeves2017}. In all these cases, the entropy of the system in the high dimensional limit can be obtained as the minimum of the so-called free entropy potential \citep{Yedidia2001,Yedidia2005,krzakala2014variational} and this allows the computation of interesting information theoretic quantities such as the mutual information between layers in a neural network \citep{Gabrie2018}. Furthermore, the global minimizer of the free entropy potential corresponds to the minimal mean squared error, which allows to determine fundamental limits to inference. Interestingly, the mean squared error achieved by AMP, predicted by the state evolution, is a stationary point of the same free entropy potential, which allows for an interesting interpretation of when the algorithm actually works \citep{Zdeborova2016} in terms of phase transitions.

%\paragraph{Tree-structured models}
Unfortunately the development of AMP algorithms faced the same caveat as probabilistic modeling: for each new model, the AMP algorithm and the associated theory (free entropy and state evolution) had to be derived and implemented separately, which can be time-consuming. However, a key observation is that the factor graphs \citep{Kschischang2001} for all the models mentioned above are tree-structured as illustrated in Figure~\ref{fig:tree_models}. Each factor corresponds to an elementary inference problem that can be solved analytically or approximately. The \tramp python package offers a unifying framework for all the models discussed above and extends to arbitrary tree-structured models. Similar to other probabilistic programming frameworks, the user only has to declare the model (here a tree-structured factor graph) then the inference, state evolution and entropy estimation are fully automated.  The implementation is also completely modularized and extending \tramp is in principle straightforward. If a new factor is needed, the user only has to solve (analytically or approximately) the elementary inference problem corresponding to this factor and implement it as a module in \tramp.

%\paragraph{Outline}

Many of the AMP algorithms previously mentioned, especially the vectorized versions considered in \citep{Schniter2016, Manoel2018, Fletcher2018},  can be stated as particular instances of the Expectation Propagation (EP) algorithm \citep{Minka_EP_2001}. 
The EP algorithm is equivalent to the Expectation Consistency framework \citep{Opper2005JMLR} which further yields an approximation for the log-evidence. Actually both approaches are solutions of the same relaxed Bethe variational problem \citep{Heskes2005}. In Section~\ref{sec:EP} we present the weak consistency derivation of EP by \citep{Heskes2005} which offers a unifying framework that extends the previously mentioned AMP algorithms to tree-structured models. These are classic results but we hope that this pedagogical review will clarify the link between the various free energy formulations of the EP/AMP algorithms. Besides it allows us to introduce the key quantities (posterior moments and log-evidence) at the heart of the \tramp implementation.
Next in Section~\ref{sec:SE} we heuristically derive the replica free entropy \citep{mezard2009information} by using weak consistency on the overlaps and conjecture the state evolution. Even if the derivation is non-rigorous, we recover earlier results derived for specific models. Interestingly, the state evolution and the free entropy potential can be reinterpreted as simple ensemble average of the posterior variances and log-evidence estimated by EP. This allows us to extend the state evolution and free entropy formalism to tree-structured models and implement them in the \tramp package. 
Finally, in Section~\ref{sec:Examples} we illustrate the package on a few examples. 
\section{Expectation Propagation \label{sec:EP}}

In this section we review the derivation of EP as a relaxed
variational problem.  First we briefly recall
the variational inference framework and the Bethe decomposition of the free
energy \citep{Yedidia2001} which is exact for tree-structured models. Then
following \cite{Heskes2005}, the Bethe variational problem can be approximately
solved by enforcing moment-matching instead of full consistency of the
marginals, which yields the EP algorithm. The EP solution consists of
exponential family distributions which satisfy a duality between natural
parameters and moments \citep{Wainwright2008}. The EP free energy \citep{Minka_free_2001} is shown to be equivalent to the AMP free energies and satisfies a tree decomposition which is at the heart of the modularization in the \tramp package. Finally we expose the \tramp implementation of EP and MAP estimation.

\subsection{Model settings \label{sec:EP_settings}}

The \tramp package is limited to 
high-dimensional tree-structured factor graphs, like the 
models presented in Figure~\ref{fig:tree_models}.
To define more precisely such models and set some notation, consider an inference problem $p(\mathbf{x}, \mathbf{y})$ where $\mathbf{x} = (x_i)_{i \in V}$ are the signals to infer and $\mathbf{y} =( y_j)_{j \in O}$ the measurements. We emphasize that in our context each signal $x_i \in \mathbb{R}^{N_i}$ and measurement $y_j \in \mathbb{R}^{N_j}$ is itself a high dimensional object.
The model is typically considered in the large $N$ limit with ratios $\alpha_i = \frac{N_i}{N} = O(1)$ and $\alpha_j = \frac{N_j}{N} = O(1)$. We will assume that $p(\mathbf{x}, \mathbf{y})$ can be factorized as a tree-structured probabilistic graphical model:
 \begin{equation}
  \label{factor_graph_unobserved}
  p(\mathbf{x},\mathbf{y}) = \frac{1}{Z_N} \prod_{k \in F} f_k(x_k; y_k), \quad
  Z_N = \int d\mathbf{x}  d\mathbf{y}  \prod_{k \in F} f_k(x_k; y_k),
\end{equation}
with factors $(f_k)_{k \in F}$. The factorization structure can be conveniently represented as a factor graph \citep{Kschischang2001}: a bipartite graph $\mathcal{G} = (V, F, E)$, where each signal $x_i$ is represented as a variable node $i \in V$ (circle), each factor $f_k$ by a factor node $k \in F$ (square), with an edge $(i,k) \in E$ connecting the variable node $i$ to the factor node $k$ if and only if $x_i$ is an argument of $f_k$. We will use the symbol $\partial$ to denote the neighbors nodes in the factor graph. Thus $i \in \partial k$ denotes the variable nodes neighboring the factor node $k$ and the arguments of the factor $f_k$ are $x_k = ( x_i )_{i \in \partial k}$. Similarly $k \in \partial i$ denotes the factor nodes neighboring the variable node $i$.

Note that while we denote $\mathbb{R}^{N_i}$ the integration domain of the signal $x_i$ our inference tasks are not limited to real high-dimensional variables. Indeed a high-dimensional binary, sparse, categorical or complex variable can always be embedded in some $\mathbb{R}^{N_i}$ and its type enforced by an appropriate factor. For instance a binary variable $x \in \pm^N$ can be enforced by a binary prior $p_0(x) = p_+ \delta_+(x) + p_- \delta_-(x) $. As a consequence, we allow generic measures for the factors, including Dirac measures. This additionally allows us to represent hard constraints as factors, for example the linear channel $z = W x$ will be represented by the factor $\delta(z-Wx)$. 
The high-dimensional factor $f_k$ has to be simple enough to lead to a tractable inference problem as will be explained in Section~\ref{sec:EP_modules}. Some representative models are given in Examples~\ref{example:glm}-\ref{example:sparse_grad}.

The goal of the inference is then to get the posterior $p(\mathbf{x} \mid \mathbf{y})$ and evidence $p(\mathbf{y})$, or equivalently the negative log-evidence known as surprisal in information theory. For the factorization Eq.~\eqref{factor_graph_unobserved} the posterior is equal to:
\begin{equation}
  \label{factor_graph_conditional}
  p(\mathbf{x} \mid \mathbf{y}) = \frac{1}{Z_N(\mathbf{y})} \prod_{k \in F} f_k(x_k; y_k), \quad
  Z_N(\mathbf{y}) = \int d\mathbf{x} \prod_{k \in F} f_k(x_k; y_k).
\end{equation}
The \emph{Helmholtz free energy} is here defined as the negative log partition
\begin{equation}
    \label{helmholtz}
    F(\mathbf{y}) = -  \ln Z_N(\mathbf{y}) = - \ln p(\mathbf{y}) -  \ln Z_N
\end{equation}
and gives the surprisal up to a constant.

\begin{remark}[Bayesian network] 
\label{remark:bayes_net}
In a Bayesian network, the factors correspond to the conditional distributions $f_k(x_k ; y_k) = p(x_k^+, y_k \mid x_k^-) $ where $x_k^+$ (resp. $x_k^-$) denote the outputs (resp. inputs) variables of the factor, besides $Z_N = 1$ so the Helmholtz free energy Eq.~\eqref{helmholtz} directly gives the surprisal. All the models considered in Figure~\ref{fig:tree_models} are Bayesian networks, with the exception of \textbf{(b)}. 
\end{remark}

\begin{example}[GLM] 
\label{example:glm}
The factor graph for the generalized linear model (GLM) is shown in Figure~\ref{fig:tree_models}~\textbf{(a)}. A high-dimensional signal $x \in \mathbb{R}^N$ is drawn from a separable prior $p_0$, a high-dimensional measurement $y \in \mathbb{R}^M$ is obtained through a separable likelihood $p_\text{out}$ from $z = Wx \in \mathbb{R}^M$, where $W$ is a given $M\times N$ matrix. The model is typically considered in the large $N$ limit with $\alpha = \frac{M}{N} = O(1)$.
The variables to infer are $\mathbf{x} = (x_i)_{i \in V} = (x, z)$, the observation $\mathbf{y} = (y_j)_{j \in O} = y $ and the factors $(f_k)_{k \in F} = \left( p_0(x), \delta(z - Wx), p_\text{out}(y|z) \right)$. 
\end{example}

\begin{example}[Low rank matrix factorization] 
\label{example:low_rank}
The factor graph for the low rank matrix factorization model considered in \citep{Lesieur2017} is shown in Figure~\ref{fig:tree_models}~\textbf{(c)}.
Two matrices $U \in \mathbb{R}^{N \times r}$ and $V \in \mathbb{R}^{M \times r}$ are drawn from separable priors $p_U$ and $p_V$, a high-dimensional measurement $Y \in \mathbb{R}^{N \times M}$ is obtained through a separable likelihood $p_\text{out}$ from $Z = \frac{UV^\intercal}{\sqrt{N}} \in \mathbb{R}^{N \times M}$. The model is typically considered in the large $N$ limit with $\alpha = \frac{M}{N} = O(1)$ and finite rank $r= O(1)$.
The variables to infer are $\mathbf{x} = (x_i)_{i \in V} = (U, V, Z)$, the observation $\mathbf{y} = (y_j)_{j \in O} = Y$ and the factors $(f_k)_{k \in F} = (p_U(U), p_V(V), \delta(Z - \frac{UV^\intercal}{\sqrt{N}}), p_\text{out}(Y|Z))$. 
\end{example}

\begin{example}[Extensive rank matrix factorization] 
\label{example:extensive_rank}
The factor graph for the extensive rank matrix factorization model considered in \citep{Kabashima2016} is shown below.
\begin{figure}[!h]
\centering
    \begin{tikzpicture}
    \node[latent, label=$\mathbb{R}^{M\times P}$]           (z)  {$Z$} ;
    \node[latent, label=$\mathbb{R}^{M\times N}$, left=of z, yshift=0.8cm]           (u)  {$F$} ;
    \node[latent, label=$\mathbb{R}^{N\times P}$, left=of z, yshift=-0.8cm]           (v)  {$X$} ;
    \node[obs, right=of z, label=$\mathbb{R}^{M\times N}$]  (y)  {$Y$} ;
    
    \factor[left=of u] {p} {$p_F$} {} {u} ;
    \factor[left=of v] {p} {$p_X$} {} {v} ;
    \factor[left=of z] {p} {$\delta$} {u, v} {z} ;
    \factor[left=of y] {p} {$p_{\rm out}$} {z} {y} ;
    
  \end{tikzpicture}
\end{figure}
This is the same factor graph as the low rank case but considered in a different asymptotic regime.
Two matrices $F \in \mathbb{R}^{M \times N}$ and $X \in \mathbb{R}^{N \times P}$ are drawn from separable priors $p_F$ and $p_X$, a high-dimensional measurement $Y \in \mathbb{R}^{M \times P}$ is obtained through a separable likelihood $p_\text{out}$ from $Z = \frac{FX}{\sqrt{N}} \in \mathbb{R}^{M \times P}$. The model is typically considered in the large $N,M,P$ limit with fixed ratios $\alpha = \frac{M}{N} = O(1)$ and $\pi = \frac{P}{N} = O(1)$.
The variables to infer are $\mathbf{x} = (x_i)_{i \in V} = (F, X, Z)$, the observation $\mathbf{y} = (y_j)_{j \in O} = Y$ and the factors $(f_k)_{k \in F} = \left( p_F(F), p_X(X), \delta(Z - \tfrac{FX}{\sqrt{N}}), p_\text{out}(Y|Z) \right)$. 
\end{example}

\begin{example}[Sparse gradient regression]
\label{example:sparse_grad}
The factor graph for this model is shown is shown in Figure~\ref{fig:tree_models}~\textbf{(b)}. Compared to the GLM, we wish to infer a signal $x \in \mathbb{R}^N$ which gradient $x' = \nabla x \in \mathbb{R}^{d \times N}$ is sparse (we view the signal as having $d$ axis, for instance $d=2$ for an image, and consequently the gradient is taken along $d$ directions). The sparsity is enforced by a Gauss-Bernoulli prior $\mathcal{N}_\rho = [1-\rho] \delta + \rho \mathcal{N}$. 
The variables to infer are $\mathbf{x} = (x_i)_{i \in V} = (x, x', z)$, the observation $\mathbf{y} = (y_j)_{j \in O} = y  $ and the factors $(f_k)_{k \in F} = \left( p_0(x), \mathcal{N}_\rho(x'), \delta(z - Wx), p_\text{out}(y|z) \right)$. The factor graph is not a Bayesian network, in particular the partition function Eq.~\eqref{factor_graph_unobserved} is equal to:
\begin{equation}
Z_N = \int_{\mathbb{R}^N} dx \mathcal{N}(x)\mathcal{N}_\rho(\nabla x) \neq 1.
\end{equation}
\end{example}

\subsection{Bethe free energy}

We briefly recall the Bethe variational formulation of the Belief Propagation algorithm following \citep{Yedidia2001} and refer the reader to this reference or \citep{Wainwright2008} for further details. 
We are interested in computationally hard inference problems and seek an
approximation $\tilde{p}$ of the posterior distribution $\tilde{p}(\mathbf{x}) \simeq p(\mathbf{x} \mid \mathbf{y})$. Consider such an approximation $\tilde{p}$ and the following functional
\begin{equation}
  \label{variational_free_energy}
  \mF[\tilde{p}] = \KL [ \tilde{p}(\mathbf{x})  \Vert p(\mathbf{x} \mid \mathbf{y}) ] + F(\mathbf{y})\,,
\end{equation}
called the \emph{variational free energy}. As the KL divergence is always
positive and equal to zero only if the two distributions are equal, we can formally get
the posterior and the Helmholtz free energy $F(\mathbf{y})$ as the solution of the variational
problem:
\begin{equation}
  \label{variational_pb}
  F(\mathbf{y}) = \min_{\tilde{p}} \mF[\tilde{p}]  \quad \text{for} \quad \tilde{p}^\star(\mathbf{x}) = p(\mathbf{x} \mid \mathbf{y}).
\end{equation}
However, for a tree-structured model it can be shown that the posterior
factorizes as
\begin{equation}
  p(\mathbf{x} \mid \mathbf{y}) =
  \frac{\prod_{k \in F} p(x_k \mid \mathbf{y})}{\prod_{i \in V} p(x_i \mid \mathbf{y})^{n_i - 1}}\,,
\end{equation}
where $p(x_i \mid \mathbf{y})$ is the marginal of the variable $x_i$, $p(x_k \mid \mathbf{y})$ is
the joint marginal over $x_k = (x_i)_{i \in \partial k}$ and
$n_i = |\partial i|$ is the number of neighbor factors of the variable $x_i$.
Therefore we can restrict the variational problem Eq.~\eqref{variational_pb}
to distributions of the form
\begin{equation}
  \label{tree_factorization}
  \tilde{p}(\mathbf{x}) = \frac{\prod_{k \in F} \tilde{p}_k(x_k)}{\prod_{i \in V} \tilde{p}_i(x_i)^{n_i - 1}}\,,
\end{equation}
and minimize over the collection of variable marginals $\tilde{p}_V = (\tilde{p}_i)_{i \in V}$ and factor marginals $\tilde{p}_F = (\tilde{p}_k)_{k \in F}$. This collection however has to satisfy a strong self-consistency constraint: whenever the variable $x_i$ is an argument of the factor $f_k$, the $i$-marginal of the factor marginal $\tilde{p}_k$ must give back the variable marginal $\tilde{p}_i$. In other words the collection of marginals $(\tilde{p}_V, \tilde{p}_F)$ must belong to the set:
\begin{equation}
  \label{strong_consistency}
  \mM = \left\{
      (\tilde{p}_V,\tilde{p}_F) \quad : \quad
      \forall (i, k) \in E, \quad
      \tilde{p}_i(x_i) = \tilde{p}_k(x_i) = \int dx_{k\setminus i}\,\tilde{p}_k(x_k)
  \right\}.
\end{equation}
For distributions of the type Eq.~\eqref{tree_factorization}, the variational free
energy Eq.~\eqref{variational_free_energy}
is equal to
\begin{align}
   \notag
  &\mF_\text{Bethe}[\tilde{p}_V,\tilde{p}_F] =
  \sum_{k \in F} F_k[\tilde{p}_k] + \sum_{i \in V} (1 - n_i) F_i[\tilde{p}_i] \, , \\
  \label{bethe_free_energy}
  &\text{with} \quad
  F_k[\tilde{p}_k] = \KL [ \tilde{p}_k \Vert f_k]              \, , \quad
  F_i[\tilde{p}_i] = - \textrm{H} [\tilde{p}_i]              \, ,
\end{align}
called the \emph{Bethe free energy} \citep{Yedidia2001}, where $\KL$ and $\textrm{H}$ denote respectively the Kullback-Leibler divergence and the entropy.
Therefore for a tree-structured model we have:
\begin{equation}
  \label{bethe_variational_pb}
  F(\mathbf{y}) = \min_{(\tilde{p}_V,\tilde{p}_F) \in \mathcal{M}} \mF_\text{Bethe}[\tilde{p}_V,\tilde{p}_F]
  \quad \text{at} \quad
  \begin{cases}
    \tilde{p}_k^\star(x_k) = p(x_k \mid \mathbf{y}) 
    &\text{for all } k \in F \\
    \tilde{p}_i^\star(x_i) = p(x_i \mid \mathbf{y}) 
    &\text{for all } i \in V
  \end{cases}
\end{equation}
The solution of this Bethe variational problem actually leads to the
Belief Propagation algorithm \citep{Pearl1988, Yedidia2001}.

\subsection{Weak consistency \label{sec:weak_consistency}}

For tree-structured models the Bethe variational problem
Eq.~\eqref{bethe_variational_pb} yields the exact posterior and
Helmoltz free energy, and is solved by the Belief Propagation algorithm. Unfortunately for the models introduced in Figure~\ref{fig:tree_models}, which involve high-dimensional vectors or matrices,
the Belief Propagation algorithm is not tractable.
Following \cite{Heskes2005}
we consider instead a relaxed version of the Bethe variational problem, by replacing the strong consistency constraint Eq.~\eqref{strong_consistency} by
the \emph{weak consistency} constraint:
\begin{equation}
  \label{weak_consistency}
  \mathcal{M}_\phi = \left\{
      (\tilde{p}_V,\tilde{p}_F) \quad : \quad
      \forall (i,k) \in E, \quad
      \mathbb{E}_{\tilde{p}_i} \phi_i(x_i) =
      \mathbb{E}_{\tilde{p}_k} \phi_i(x_i)
  \right\}\,,
\end{equation}
for  a collection $\phi = (\phi_i)_{i \in V}$ of sufficient statistics $\phi_i : \mathbb{R}^{N_i} \to \mathbb{R}^{d_i}$ for each variable $x_i$. In other words instead of requiring the full consistency of the marginals we only require moment matching. The collection $\phi$ is a choice and each choice leads to a different approximation scheme. As the notation suggests, one can choose different sufficient statistics for each variable $x_i$. Following \cite{Wainwright2008}  we use $\langle \lambda_i , \phi_i(x_i) \rangle$ to denote the Euclidian inner product in $\mathbb{R}^{d_i}$ of the so-called natural parameter $\lambda_i \in \mathbb{R}^{d_i}$ with $\phi_i(x_i) \in \mathbb{R}^{d_i}$. Currently the \tramp package only supports isotropic Gaussian beliefs (Example~\ref{example:isotropic_beliefs}) but we plan to include more generic Gaussian beliefs (Examples~\ref{example:diagonal_beliefs}-\ref{example:structured_beliefs}) 
in future versions of the package. The \emph{relaxed Bethe variational problem}:
\begin{equation}
  \label{relaxed_bethe_variational_pb}
  F_\phi(\mathbf{y}) = \min_{ (\tilde{p}_V,\tilde{p}_F)  \in \mathcal{M}_\phi} \mF_\text{Bethe}[\tilde{p}_V,\tilde{p}_F] \simeq F(\mathbf{y})
  \quad \text{at} \quad
  \begin{cases}
    \tilde{p}_k^\star(x_k) \simeq p(x_k \mid \mathbf{y}) 
    &\text{for all } k \in F \\
    \tilde{p}_i^\star(x_i) \simeq p(x_i \mid \mathbf{y}) 
    &\text{for all } i \in V
  \end{cases}
\end{equation}
leads to the following solution as proven by \citep{Heskes2005}. Let $\lambda_{i \to k} \in \mathbb{R}^{d_i}$ denotes the Lagrange multiplier associated to the moment matching constraint  
$\mathbb{E}_{\tilde{p}_i} \phi_i(x_i) = \mathbb{E}_{\tilde{p}_k} \phi_i(x_i)$. 
The (approximate) factor marginal $\tilde{p}_k^\star(x_k)$ belongs to the exponential family
\begin{equation}
    \label{factor_marginal}
    p_k(x_k \mid \lambda_k)  = \frac{1}{Z_k[\lambda_k]} f_k(x_k; y_k) 
    e^{\langle \lambda_k , \phi_k(x_k) \rangle}, \quad
    Z_k[\lambda_k] = \int dx_k  f_k(x_k; y_k) e^{\langle \lambda_k , \phi_k(x_k) \rangle}
\end{equation}
with natural parameter and sufficient statistics
\begin{equation}
    \label{factor_natural_param}
    \lambda_k = (\lambda_{i \to k} )_{i \in \partial k}, \quad
    \phi_k(x_k) = \left( \phi_i(x_i) \right)_{i \in \partial k}, \quad
    \langle \lambda_k , \phi_k(x_k) \rangle = \sum_{i \in \partial k} \langle \lambda_{i \to k}, \phi_i(x_i) \rangle.
\end{equation}
The (approximate) variable marginal $\tilde{p}_i^\star(x_i)$ belongs to the exponential family
\begin{equation}
    \label{variable_marginal}
    p_i(x_i \mid \lambda_i) = \frac{1}{Z_i[\lambda_i]} 
    e^{\langle \lambda_i , \phi_i(x_i) \rangle }, \quad
    Z_i[\lambda_i] = \int dx_i  e^{\langle \lambda_i , \phi_i(x_i) \rangle}
\end{equation}
with natural parameter given by
\begin{equation}
\label{lambda_variable}
(n_i - 1) \lambda_i = \sum_{k \in \partial i} \lambda_{i \to k},
\end{equation}
or introducing the factor-to-variable messages $\lambda_{k \to i} = \lambda_i - \lambda_{i \to k}$:
\begin{equation}
    \lambda_i = \sum_{k \in \partial i} \lambda_{k \to i}.
\end{equation}
The moment matching condition can be written as:
\begin{equation}
    \label{moment_matching}
    \mu_i^\star = \mu_i[\lambda_i] = \mu_i^k[\lambda_k]
\end{equation}
where $\mu_i[\lambda_i] = \mathbb{E}_{p_i(x_i \mid \lambda_i)} \phi_i(x_i)$ and
$\mu_i^k[\lambda_k] = \mathbb{E}_{p_k(x_k \mid \lambda_k)} \phi_i(x_i)$ are the
moments of the variable $x_i$ as estimated by the variable and factor marginal
respectively. This is the fixed point searched by the EP algorithm
\citep{Minka_EP_2001}. 

\begin{example}[Isotropic Gaussian belief] 
\label{example:isotropic_beliefs}
It corresponds to the sufficient statistics $\phi_i(x_i) = (x_i, -\frac{1}{2} \Vert x_i \Vert^2)$ so $d_i = N_i + 1$. The associated natural parameters are $\lambda_i = (b_i, a_i) $ with $b_i \in \mathbb{R}^{N_i}$ and scalar precision $a_i \in \mathbb{R}_+$. The inner product leads to an isotropic Gaussian belief $e^{\langle \lambda_i , \phi_i(x_i) \rangle} = e^{-\frac{a_i}{2} \Vert x_i \Vert^2 + b_i^\intercal x_i} $on $x_i$.
\end{example}

\begin{example}[Diagonal Gaussian belief] 
\label{example:diagonal_beliefs}
It corresponds to the sufficient statistics $\phi_i(x_i) = (x_i, -\frac{1}{2} x_i^2)$ so $d_i = 2 N_i$ and the precision $a_i\in \mathbb{R}_+^{N_i}$ is a vector.
\end{example}

\begin{example}[Full covariance Gaussian belief] 
\label{example:full_cov_beliefs}
It corresponds to the sufficient statistics $\phi_i(x_i) = (x_i, -\frac{1}{2}x_i x_i^\intercal)$  so $d_i = N_i + \frac{N_i (N_i+1)}{2}$ and the precision $a_i$  is a $N_i \times N_i$ positive symmetric matrix. Note however that such a belief will be computationally demanding as inverting the precision matrix will take $O(N_i^3)$ time.
\end{example}

\begin{example}[Structured Gaussian beliefs]
\label{example:structured_beliefs}
When the high dimensional variable $x_i$ has an inner structure with multiple indices, a more complex covariance structure can be envisioned.
For instance in the low rank matrix factorization problem (Example~\ref{example:low_rank}) \cite{Lesieur2017} consider for $U \in \mathbb{R}^{N \times r}$ a Gaussian belief with a full covariance in the second coordinate but diagonal in the first. 
In other words the low rank matrix $U = (U_n)_{n=1}^N$ is viewed as a collection of vectors  $U_n \in \mathbb{R}^r$ with sufficient statistics $\phi(U) = (U_n, -\frac{1}{2} U_n U_n^\intercal)_{n=1}^N$, natural parameters $\lambda_U = ( b_n, a_n)_{n=1}^N$ with $b_n \in \mathbb{R}^r$ and $a_n$ is a $r\times r$ positive symmetric matrix.  
Then $d_U = N r + N \frac{r(r+1)}{2}$ and $\lambda_U = (b_U, a_U)$ with $b_U \in \mathbb{R}^{N \times r}$ and $a_U \in \mathbb{R}^{N \times r \times r}$. 
\end{example}

\subsection{Moments and natural parameters duality \label{sec:exp_family_duality}}

As the solution of the relaxed Bethe variational problem involves exponential family distributions, it is useful to recall some of their basic properties \citep{Wainwright2008}. The log-partitions
\begin{align}
  \label{factor_log_partition}
  A_k[\lambda_k] &= \ln Z_k[\lambda_k] = \ln \int dx_k\, f_k(x_k; y_k) \, e^{\langle \lambda_k , \phi_k(x_k) \rangle}\,, \\
  A_i[\lambda_i] &= \ln Z_i[\lambda_i]  = \ln \int dx_i\, e^{\langle \lambda_i,  \phi_i(x_i) \rangle}\,,
\end{align}
are convex functions and provide the bijective mappings between the convex set of natural parameters and the convex set of moments:
\begin{align}
  \label{factor_moment}
  \mu_k[\lambda_k] &=  \mathbb{E}_{p_k(x_k \mid \lambda_k)} \phi_k(x_k) = \partial_{\lambda_k} A_k[\lambda_k]\,, \\
  \label{variable_moment}
  \mu_i[\lambda_i] &= \mathbb{E}_{p_i(x_i \mid \lambda_i)} \phi_i(x_i)\; = \partial_{\lambda_i} A_i[\lambda_i]\,,
\end{align}
and the inverse mappings are given by:
\begin{align}
\lambda_k[\mu_k] &= \partial_{\mu_k} G_k[\mu_k]\,, \\
  \lambda_i[\mu_i] &= \partial_{\mu_i} G_i[\mu_i]\,,
\end{align}
where $G_i[\mu_i]$ and $G_k[\mu_k]$
are the Legendre transformations (convex conjugates) of the log-partitions and
are equal to the KL divergence and the negative entropy respectively:
\begin{align}
  \label{factor_legendre}
  G_k[\mu_k] &= \max_{\lambda_k} \, \langle \lambda_k, \mu_k \rangle - A_k[\lambda_k] 
  = \KL [p_k(x_k \mid \lambda_k) \Vert f_k(x_k ; y_k)]\,, \\
  \label{variable_legendre}
  G_i[\mu_i] &=  \max_{\lambda_i} \, \langle \lambda_i, \mu_i \rangle - A_i[\lambda_i] 
  = - \textrm{H}[p_i(x_i \mid \lambda_i)] \,.
\end{align}
We recall that for the factor marginal the natural parameter, sufficient statistics and inner product are given by Eq.~\eqref{factor_natural_param}. The corresponding moment is  $\mu_k = (\mu_i^k)_{i \in \partial k}$ and the mapping Eq.~\eqref{factor_moment} and the inner product in Eq.~\eqref{factor_legendre} are explicitly:
\begin{align}
        \label{factor_moment_explicit}
      \mu_i^k[\lambda_k] &=  \mathbb{E}_{p_k(x_k \mid \lambda_k)} \phi_i(x_i) 
      = \partial_{\lambda_{i \to k}} A_k[\lambda_k]  \quad \text{for all }  i \in \partial k, \\
    \langle \lambda_k, \mu_k \rangle &= \sum_{i \in \partial k} 
    \langle \lambda_{i \to k}, \mu_i^k \rangle. 
\end{align}

\begin{example}[Duality for isotropic Gaussian beliefs]
\label{example:duality_isotropic}
Let us consider isotropic Gaussian beliefs (Example~\ref{example:isotropic_beliefs}) with natural parameters $\lambda_i = (b_i, a_i) \in \mathbb{R}^{N_i} \times \mathbb{R}_+$. The corresponding moments $\mu_i = (r_i, -\tfrac{N_i}{2} \tau_i)$ are the mean $r_i = \EE_{p_i} x_i \in \mathbb{R}^{N_i}$ and second moment  $\tau_i = \EE_{p_i} \frac{\Vert x_i \Vert^2}{N_i} \in \mathbb{R}_+$. The variable marginal Eq.~\eqref{variable_marginal} is the isotropic Gaussian 
\begin{equation}
   \label{variable_marginal_isotropic}
    p_i(x_i \mid a_i, b_i) = e^{ -\frac{a_i}{2} \Vert x_i \Vert^2 + b_i^\intercal x_i -A_i[b_i, a_i]} = \mathcal{N}(x_i \mid r_i, v_i)
\end{equation}
with mean $r_i \in \mathbb{R}^{N_i}$ and variance $v_i \in \mathbb{R}_+$. The mapping between the two parametrizations is particularly simple:
\begin{equation}
    \label{variable_mean_variance_isotropic}
    r_i = \frac{b_i}{a_i}, \quad v_i = \frac{1}{a_i} \quad \text{and} \quad
    b_i = \frac{r_i}{v_i}, \quad a_i = \frac{1}{v_i}.
\end{equation}
The variable log-partition
\begin{equation}
    \label{variable_log_partition_isotropic}
    A_i[a_i,b_i] = \frac{\Vert b_i \Vert^2 }{2a_i} + \frac{N_i}{2}\ln\frac{2\pi}{a_i}, 
\end{equation}
gives consistently the forward mapping $r_i = \partial_{b_i} A_i = \frac{b_i}{a_i}$
and $-\frac{N_i}{2} \tau_i = \partial_{a_i} A_i$ with $\tau_i = \frac{\Vert r_i \Vert^2}{N_i} + v_i$ and $v_i = \frac{1}{a_i}$.
The variable negative entropy
\begin{equation}
    G_i[r_i,\tau_i] = -\frac{N_i}{2} \ln 2\pi e v_i \quad \text{with} \quad v_i = \tau_i -\frac{\Vert r_i \Vert^2}{N_i}
\end{equation}
gives consistently the inverse mapping
$b_i = \partial_{r_i} G_i = \frac{r_i}{v_i}$ and $-\frac{N_i}{2} a_i = \partial_{\tau_i} G_i$ with $a_i = \frac{1}{v_i}$. The moment matching condition Eq.~\eqref{moment_matching} is equivalent to match the mean $r_i = r_i^k$ and isotropic variance $v_i = v_i^k$. The factor marginal Eq.~\eqref{factor_marginal} is the factor titled by Gaussian beliefs:
\begin{equation}
    \label{factor_marginal_isotropic}
    p_k(x_k \mid a_k, b_k) = f(x_k; y_k) e^{-\frac{1}{2} a_k \Vert x_k \Vert^2 + b_k^\intercal x_k - A_k[a_k, b_k]}
\end{equation}
where following Eq.~\eqref{factor_natural_param} we denote compactly the inner product by:
\begin{equation}
    -\frac{a_k}{2} \Vert x_k \Vert^2 + b_k^\intercal x_k  = \sum_{i \in \partial k} -\frac{a_{i \to k}}{2} \Vert x_i \Vert^2 + b_{i \to k}^\intercal x_i
\end{equation}
The factor log-partition, mean and isotropic variance are explicitly given by:
\begin{align}
\label{factor_log_partition_isotropic}
&A_k[a_k, b_k] = \ln \int dx_k \, f(x_k; y_k) \, e^{-\frac{1}{2} a_k \Vert x_k \Vert^2 + b_k^\intercal x_k} , \\
\label{factor_mean_isotropic}
&r_i^k[a_k, b_k] = \EE_{p_k(x_k \mid a_k, b_k)} x_i = \partial_{b_{i \to k}} A_k[a_k, b_k], \\ 
\label{factor_variance_isotropic}
&v_i^k[a_k, b_k] = \langle \mathrm{Var}_{p_k(x_k \mid a_k, b_k)}(x_i) \rangle 
= \langle \partial^2_{b_{i \to k}} A_k[a_k, b_k] \rangle
\end{align}
where $\langle \cdot \rangle$ denotes the average over components. 
Several factor log-partitions with isotropic Gaussian beliefs are given in Appendix~\ref{app:modules}.
\end{example}

\subsection{Tree decomposition of the free energy}

We now present several but equivalent free energy formulations of the EP and AMP algorithms. The Expectation Consistency (EC) Gibbs free energy \citep{Opper2005NIPS} is a function defined over the posterior moments $\mu_V = (\mu_i)_{i \in V}$:
\begin{equation}
  \label{gibbs_free_energy}
  G[\mu_V] = \sum_{k \in F} G_k[\mu_k] + \sum_{i \in V} (1-n_i) G_i[\mu_i]
  \quad \text{where} \quad
  \mu_k = (\mu_i)_{i\in \partial k} .
\end{equation}
The EP free
energy \citep{Minka_free_2001} is a function defined over the variable $\lambda_V = (\lambda_i)_{i \in V}$ and factor $\lambda_F = (\lambda_k)_{k \in F}$ natural parameters:
\begin{equation}
  \label{minka_free_energy}
  A[\lambda_V, \lambda_F] =
  \sum_{k \in F} A_k[\lambda_k] + \sum_{i \in V} (1-n_i) A_i[\lambda_i].
\end{equation}
The \tramp free energy is the same as the EP free energy
but is parametrized in term of
factor to variable messages and variable to factor
messages $\lambda_E = (\lambda_{k \to i}, \lambda_{i \to k})_{(i,k) \in E}$:
\begin{align}
  \label{tramp_free_energy}
  &A[\lambda_E]
  = \sum_{k \in F} A_k[\lambda_k]
  - \sum_{(i,k) \in E} A_i[\lambda_i^k]
  + \sum_{i \in V} A_i[\lambda_i], \\
  \label{message_param}
  \text{where} \quad
  &\lambda_k = ( \lambda_{i \to k} )_{i \in \partial k}
  \quad
  \lambda_i^k = \lambda_{i \to k} + \lambda_{k \to i},
  \quad
  \lambda_i = \sum_{k \in \partial i} \lambda_{k \to i}.
\end{align}
The parametrization Eq.~\eqref{message_param} in term of messages has a nice
interpretation: the variable natural parameter is the sum of the incoming
messages (coming from the neighboring factors), the factor natural parameter is
the set of the incoming messages (coming from the neighboring variables). 

\begin{proposition}
\label{prop:free_energy}
The relaxed Bethe variational problem Eq.~\eqref{relaxed_bethe_variational_pb} can be formulated in term of the posterior moments using the EC Gibbs free energy, in term of the factor and variable natural parameters using the EP free energy, or in term of the natural parameters messages using the \tramp free energy:
\begin{align}
\label{gibbs_variational}
F_\phi(\mathbf{y}) &= \min_{\mu_V} G[\mu_V] \\
\label{minka_variational}
    &= \min_{\lambda_V} \max_{\lambda_F}
    - A[\lambda_V,\lambda_F]
    \quad\text{s.t.}\quad \forall i \in V: (n_i - 1)
    \lambda_i = \sum_{k \in \partial i} \lambda_{i \to k} \\
\label{tramp_variational}
    &= \min \extr_{\lambda_E} - A[\lambda_E].
\end{align}
Besides, any stationary point of the free energies
(not necessarily the global optima)
is an EP fixed point:
\begin{equation}
  \label{ep_fixed_point}
  \mu_i = \mu_i^k, \quad
  (n_i - 1) \lambda_i = \sum_{k \in \partial i} \lambda_{i \to k}.
\end{equation}
\end{proposition}
The $\min \extr_{\lambda_E}$ notation in Eq.~\eqref{tramp_variational} means to search for stationary (in general saddle) points of $A[\lambda_E]$ and among these critical points select the minimizer of $G$. 

\paragraph{Proofs} There are several but separate derivations of these equivalences in the literature.
In \citep{Minka_free_2001} the optimization problem Eq.~\eqref{minka_variational} is shown to lead to the EP algorithm, where $A[\lambda_V, \lambda_F]$ is called the EP dual energy function while the relaxed Bethe variational problem is called the EP primal energy function.
\cite{Opper2005JMLR, Opper2005NIPS} further show that the minimization Eq.~\eqref{gibbs_variational}, or equivalently the dual saddle point problem Eq.~\eqref{minka_variational}, yields the so-called EC approximation of $-\ln Z_N(\mathbf{y})$, here denoted by $F_\phi(\mathbf{y})$. The EC approximation is presented for a very simple factor graph with only one variable $x$ and two factors $f_q(x)$ and $f_r(x)$ but can be straightforwardly extended to a tree-structured factor graph which yields $A[\lambda_V, \lambda_F]$ and $G[\mu_V]$. Finally, \cite{Heskes2005} unify the two formalisms as solutions to the relaxed Bethe variational problem. Proposition~\ref{prop:free_energy} is the straightforward application of these ideas to the tree-structured models considered in this manuscript. We present in Appendix~\ref{app:free_energy} a condensed proof for the reader convenience using the duality between moments and natural parameters. 

\paragraph{Exact tree decomposition} The Bethe, EC Gibbs, EP and \tramp free energies all follow the same tree decomposition: a sum over factors $-$ sum over edges
$+$ sum over variables. The same tree decomposition holds for the Helmholtz free energy $F(\mathbf{y})$ and the EC approximation $F_\phi(\mathbf{y})$. Indeed $F(\mathbf{y})$ is simply the Bethe free energy evaluated at the true marginals according to Eq.~\eqref{bethe_variational_pb} and $F_\phi(\mathbf{y})$ is the Minka/EC Gibbs/\tramp free energy evaluated at the optimal EP fixed point according to Proposition~\ref{prop:free_energy}.
This tree decomposition is at the heart of the modularization (Section~\ref{sec:EP_modules}) in the \tramp package. We emphasize that this tree decomposition is exact, the approximate part in the inference comes from relaxing the full consistency of the marginals to moment-matching, which effectively projects the marginals onto exponential family approximate marginals Eq.~\eqref{factor_marginal} and Eq.~\eqref{variable_marginal} indexed by a finite-dimensional natural parameter. It is this projection of the beliefs onto finite-dimensional exponential family distributions which makes the inference tractable. Note however that the factor log-partition Eq.~\eqref{factor_log_partition} involves an integration over the high-dimensional factor $f_k(x_k ;y_k)$ and can still be a challenge to compute or approximate. See Section~\ref{sec:EP_modules} for the kind of factors that the \tramp package can currently handle.

\paragraph{Iterative schemes} The fixed point Eq.~\eqref{ep_fixed_point} consists of the moment matching constraint $\mu_i = \mu_i^k$ and the natural parameter constraint $(n_i - 1) \lambda_i = \sum_{k \in \partial i} \lambda_{i \to k}$.
The three optimization problems in Proposition~\ref{prop:free_energy} suggest different iterative schemes to reach this fixed point. As shown by \citep{Minka_free_2001, Heskes2005}, the optimization Eq.~\eqref{minka_variational} of the EP free energy naturally suggests the EP algorithm, where the natural parameter constraint is enforced at each iteration. The iterative scheme suggested by the \tramp free energy will be presented in Section~\ref{sec:tramp_iteration}. The direct minimization Eq.~\eqref{gibbs_variational} of the EC Gibbs free energy enforces the moment-matching constraint at each iteration. Note that the EC Gibbs free energy is in general not convex \citep{Opper2005NIPS} because:
\begin{equation}
    G[\mu_V] = \sum_{k \in F} \underbrace{G_k[\mu_k]}_{\text{convex}} + \sum_{i \in V} \underbrace{(1 - n_i)}_{\leq 0} \underbrace{G_i[\mu_i]}_{\text{convex}}
\end{equation}
Message passing procedures can only aim\footnote{Actually the EP and \tramp algorithms are not even guaranteed to converge, although damping the updates often works in practice. The double loop algorithm \citep{Heskes2002} is guaranteed to converge but is usually very slow.} at a local minimum. According to Eq.~\eqref{gibbs_variational}, the global minimum / minimizer is expected to give the best approximation $F_\phi(\mathbf{y})$ of the surprisal / posterior moments. The algorithm is said to be in a computational hard phase if on typical instances the message passing procedure converges towards a sub-optimal minimum.

\paragraph{Relationship to AMP} Finally we note that many AMP algorithms, such as LowRAMP \citep{Lesieur2017} for the low-rank matrix factorization problem, GAMP \citep{Zdeborova2016} for the GLM,  or \citep{Kabashima2016} for the extensive rank matrix factorization problem, also follow a free energy formulation. The corresponding AMP free energies can be shown to be equivalent to Proposition~\ref{prop:free_energy} through appropriate Legendre transformations using the moment/natural parameter duality. They therefore seek the same fixed point Eq.~\eqref{ep_fixed_point} but yield another iterative schemes (the AMP algorithms) to find this fixed point. The EC Gibbs free energy is actually equal to the so-called variational Bethe energy in AMP literature (Examples~\ref{example:glm_free_energy}-\ref{example:extensive_rank_free_energy}).  This is quite remarkable as the derivation of AMP algorithms  and corresponding variational Bethe free energies follows a different path. In contrast with the approach presented here, it starts by unfolding the factor graph at the level of individual scalar components (as a result the factor graph is dense and far from being a tree) and then considers the relaxation obtained in the asymptotic limit \citep{Zdeborova2016}. The tree decomposition in Proposition~\ref{prop:free_energy} (at the level of the high-dimensional factors and variables) is usually recovered "after the fact".

\begin{remark}[Variable neighbored by two factors] 
\label{remark:two_neighbors}
When the variable, say $x$, has two factor neighbors,
say $f_-$ and $f_+$, the fixed point messages must satisfy:
\begin{equation}
    \label{two_neighbors}
    \lambda_x^+ \doteq \lambda_{f_- \to x} = \lambda_{x \to f_+}, \qquad
    \lambda_x^- \doteq \lambda_{f_+ \to x} = \lambda_{x \to f_-}.
\end{equation}
\begin{figure}[!h]
    \centering
    \begin{tikzpicture}
      \node[latent] (x) {$x$} ;
      \factor[left=2 of x] {f} {$f_-$} {} {x} ;
      \factor[right=2 of x] {g} {$f_+$} {x} {} ;
      \node[above=0.5 of f.center] (x1) {} ;
      \node[above=0.5 of x.center] (y1) {};
      \draw[->, blue] (x1) -- (y1) node[midway,above] {$\lambda_x^+$} ;
      \node[above=0.5 of x.center] (x2) {} ;
      \node[above=0.5 of g.center] (y2) {};
      \draw[->, blue] (x2) -- (y2) node[midway,above] {$\lambda_x^+$} ;
      \node[below=0.5 of f.center] (x3) {} ;
      \node[below=0.5 of x.center] (y3) {};
      \draw[<-, green] (x3) -- (y3) node[midway,below] {$\lambda_x^-$} ;
      \node[below=0.5 of x.center] (x4) {} ;
      \node[below=0.5 of g.center] (y4) {};
      \draw[<-, green] (x4) -- (y4) node[midway,below] {$\lambda_x^-$} ;
    \end{tikzpicture}
\end{figure}
In many cases, for instance all the models in Figure~\ref{fig:tree_models} except \textbf{(b)}, each variable has only two factor neighbors, the reparametrization Eq.~\eqref{two_neighbors} allows to reduce by half the number of messages. We use this reparametrization in the Examples~\ref{example:glm_free_energy}-\ref{example:extensive_rank_free_energy} discussed below.
\end{remark}

\begin{example}[GLM]
\label{example:glm_free_energy}
The \tramp free energy for the GLM (Example~\ref{example:glm}), with isotropic Gaussian beliefs (Example~\ref{example:isotropic_beliefs}), is given by:
\begin{align}
    \notag
    &A[a_x^\pm, b_x^\pm, a_z^\pm, b_z^\pm] \\
    &= A_{p_0}[a_x^-, b_x^-] + A_W[a_x^+, b_x^+, a_z^-, b_z^-] + A_{p_\text{out}}[a_z^+, b_z^+]
    - A_x[a_x, b_x] - A_z[a_z, b_z] 
\end{align}
where $a_x = a_x^+ + a_x^-$, $b_x =  b_x^+ + b_x^-$ (idem $z$). 
The factor log-partitions are given in Appendix~\ref{app:modules}.
The EC Gibbs free energy is given by:
\begin{equation}
    G[r_x, \tau_x, r_z, \tau_z] = G_{p_0}[r_x, \tau_x] + G_{p_\text{out}}[r_z, \tau_z] 
    - \tilde{G}_W[r_x, \tau_x, r_z, \tau_z]
\end{equation}
with $\tilde{G}_W = G_x + G_z - G_W$ given in Appendix~\ref{module:gibbs_linear} for a generic linear channel. In particular when the matrix $W$ has iid entries (Appendix~\ref{module:iid_entries}) one recovers exactly the variational Bethe free energy of \citep{krzakala2014variational} which is shown to be equivalent to the GAMP free energy.
\end{example}

\begin{example}[Low rank matrix factorization]
\label{example:low_rank_free_energy}
The \tramp free energy for the low rank factorization (Example~\ref{example:glm}), with isotropic (Example~\ref{example:isotropic_beliefs}) or structured (Example~\ref{example:structured_beliefs}) Gaussian beliefs on $U$ and $V$, and diagonal Gaussian beliefs (Example~\ref{example:diagonal_beliefs}) on $Z$, is given by:
\begin{align}
\notag
    &A[a_U^\pm, b_U^\pm, a_V^\pm, b_V^\pm] \\
    &= 
    A_{p_U}[a_U^-, b_U^-] + A_{p_V}[a_V^-, b_V^-] 
    + A_\delta[a_U^+, b_U^+, a_V^+, b_V^+; a_Z^-, b_Z^-]
    - A_U[a_U, b_U] - A_V[a_V, b_V] 
\end{align}
where $a_U = a_U^+ + a_U^-$, $b_U =  b_U^+ + b_U^-$ (idem $V$).
The factor log-partitions are given in Appendix~\ref{app:modules}. In the large $N$ limit, the output channel do not contribute to the free energy and its net effect is to send the constant messages:
\begin{equation}
    a_Z^- = - \partial_z^2 \ln p_\text{out}(y \mid z) \vert_{z=0, y=Y}, \quad
    b_Z^- = \partial_z \ln p_\text{out}(y \mid z) \vert_{z=0, y=Y}, 
\end{equation}
a phenomenon known as channel universality in \citep{Lesieur2017}. The messages $a_Z^-, b_Z^- \in \mathbb{R}^{N \times M}$ are thus viewed as fixed parameters.
The EC Gibbs free energy is given by:
\begin{equation}
    G[r_U, \tau_U, r_V, \tau_V] = G_{p_U}[r_U, \tau_U] + G_{p_V}[r_V, \tau_V] - \tilde{G}_\delta[r_U, \tau_U, r_V, \tau_V; a_Z^-, b_Z^-]
\end{equation}
where $\tilde{G}_\delta = G_U + G_V - G_\delta$ is derived in \citep{Lesieur2017} using a  Plefka-Georges-Yedidia expansion \citep{plefka1982convergence, georges1991expand}, which is asymptotically exact in the large $N$ limit:
\begin{align}
    \notag
    &\tilde{G}_\delta[r_U, \tau_U, r_V, \tau_V; a_Z^-, b_Z^-] \\
    \label{lowramp_gibbs}
    &= \frac{1}{2} 
    \sum_{n=1}^N \sum_{m=1}^M
    \frac{(b_Z^-)_{nm}^2}{N} \Tr (\Sigma_U^n \Sigma_V^m) 
    + 2\frac{(b_Z^-)_{nm}}{\sqrt{N}} (r_U r_V^T)_{nm} 
    - \frac{(a_Z^-)_{nm}}{N} \Tr \tau_U^n \tau_V^m 
\end{align}
One recovers exactly\footnote{We think there is a typo in Eq.~(111) of \citep{Lesieur2017} and that Eq.~\eqref{lowramp_gibbs} is the correct expression.} the variational Bethe free energy of \citep{Lesieur2017} which is shown to be equivalent to the LowRAMP free energy. 
\end{example}

\begin{example}[Extensive rank matrix factorization]
\label{example:extensive_rank_free_energy}
The \tramp free energy for the extensive rank factorization (Example~\ref{example:extensive_rank}), with diagonal Gaussian beliefs (Example~\ref{example:diagonal_beliefs}), is given by:
\begin{align}
\notag
    A[a_F^\pm, b_F^\pm, a_X^\pm, b_X^\pm, a_Z^\pm, b_Z^\pm, ] &=
    A_{p_F}[a_F^-, b_F^-] + A_{p_X}[a_X^-, b_X^-] 
    + A_\delta[a_F^+, b_F^+, a_X^+, b_X^+, a_Z^-, b_Z^-] \\
    &- A_F[a_F, b_F] - A_X[a_X, b_X] - A_Z[a_Z, b_Z]
\end{align}
where $a_F = a_F^+ + a_F^-$, $ b_F =  b_F^+ + b_F^-$ (idem $X,Z$). The factor log-partitions are given in Appendix~\ref{app:modules}. Contrary to the low rank case, the output channel does contribute to the free energy and the messages $a_Z^-, b_Z^-$ are no longer constant.
The EC Gibbs free energy is given by:
\begin{equation}
    G[r_F, \tau_F, r_X, \tau_X] = G_{p_F}[r_F, \tau_F] + G_{p_X}[r_X, \tau_X] + G_{p_Z}[r_Z, \tau_Z] - \tilde{G}_\delta[r_F, \tau_F, r_X, \tau_X, r_Z, \tau_Z]
\end{equation}
One recovers exactly the variational Bethe free energy of \citep{Kabashima2016}, which is shown to be equivalent to the AMP free energy, with $\tilde{G}_\delta = G_F + G_X + G_Z - G_\delta$ given by:
\begin{align}
    \notag
    &\tilde{G}_\delta[r_F, \tau_F, r_X, \tau_X, r_Z, \tau_Z] = 
    -\frac{1}{2} \sum_{m=1}^M \sum_{p=1}^P
    \frac{v_Z^{m p}}{V_{m p}} + \ln 2\pi V_{m p} + \frac{(v_F v_X)_{m p}}{N} g_{m p}^2
    \\
    \label{extensive_rank_gibbs}
    &\text{with} \quad
    g  = \frac{r_Z- \tfrac{r_F r_X}{\sqrt{N}} }{ \tfrac{v_F v_X}{N}}, 
    \quad
    V = \frac{v_F v_X + r_F^2 v_X + v_F r_X^2}{N} = \frac{\tau_F \tau_X - r_F^2 r_X^2}{N} .
\end{align}
However we warn the reader that the derivation of \citep{Kabashima2016} wrongly assumes that $Z$ behaves as a multivariate Gaussian as pointed out by \citep{Maillard2021}. In consequence the free energy is not asymptotically exact and we expect the AMP algorithm of \citep{Kabashima2016} to be sub-optimal. See \citep{Maillard2021} for the corrections to the expression Eq.~\eqref{extensive_rank_gibbs} of $\tilde{G}_\delta$.
\end{example}

\subsection{Tree-AMP implementation of Expectation Propagation \label{sec:tramp_iteration}}

The \tramp implementation of EP works with the full set of messages
$(\lambda_{k \to i}, \lambda_{i \to k})_{(i,k) \in E}$. 
% In the parametrization Eq.~\eqref{message_param} we recall that the factor natural [aramater is given by $\lambda_k=(\lambda_{i \to k})_{i \in \partial k}$, the edge natural parameter  by $\lambda_i^k = \lambda_{i \to k} + \lambda_{k \to i}$, and the variable natural parameter by $\lambda_i = \sum_{k \in \partial i} \lambda_{k \to i}$. 
Due to the parametrization Eq.~\eqref{message_param} and the moment functions Eqs~\eqref{variable_moment} and \eqref{factor_moment_explicit}, a stationary point of $A[\lambda_E]$ satisfies:
\begin{align}
  &\partial_{\lambda_{k \to i}} A[\lambda_E] = 0 \implies
  \mu_i[\lambda_i^k] = \mu_i[\lambda_i] \implies
  \lambda_i^k = \lambda_i \,,\\
  &\partial_{\lambda_{i \to k}} A[\lambda_E]  = 0 \implies
  \mu_i[\lambda_i^k] = \mu_i^k[\lambda_k]  \implies
  \lambda_i^k = \lambda_i[\mu_i^k[\lambda_k]]\,,
\end{align}
which suggests the iterative procedure summarized in Algorithm~\ref{algo:tramp},
where $E_+$ denotes a topological ordering of the edges and $E_-$
the reverse ordering.

The message-passing schedule seems the most natural: iterate over the
edges in topological order (forward pass) then iterate in reverse
topological order (backward pass) and repeat until convergence.
In fact, if the exponential beliefs and the factors are
conjugate\footnote{For example Gaussian
beliefs and the factors are either linear transform or Gaussian noise, prior or
likelihood} then the moment-matching is exact and Algorithm~\ref{algo:tramp} is 
actually equivalent to exact belief
propagation, where one forward pass and one backward pass yield the
exact marginals \citep{Bishop2006}.

\vspace{1em}
\begin{algorithm}[H]
\label{algo:tramp}
\DontPrintSemicolon
\SetKwInput{KwInput}{input}
\SetKwInput{KwOutput}{output}
\SetKw{Initialize}{initialize}
%\KwInput{observations $\mathbf{y}$}
%\KwOutput{posterior natural parameters $(\lambda_i)_{i \in V}$}
\Initialize{$\lambda_{i\to k},\lambda_{k\to i} = 0$}\\
\Repeat{convergence}{
  \ForEach(\tcp*[f]{ forward and backward pass}){edge $e \in E_+ \cup E_-$}{
    \If{$e = k \to i$}{
      $\lambda_i^k = \lambda_i[\mu_i^k[\lambda_k]]$
      \tcp*[f]{moment-matching }\\
      $\lambda_{k \to i}^\text{new} = \lambda_i^k - \lambda_{i \to k}$
      \tcp*[f]{message $f_k \to x_i$ }
    }
    \If{$e = i \to k$}{
      $\lambda_i = \sum_{k' \in \partial i} \lambda_{k' \to i}$
      \tcp*[f]{posterior }\\
      $\lambda_{x \to f}^\text{new} = \lambda_i - \lambda_{k \to i}$
      \tcp*[f]{message $x_i \to f_k$ }
    }
  }
}
\caption{Generic \tramp algorithm}
\end{algorithm}
\vspace{1em}

For isotropic or diagonal Gaussian beliefs (Examples~\ref{example:isotropic_beliefs} and \ref{example:diagonal_beliefs}) the moment-matching is equivalent to match the mean
$r_i = r_i^k$ and variance $v_i = v_i^k$ which leads to Algorithm~\ref{algo:gaussian_EP}. 
When the variable $x_i$ has only two neighbor factors, say $f_k$ and $f_l$,
the $x_i \to f_k$ update is particularly simple. The variable just passes through the corresponding messages:
\begin{equation}
    \label{message_x_pass}
    a_{i \to k}^\text{new} = a_{l \to i}, \quad
    b_{i \to k}^\text{new} = b_{l \to i}
\end{equation}
in compliance with Remark~\ref{remark:two_neighbors}. Algorithm~\ref{algo:gaussian_EP} can be straightforwardly extended to full covariance and structured Gaussian beliefs (Example~\ref{example:full_cov_beliefs} and \ref{example:structured_beliefs}) using the moment-matching update
\begin{equation*}
    a_i^k = v_i^k[a_k,b_k]^{-1}, \quad b_i^k = v_i^k[a_k,b_k]^{-1} r_i^k[a_k, b_k]
\end{equation*}
where $v_i^k$ is now a covariance matrix. 

\vspace{1em}
\begin{algorithm}[H]
\label{algo:gaussian_EP}
\DontPrintSemicolon
\SetKwInput{KwInput}{input}
\SetKwInput{KwOutput}{output}
\SetKw{Initialize}{initialize}
%\KwInput{observations $\mathbf{y}$}
%\KwOutput{posterior natural parameters $(a_i, b_i)_{i \in V}$}
\Initialize{$a_{i\to k},b_{i \to k},a_{k\to i},b_{k \to i} = 0$}\\
\Repeat{convergence}{
  \ForEach(\tcp*[f]{ forward and backward pass}){edge $e \in E_+ \cup E_-$}{
    \If{$e = k \to i$}{
      $a_i^k = \frac{1}{v_i^k[a_k, b_k]}, \quad b_i^k = \frac{r_i^k[a_k, b_k]}{v_i^k[a_k, b_k]}$
      \tcp*[f]{moment-matching }\\
      $a_{k \to i}^\text{new} = a_i^k - a_{i \to k}, \quad b_{k \to i}^\text{new} = b_i^k - b_{i \to k}$
      \tcp*[f]{message $f_k \to x_i$ }
    }
    \If{$e = i \to k$}{
      $a_i = \sum_{k' \in \partial i} a_{k' \to i}, \quad b_i = \sum_{k' \in \partial i} b_{k' \to i} $
      \tcp*[f]{posterior }\\
      $a_{i \to k}^\text{new} = a_i - a_{k \to i}, \quad b_{i \to k}^\text{new} = b_i - b_{k \to i}$
      \tcp*[f]{message $x_i \to f_k$ }
    }
  }
}
\caption{Expectation propagation in \tramp (Gaussian beliefs)}
\end{algorithm}

\subsection{MAP estimation \label{sec:map_estimation}}

The \tramp Algorithm can also be used for maximum at posteriori (MAP) estimation, where the mode of the posterior Eq.~\eqref{factor_graph_conditional}
is the solution to the energy minimization problem:
\begin{equation}
    \label{map_problem}
    \mathbf{x}^* = \argmin_{ \mathbf{x}} E( \mathbf{x},  \mathbf{y}), \quad E( \mathbf{x},  \mathbf{y}) =\sum_{k \in F} E_k(x_k ;y_k)
\end{equation}
where $E_k(x_k ; y_k)$ is the energy associated to the factor $f_k(x_k ; y_k) = e^{-E_k(x_k ;y_k)}$. Conversely any optimization problem of the form Eq.~\eqref{map_problem}, with a tree-structured graph of penalties/constraints, can be viewed as the MAP estimation of Eq.~\eqref{factor_graph_conditional} with pseudo-factors $f_k(x_k ; y_k) = e^{-E_k(x_k ;y_k)}$. Following \citep{Manoel2018} for the derivation of the TV-VAMP algorithm, that we generalize to tree-structured models in Appendix~\ref{app:map_estimation}, the MAP estimation / energy minimization can be derived by introducing an inverse temperature $\beta$ in the posterior Eq.~\eqref{factor_graph_conditional} and considering the zero temperature limit,  as usually done in statistical physics literature \citep{mezard2009information}.

\begin{proposition}[MAP estimation in \tramp]
\label{prop:map_estimation}
The energy minimization / MAP estimation problem can be formulated as the $\beta \to \infty$ limit of Proposition~\ref{prop:free_energy}:
\begin{equation}
    \min_{\mathbf{x}} E(\mathbf{x}, \mathbf{y}) = \min \extr_{a_E, b_E} - A[a_E, b_E] 
\end{equation}
with MAP variable log-partition, mean and variance:
\begin{equation}
    A_i[a_i, b_i] = \frac{\Vert b_i \Vert^2}{2a_i}, \quad
    r_i[a_i, b_i] = \frac{b_i}{a_i}, \quad
    v_i[a_i, b_i] = \frac{1}{a_i}, 
\end{equation}
and MAP factor log-partition, mean and variance:
\begin{align}
   \label{map_log_partition}
    &A_k[a_k, b_k] =  \frac{\Vert b_k\Vert^2}{2a_k} 
    - \mathcal{M}_{\tfrac{1}{a_k}E_k(\cdot\; ; \; y_k)}\left( \frac{b_k}{a_k} \right) ,
    \\
    \label{map_mean_variance}
    &r_k[a_k, b_k] = \mathrm{prox}_{\tfrac{1}{a_k}E_k(\cdot\; ; \; y_k)}\left( \frac{b_k}{a_k} \right) , \quad
    v_k[a_k, b_k] = \langle \partial_{b_k} r_k[a_k, b_k] \rangle ,
\end{align}
where we introduce the Moreau envelop 
$\mathcal{M}_g(y) = \min_x \{g(x) + \tfrac{1}{2}\Vert x - y \Vert^2\}$ 
and the proximal operator
$\mathrm{prox}_g(y) = \argmin_x \{g(x) + \tfrac{1}{2}\Vert x - y \Vert^2\} $.
A stationary point of $A[a_E, b_E]$ can be searched with the \tramp Algorithm~\ref{algo:gaussian_EP}. At the optimal fixed point, the "means" $(r_i)_{i \in V}$ actually yield the MAP estimate / minimizer $\mathbf{x}^* = (x_i^*)_{i \in V}$.
\end{proposition}

The TV-VAMP algorithm is recovered as a special case (Section~\ref{sec:related_algos}).
See the discussion in \citep{Manoel2018} for the close relationship to proximal methods in optimization \citep{Parikh2014}, in particular the "variances" $(v_i)_{i \in V}$ can be viewed as adaptive stepsizes in the Peaceman-Rachford splitting.

\subsection{Expectation Propagation modules \label{sec:EP_modules}}

In the weak consistency framework, we have the freedom to choose any kind of approximate beliefs, that is choose a set of sufficient statistics $\phi_i$ for each variable $x_i$.
Each choice leads to a different approximate inference scheme,
but all are implemented  by the same message passing Algorithm~\ref{algo:tramp}.
Of course the algorithm requires each relevant module to be implemented:
in practice the variable $x_i$ (resp. factor $f_k$) module should be able to compute
the log-partition $A_i[\lambda_i]$ (resp. $A_k[\lambda_k]$) and its associated moment function $\mu_i[\lambda_i]$ (resp. $\mu_k[\lambda_k]$). Note that the definition of the module directly depends on the choice 
of sufficient statistics $\phi$, so choosing a different kind of approximate beliefs 
actually leads to a distinct module. Currently the \tramp package only supports isotropic Gaussian beliefs (Example~\ref{example:isotropic_beliefs}) but we plan to include more generic Gaussian beliefs (Examples~\ref{example:diagonal_beliefs}-\ref{example:structured_beliefs}) 
in future versions of the package. The corresponding variable log-partition $A_i[a_i, b_i]$, mean $r_i[a_i, b_i]$ and variance $r_i[a_i, b_i]$ and factor log-partition $A_k[a_k, b_k]$, mean $r_k[a_k, b_k]$ and variance $v_k[a_k, b_k]$ are presented in Example~\ref{example:duality_isotropic}. The implementation is detailed in the documentation\footnote{ See \href{https://sphinxteam.github.io/tramp.docs/0.1/html/implementation.html}{https://sphinxteam.github.io/tramp.docs/0.1/html/implementation.html}}. We list below the modules considered in \tramp.

\paragraph{Variable modules}
A list of approximate beliefs is presented in 
Appendix~\ref{module:variable}.
As shown, such beliefs can be defined over many types of variable: binary, sparse, real, constrained to an interval, or circular
for instance. The associated variable modules correspond to well known exponential family distributions,
including the Gauss-Bernoulli for a sparse variable. Even if the \tramp package only implements isotropic Gaussian beliefs, the variable modules are useful to derive the factor
modules. 

\paragraph{Analytical vs approximate factor modules}
Note that the factor log-partition Eq.~\eqref{factor_log_partition} involves the high-dimensional factor $f_k(x_k ;y_k)$ and can thus be a challenge to compute or approximate.
Nonetheless, many factor modules implemented in the \tramp package can be analytically derived, which means providing an explicit formula for the factor log-partition
$A_k[a_k, b_k]$, mean $r_k[a_k, b_k]$ and variance $v_k[a_k, b_k]$. The following analytical modules are derived in Appendix~\ref{app:modules}:
\begin{itemize}
    \item linear channels which include the rotation channel, the discrete Fourier transform and convolutional filters as special cases;
    \item separable priors such as the Gaussian, binary, Gauss-Bernoulli, and positive priors;
    \item separable likelihoods such as the Gaussian or a deterministic likelihood like observing the sign, absolute value, modulus or phase;
    \item separable channels such additive Gaussian noise or the piecewise linear activation channel.
\end{itemize}
For other modules, that we did not manage to obtain analytically, one resorts to an approximation or an algorithm to estimate the log-partition $A_k[a_k,b_k]$ and the associated mean $r_k[a_k, b_k]$ and variance $v_k[a_k, b_k]$.
One such example in the \tramp package is the low rank factorisation module $Z = \frac{UV^\intercal}{\sqrt{N}}$, for which we use the AMP algorithm developed in \citep{Lesieur2017} to estimate $A_k[a_k, b_k]$, $r_k[a_k, b_k]$ and $v_k[a_k, b_k]$.

\paragraph{MAP modules} The maximum a posteriori (MAP) modules are worth mentioning especially due to their connection to proximal methods in optimization \citep{Parikh2014}.  They are of course used in MAP estimation (Section~\ref{sec:map_estimation}) -- where all modules are MAP modules -- or can be used in isolation for a specific factor to approximate. Indeed, for any factor $f_k(x_f; y_k)  = e^{-E_k(x_k ;y_k)}$, one can use the Laplace method to obtain the MAP approximation Eqs~\eqref{map_log_partition}-\eqref{map_mean_variance} to the log-partition, mean and variance. 
Two such MAP modules are implemented in the \tramp package for the penalties $E(x) = \lambda \Vert x \Vert_1$ and $E(x) = \lambda \Vert x \Vert_{2,1}$ associated to the $\ell_1$ and $\ell_{2,1}$ norms. We recall that the  $\ell_{2,1}$ norm is defined as:
\begin{equation}
    \Vert x \Vert_{2,1} = \sum_{n=1}^N  \Vert x_n \Vert_2 =  \sum_{n=1}^N \sqrt{ \sum_{l=1}^d x_{ln}^2 } \quad \text{for} \quad x \in \mathbb{R}^{d \times N}.
\end{equation}
The corresponding proximal operators are the soft thresholding and group soft thresholding operators.

\subsection{Related algorithms \label{sec:related_algos}}

We recover several algorithms as special cases of Algorithm~\ref{algo:gaussian_EP}. 
For instance the G-VAMP \citep{Schniter2016},
TV-VAMP \citep{Manoel2018} and
ML-VAMP \citep{Fletcher2018} algorithms correspond respectively to the factor
graphs Figure~\ref{fig:tree_models}~(\textbf{a}), (\textbf{b}) and (\textbf{d}). 
In this subsection, we make explicit the equivalence with these algorithms and argue that the modularity of \tramp allows to tackle a greater variety of inference tasks and optimization problems.

\begin{figure}[t]
  \centering
   (\textbf{a})\hspace{1cm}
    \begin{tikzpicture}
    \node[latent]                     (z0)  {$z_0$} ;
    \node[latent, right=of x]         (z1)  {$z_1$} ;
    \node[latent, right=of z1]        (z2)  {$z_2$} ;
    \node[latent, right=of z2]        (d)  {$\cdots$} ;
    \node[latent, right=of d]         (zm)  {$z_{L-1}$} ;
    \node[obs, right=of L]            (zL)  {$z_L$} ;

    \factor[left=of z0] {p} {$p_0$} {} {z0} ;
    \factor[left=of z1] {p} {$p_1$} {z0} {z1} ;
    \factor[left=of z2] {p} {$p_2$} {z1} {z2} ;
    \factor[left=of d]  {p} {} {z2} {d} ;
    \factor[left=of zm] {p} {$p_{L-1}$} {d} {zm} ;
    \factor[left=of zL] {out} {$p_L$} {zm} {zL} ;
  \end{tikzpicture}
  
  \vspace{1cm}
 \begin{minipage}[c]{0.45\linewidth}
    (\textbf{b}) \hspace{1cm}
    \begin{tikzpicture}
    \node[latent]  (x)  {$z_l$} ;
    \factor[left=1.8 of x.center] {p} {left:$p_l$} {x} {} ;
    \factor[right=1.8 of x.center] {q} {right:$p_{l+1}$} {x} {} ;
    \node[below=0.5 of p.center]  (p1) {} ;
    \node[right=1.3 of p1] (p2) {};
    \node[below=0.5 of q.center]  (q1) {};
    \node[left=1.3 of q1]  (q2) {};
    \node[above=0.5 of p.center]  (p3) {} ;
    \node[right=1.3 of p3] (p4) {};
    \node[above=0.5 of q.center]  (q3) {};
    \node[left=1.3 of q3]  (q4) {};
    \draw[->, green] (p1) -- (p2) node[midway,below] {$a_l^+ b_l^+$}  ;
    \draw[->, green] (q2) -- (q1) node[midway,below] {$a_l^+ b_l^+$}  ;
    \draw[->, red] (p4) -- (p3) node[midway,above] {$a_l^- b_l^-$}  ;
    \draw[->, red] (q3) -- (q4) node[midway,above] {$a_l^- b_l^-$}  ;
    \end{tikzpicture}
    \end{minipage}
    \begin{minipage}[c]{0.45\linewidth}
    (\textbf{c})\hspace{1cm}
    \begin{tikzpicture}
    \node[latent] (x) {$z_{l-1}$} ;
    \node[latent, right=3.6 of x.center] (y) {$z_l$} ;
    \factor[left=1.8 of y.center] {p} {$p_l$} {x,y} {} ;
    \node[below=0.5 of x.center]  (x1) {} ;
    \node[right=1.6 of x1] (x2) {};
    \node[below=0.5 of y.center]  (y1) {};
    \node[left=1.6 of y1]  (y2) {};
    \node[above=0.5 of x.center]  (x3) {} ;
    \node[right=1.6 of x3] (x4) {};
    \node[above=0.5 of y.center]  (y3) {};
    \node[left=1.6 of y3]  (y4) {};
    \draw[->, blue] (x1) -- (x2) node[midway,below] {$a_{l-1}^+ b_{l-1}^+$}  ;
    \draw[->, blue] (y1) -- (y2) node[midway,below] {$a_l^- b_l^-$}  ;
    \draw[->, red] (x4) -- (x3) node[midway,above] {$a_{l-1}^- b_{l-1}^-$}  ;
    \draw[->, green] (y4) -- (y3) node[midway,above] {$a_l^+ b_l^+$}  ;
  \end{tikzpicture}
\end{minipage}
   \caption{
   Message passing in ML-VAMP.
    (\textbf{a}) Multi-layer generalized linear model. 
    (\textbf{b}) The variable $z_l$ passes 
    through the forward message $a_l^+ b_l^+$ (green) during the forward pass, and the backward message $a_l^- b_l^-$ (red) during the backward pass.
    (\textbf{c}) The factor $p_l$ always takes as inputs the messages $a_{l-1}^+ b_{l-1}^+$ and $a_l^- b_l^-$ (blue). It outputs the message  $a_l^+ b_l^+$ (green) during the forward pass, and the message $a_{l-1}^- b_{l-1}^-$ (red) during the backward pass.
  }
  \label{fig:ml_vamp_message}
\end{figure}
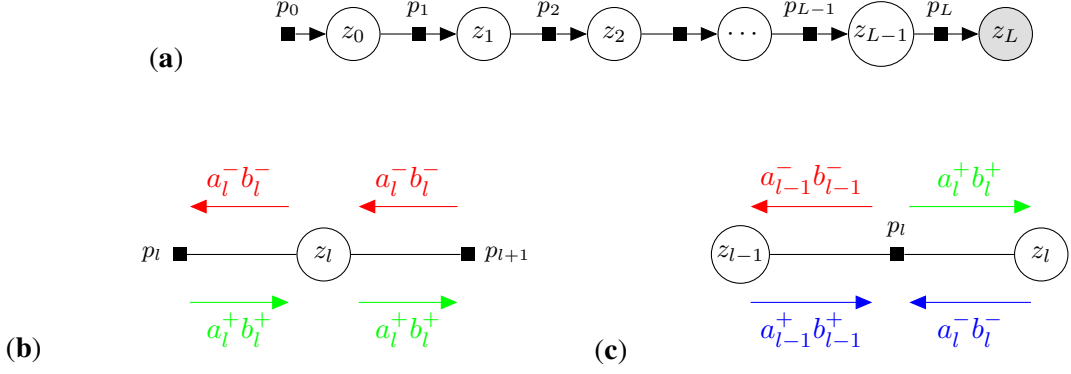

\paragraph{Inference in multi-layer network} The ML-VAMP algorithm \citep{Fletcher2018} performs inference in multi-layer networks such as Figure~\ref{fig:tree_models}~(\textbf{d}). 
The one layer case reduces to the G-VAMP algorithm \citep{Schniter2016} for GLM such as Figure~\ref{fig:tree_models}~(\textbf{a}). 
Following \cite{Fletcher2018} let us consider the multi-layer model:
\begin{equation}
    p(\mathbf{z}) = \prod_{l=0}^L p_l(z_l \mid z_{l-1})
\end{equation}
where $\mathbf{z}= \{z_l\}_{l=0}^L$ includes the measurement $y = z_L$ and the signals $\mathbf{x} = \{z_l\}_{l=0}^{L-1}$ to infer. The corresponding factor graph is displayed in Figure~\ref{fig:ml_vamp_message}~(\textbf{a}). The model generally consists of a succession of linear channels (with possibly a bias and additive Gaussian noise) and separable non-linear activations; 
however, it is not necessary to specify further the architecture as all factors are treated on the same footing in both ML-VAMP and Algorithm~\ref{algo:gaussian_EP}.

We are interested in the isotropic Gaussian beliefs version of Algorithm~\ref{algo:gaussian_EP}.  According to Eq.~\eqref{message_x_pass}, the $z_l \to p_{l+1}$ update during the forward pass leads to
\begin{equation}
    a_l^+ \doteq a_{z_l \to p_{l+1}} = a_{p_l \to z_l} , 
    \quad 
    b_l^+ \doteq b_{z_l \to p_{l+1}} = b_{p_l \to z_l},
\end{equation}
while the $z_l \to p_l$ update  during the backward pass leads to
\begin{equation}
    a_l^- \doteq a_{z_l \to p_l} = a_{p_{l+1} \to z_l}  , 
    \quad 
    b_l^- \doteq b_{z_l \to p_l} = b_{p_{l+1} \to z_l} .
\end{equation}
Each variable in Figure~\ref{fig:ml_vamp_message}~(\textbf{a}) has exactly two neighbors, so each variable just passes through the corresponding messages as illustrated in Figure~\ref{fig:ml_vamp_message}~(\textbf{b}). 

The $p_l \to z_l$ update during the forward pass leads to
\begin{align}
    \label{pl_forward_moment}
    &r_l = r_{z_l}^{p_l}[a_{l-1}^+ , b_{l-1}^+ ,a_{l}^- , b_{l}^-], \quad 
    v_l = v_{z_l}^{p_l}[a_{l-1}^+ , b_{l-1}^+ , a_{l}^- , b_{l}^-], \\
     \label{pl_forward_update}
    &a_l = \frac{1}{v_l}, \quad 
    b_l = \frac{r_l}{v_l}, \quad
    a_l^+ = a_l - a_l^-, \quad 
    b_l^+ = b_l - b_l^-. 
\end{align}
while the $p_l \to z_{l-1}$ update  during the backward pass leads to
\begin{align}
    \label{pl_backward_moment}
    &r_{l-1} = r_{z_{l-1}}^{p_l}[a_{l-1}^+ , b_{l-1}^+ , a_{l}^- , b_{l}^-], \quad 
    v_{l-1} = v_{z_{l-1}}^{p_l}[a_{l-1}^+  , b_{l-1}^+ , a_{l}^- , b_{l}^-],  \\
    \label{pl_backward_update} 
    &a_{l-1} = \frac{1}{v_{l-1}} \quad 
    b_{l-1} = \frac{r_{l-1}}{v_{l-1}}, \quad
    a_{l-1}^- = a_{l-1} - a_{l-1}^+, \quad 
    b_{l-1}^- = b_{l-1} - b_{l-1}^+. 
\end{align}
as illustrated in Figure~\ref{fig:ml_vamp_message}~(\textbf{c}). 
The ML-VAMP forward pass computes $r_l$ and $v_l$ using Eq.~\eqref{pl_forward_moment} and updates the message according to Eq.~\eqref{pl_forward_update} where these quantities in \citep{Fletcher2018} are denoted by:
\begin{equation}
a_l^\pm = \gamma_l^\pm,  \quad 
b_l^\pm = \gamma_l^\pm r_l^\pm, \quad
r_l = \hat{z}_l^+, \quad 
a_l = \eta_l^+ .
\end{equation}
Similarly the backward pass in the ML-VAMP algorithm is equivalent to Eqs~\eqref{pl_backward_moment} and \eqref{pl_backward_update}. Finally the equivalence also holds for the prior $p_0(z_0)$
and the likelihood $p_L(y|z_{L-1})$. The prior is only used during the forward pass, it receives the
backward message $a_0^-,b_0^-$ as input and outputs the forward message $a_0^+,b_0^+$. The likelihood is only used during the backward pass, it receives the
forward message $a_{L-1}^+,b_{L-1}^+$ as input and outputs the backward message $a_{L-1}^-,b_{L-1}^-$.

Therefore the EP Algorithm~\ref{algo:gaussian_EP} with isotropic Gaussian beliefs is exactly equivalent to the ML-VAMP algorithm. It offers a direct generalization to any tree-structured model, for instance the tree network of GLMs considered in \citep{Reeves2017}.

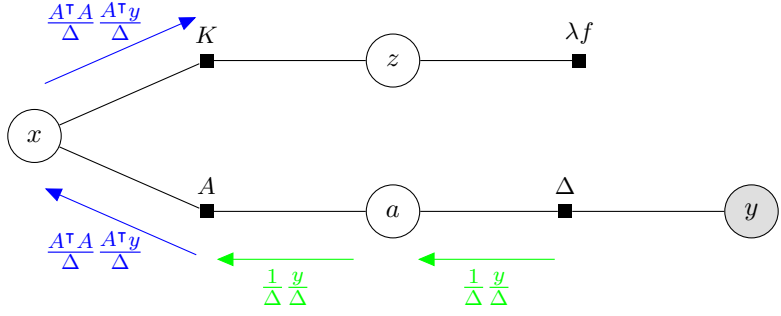
\begin{figure}[t]
  \centering
  \begin{tikzpicture}
    \node[latent] (x) {$x$} ;
    \node[latent, right=4 of x, yshift=+1cm] (z) {$z$} ;
    \node[latent, right=4 of x, yshift=-1cm] (a) {$a$} ;
    \node[obs, right=4 of a] (y) {$y$} ;
    \factor[left=2 of z] {K} {$K$} {x, z} {};
    \factor[left=2 of a] {A} {$A$} {x, a} {} ;
    \factor[left=2 of y] {D} {$\Delta$} {a, y} {} ;
    \factor[right=2 of z] {f} {$\lambda f$} {z} {} ;
    \node[below=0.5 of D.center]  (x1) {} ;
    \node[left=1.8 of x1] (x2) {};
    \draw[->, green] (x1) -- (x2) node[midway,below] {$\tfrac{1}{\Delta}\tfrac{y}{\Delta}$} ;
    \node[below=0.5 of A.center]  (x3) {} ;
    \node[right=1.8 of x3] (x4) {};
    \draw[->, green] (x4) -- (x3) node[midway,below] {$\tfrac{1}{\Delta}\tfrac{y}{\Delta}$} ;
    \node[below=0.5 of A.center] (x5) {} ;
    \node[below=0.5 of x.center] (x6) {};
    \draw[->, blue] (x5) -- (x6) node[midway,below,xshift=-2ex] {$\tfrac{A^\intercal A}{\Delta}\tfrac{A^\intercal y}{\Delta}$} ;
    \node[above=0.5 of x.center] (x7) {} ;
    \node[above=0.5 of K.center] (x8) {};
    \draw[->, blue] (x7) -- (x8) node[midway,above,xshift=-2ex] {$\tfrac{A^\intercal A}{\Delta}\tfrac{A^\intercal y}{\Delta}$} ;
  \end{tikzpicture}
  \caption{
    Message passing in TV-VAMP.
  }
  \label{fig:tv_vamp_message}
\end{figure}

\paragraph{Optimization with non-separable penalties}
We now turn to the TV-VAMP algorithm \citep{Manoel2018} designed to solve optimization problem of the form:
\begin{equation}
    x^* = \argmin \frac{1}{2\Delta} \Vert y - Ax \Vert^2 + \lambda f(Kx).
\end{equation}
This corresponds to the MAP estimate (Section~\ref{sec:map_estimation}) for the factor graph displayed in Figure~\ref{fig:tv_vamp_message}. Of particular interest is the case $K = \nabla$ and $f(z) = \Vert z \Vert_{2,1}$ which is identical to the total variation penalty for $x$. 

We are interested in the version of Algorithm~\ref{algo:gaussian_EP} with isotropic Gaussian beliefs on all variables except $x$ for which we consider a full covariance belief.
The penalty term $\lambda f$ would correspond to a factor $e^{-\lambda f(z)}$ in a probabilistic setting, but here we are only considering the MAP module for which the mean and variance are given by Eq.~\eqref{map_mean_variance}:
\begin{equation}
r_f[a_f , b_f] 
= \eta_{\tfrac{\lambda}{a_f}} 
\left( \frac{b_f}{a_f} \right), \quad
v_f[a_f , b_f]
= \frac{1}{a_f} 
\left\langle 
\nabla\eta_{\tfrac{\lambda}{a_f}} 
\left( \frac{b_f}{a_f} \right) 
\right\rangle ,
\end{equation}
where we introduce the function $\eta_{\lambda}(x) = \mathrm{prox}_{\lambda f} (x)$ following \cite{Manoel2018}.

First note that each variable in Figure~\ref{fig:tv_vamp_message} has exactly two neighbors. According to Eq.~\eqref{message_x_pass} the message passing is then particularly simple: the variable just passes through the corresponding messages.
The Gaussian likelihood $\Delta$
(Appendix~\ref{module:gaussian_likelihood}) leads to the messages:
\begin{equation}
 a_{a \to A} = a_{\Delta \to a}  = \frac{1}{\Delta}, \quad 
 b_{a \to A} = b_{\Delta \to a}  = \frac{y}{\Delta},
\end{equation}
and the linear channel $A$ with full covariance belief on $x$ (Appendix~\ref{module:linear_full_cov})
leads to the messages:
\begin{equation}
 a_{x \to K} = a_{A \to x}  = \frac{A^\intercal A}{\Delta}, \quad
 b_{x \to K} = b_{A \to x}  = \frac{A^\intercal y}{\Delta}.
\end{equation}
This stream of constant messages from the likelihood $\Delta$ up to factor $K$ is displayed on Figure~\ref{fig:tv_vamp_message}. 
The $K \to z$ update for the linear channel $K$ with isotropic belief on $z$ (Appendix~\ref{module:linear}) leads to the messages:
\begin{align}
    \label{tv_vamp_rx}
    &r_x^K =  \Sigma_x^K \left[
        \frac{A^\intercal y}{\Delta} + K^\intercal  b_{z \to K}    
    \right], \quad
    \Sigma_x^K = \left[
        \frac{A^\intercal A}{\Delta} + a_{z \to K} K^\intercal  K    
    \right]^{-1} , \\
    &r_z^K = K r_x^K, \quad
    v_z^K = \frac{1}{N_z} \mathrm{Tr}\left[ K \Sigma_x^K K^\intercal \right] , \\ 
    &a_{z \to f} = a_{K \to z} = \frac{1}{v_z^K} - a_{z \to K}, \quad
    b_{z \to f} = b_{K \to z} = \frac{r_z^K}{v_z^K} - b_{z \to K},
\end{align}
and the $f \to z$ update for the MAP module $\lambda f$ leads to the messages:
\begin{align}
    \label{tv_vamp_rz}
    &r_z^f = \eta_{\frac{\lambda}{a_{z \to f}}} \left(
        \frac{b_{z \to f}}{a_{z \to f}}
    \right), \quad
    v_z^f = \frac{1}{a_{z \to f}} 
    \left\langle 
        \nabla \eta_{\frac{\lambda}{a_{z \to f}}} 
        \left(
            \frac{b_{z \to f}}{a_{z \to f}}
        \right) 
    \right\rangle , \\
    \label{tv_vamp_a_zk}
    &a_{z \to K} = a_{f \to z} = \frac{1}{v_z^f} - a_{z \to f}, \quad
    b_{z \to K} = b_{f \to z} = \frac{r_z^f}{v_z^f} - b_{z \to f} .
\end{align}

In \citep{Manoel2018} the following quantities at iteration $t$ are denoted by:
\begin{equation}
a_{z \to K} = \rho^t, \quad 
b_{z \to K} = u^t, \quad 
r_x^K = x^t, \quad 
v_z^K = \sigma_x^t, \quad
r_z^f = z^t, \quad v_z^f = \sigma_z^t
\end{equation}
Then Eqs~\eqref{tv_vamp_rx}, \eqref{tv_vamp_rz} and \eqref{tv_vamp_a_zk} are exactly equivalent to the Eqs~(24), (25) and (26) of \citep{Manoel2018} defining the TV-VAMP algorithm.

As discussed in greater detail by \cite{Manoel2018}, the TV-VAMP algorithm is closely related to proximal methods: it can be viewed as the Peaceman-Rachford splitting where the step-size $\rho^t$ is set adaptively. Then Algorithm~\ref{algo:gaussian_EP} offers a  generalization to any optimization problem for which the factor graph of penalty/constraints is tree-structured (Section~\ref{sec:map_estimation}). For instance, while the TV-VAMP can only solve linear regression with a TV penalty,
Algorithm~\ref{algo:gaussian_EP} can be easily applied to a classification setting: one just needs to replace the Gaussian likelihood $\Delta$ by the appropriate likelihood.
Also Algorithm~\ref{algo:gaussian_EP} offers more flexibility in designing the approximate inference scheme: for example one can choose isotropic or diagonal Gaussian belief for $x$ to  alleviate the computational burden of inverting a matrix in Eq.~\eqref{tv_vamp_rx}.

\subsection{EP, EC, AdaTAP and Message-Passing}
There is a long history behind the methods used in this section and the literature on statistical physics.
In particular, broadening the class of matrices amenable to mean-field treatments was the motivation behind a decades long
series of works. 

\cite{parisi1995mean} were among the pioneers in this direction by deriving mean-field equations for orthogonal matrices. The adaTAP approach of \cite{csato2002tap}, and their reinterpretation as a  particular case of the Expectation Propagation algorithm \citep{Minka_EP_2001} allowed for a generic reinterpretation of these ideas as an approximation of the log partition named
Expectation Consistency (EC) \citep{Heskes2005,Opper2005JMLR,Opper2005NIPS}.  Many works then applied these ideas to problems such as the perceptron  \citep{shinzato2008perceptron,shinzato2008learning,kabashima2003cdma}. 

All these ideas were behind the recent renewal of interest of message-passing algorithms with generic rotationally invariant matrices \citep{Schniter2016,ma2017orthogonal,Cakmak2014,Cakmak2016}. In a recent work, \cite{maillard2019high} showed the consistency and the equivalence of these approaches.

\section{State evolution and free entropy\label{sec:SE}}

In this section we present an heuristic derivation of the free entropy and state evolution formalisms for the tree-structured models considered in Section~\ref{sec:EP_settings}. 
There is now a very vast literature on the state evolution and free entropy formalisms applied to machine learning models \citep{Zdeborova2016} so an exhaustive review is beyond the scope of this manuscript, however see Examples~\ref{example:glm_condition} and \ref{example:low_rank_condition} for some representative prior works. Our primary goal in this section is to tie these results together in an unifying framework, extend them to tree-structured factor graphs and justify the modularization of the free entropy and state evolution as done in the \tramp package. 

We first heuristically derive the so-called \emph{replica free entropy} \citep{mezard2009information} using weak consistency \citep{Heskes2005} on the overlaps. 
We expect our formulas to be valid only when the overlaps are the relevant order parameters needed to describe the ensemble average.
The replica symmetric solution is exposed in Section~\ref{sec:RS} and we present more specifically the Bayes-optimal setting in Section~\ref{sec:bayes_optimal}. 
The solution is easily interpreted as local ensemble averages defined for each factor and variable and allows us to conjecture the state evolution of the EP Algorithm~\ref{algo:gaussian_EP}.
This effective ensemble average is at the heart of the modularization of the free entropy and state evolution formalisms in the \tramp package.
We  express the free entropy potentials using information theoretic quantities in Section~\ref{sec:information_theory} and recover \citep{Reeves2017} formalism as a special case. Finally we briefly list the state evolution modules currently implemented in the \tramp package. 
We emphasize that the derivation is non-rigorous and largely conjectural, however we recover many replica free entropies and state evolutions previously derived for specific models, that are conjectured to be exact or even rigorously proven in some cases (Examples~\ref{example:glm_condition} and \ref{example:low_rank_condition}).

\begin{example}[Multi-layer and tree network of GLMs]
\label{example:glm_condition}
A very general setting is the tree network of GLMs proposed by \cite{Reeves2017}, which includes the
GLM and multi-layer network as special cases. A key assumption in such models is that the weight matrices in linear channels are drawn from an orthogonally invariant ensemble. The state evolution is rigorously proven in the multi-layer case \citep{Fletcher2018} for the corresponding ML-VAMP algorithm.
The replica free entropy for the multi-layer case is derived in \citep{Gabrie2018}.
 When the entries of the weight matrices are iid Gaussian (a special case of an orthogonally-invariant ensemble) the replica free entropy was further be shown to be rigorous in the compressed sensing \citep{Reeves2016} and GLM \citep{Barbier2019} cases.
\end{example}

\begin{example}[Low rank matrix factorization]
\label{example:low_rank_condition}
The replica free entropy for the low-rank matrix factorization problem (Example~\ref{example:low_rank}) and the state evolution of the LowRAMP algorithm are derived in \citep{Lesieur2017}. The replica free entropy was further shown to be rigorous in \citep{Miolane2017, Lelarge2019}. Stacking a multi-layer GLM with a low rank factorization model was considered in \citep{Aubin2020}.
\end{example}

\subsection{Model settings}

\paragraph{Teacher-student scenario} 
We will consider a generic teacher-student scenario where the teacher generates the signals $\mathbf{x}^{(0)}$ and measurements $\mathbf{y}$. The student is only given the measurements $\mathbf{y}$ and must infer the signals. The teacher generative model is a tree-structured factor graph:
\begin{equation} 
\label{teacher}
    p^{(0)}(\mathbf{x}^{(0)} , \mathbf{y}) = \frac{1}{Z_N^{(0)}} \prod_{k \in F} f_k^{(0)}(x_k^{(0)} ; y_k),
    \quad Z_N^{(0)} = \int d\mathbf{x}^{(0)} d\mathbf{y} \prod_{k \in F} f_k^{(0)}(x_k^{(0)} ; y_k),
\end{equation}
where $\mathbf{x}^{(0)} = (x_i^{(0)})_{i \in V}$ are the (ground truth) signals and $\mathbf{y} = (y_j)_{j \in O}$ the measurements, see Section~\ref{sec:EP_settings} for more information on the factor graph notation. 
For the student, we will assume the same tree factorization as the teacher, however the student factors $f_k(x_k; y_k)$ can be mismatched. The student generative model is given by Eq.~\eqref{factor_graph_unobserved} and the student posterior by Eq.~\eqref{factor_graph_conditional}.
The student posterior mean, variance and second moment are given by:
\begin{equation}
 r_i(\mathbf{y}) = \EE_{p(\mathbf{x} \mid \mathbf{y})} x_i, \quad  
 v_i(\mathbf{y}) = \langle \mathrm{Var}_{p(\mathbf{x} \mid \mathbf{y})} x_i \rangle, \quad
 \tau_i(\mathbf{y}) = \EE_{p(\mathbf{x} \mid \mathbf{y})} \frac{\Vert x_i \Vert^2 }{N_i} 
 = q_i(\mathbf{y}) + v_i(\mathbf{y}),
 \end{equation}
where $q_i(\mathbf{y}) = \frac{ \Vert r_i(\mathbf{y})) \Vert^2}{N_i} $ is the self-overlap.
The \emph{mismatched setting} corresponds to the case where at least one of the student factors is mismatched. On the opposite, the \emph{Bayes-optimal setting} refers to the case where all the student factors match the teacher factors $f_k(x_k; y_k)  = f_k^{(0)}(x_k; y_k)$, consequently the student and teacher generative models are identical $p(\mathbf{x}, \mathbf{y}) = p^{(0)}(\mathbf{x}, \mathbf{y})$ and the student posterior Eq.~\eqref{factor_graph_conditional} is indeed Bayes-optimal, in particular the posterior mean $r_i(\mathbf{y})$ is the minimal mean-squared-error (MMSE) estimator and the posterior variance $v_i(\mathbf{y})$ the MMSE.

\paragraph{High-dimensional limit} 
We will consider the high-dimensional limit $N \to \infty$  where each signal $x_i \in \mathbb{R}^{N_i}$ is itself a high-dimensional object with scaling $\alpha_i = \frac{N_i}{N} = O(1)$. We will denote by $N_k$ the dimension of the factor $f_k$ (for instance we can choose the dimension of its inputs signals by convention) and $\alpha_k = \frac{N_k}{N} = O(1)$ the corresponding scaling. Finally we will denote $\alpha_i^k = \frac{N_i}{N_k} = O(1)$ the scaling of the variable $x_i$ wrt the factor $f_k$. In the large $N$ limit we expect the log-partition to self-average:
\begin{equation}
    A_N(\mathbf{y}) = \frac{1}{N} \ln Z_N(\mathbf{y}) \simeq \bar{A} = \lim_{N \to \infty} \EE_{ p^{(0)}(\mathbf{y}) } A_N(\mathbf{y})
\end{equation}
where the log-partition $A_N(\mathbf{y})$ is scaled by $N$ in order to be $O(1)$. The ensemble average $\bar{A}$ is called the \emph{free entropy} and gives the cross-entropy up to a constant:
\begin{equation}
    \label{cross_entropy}
    - \bar{A} =  \lim_{N \to \infty} \frac{1}{N} H[p^{(0)}(\mathbf{y}), p(\mathbf{y})] - A_N
\end{equation}
where $A_N = \frac{1}{N} \ln Z_N$ is the scaled log-partition associated to the student generative model Eq.~\eqref{factor_graph_unobserved} and $H[p , q] = - \EE_p \ln q$ denotes the differential cross-entropy. In particular when the model is a Bayesian network (Remark~\ref{remark:bayes_net}) $Z_N = 1$ and $A_N=0$ so the free entropy directly gives the cross-entropy. The goal of the free entropy formalism is to provide an analytical expression for $\bar{A}$ and describe the limiting ensemble average.

\paragraph{Replica free entropy}
The replica trick \citep{mezard1987spin} can be viewed as an heuristic method to compute
\begin{equation}
\label{SCGF}
A(n) 
= \lim_{N \to \infty} \frac{1}{N} \ln \EE_{p^{(0)}(\mathbf{y})}  Z_N(\mathbf{y})^n 
=  \lim_{N \to \infty} \frac{1}{N} \ln \EE_{p^{(0)}(\mathbf{y})}  e^{N n A_N(\mathbf{y})} 
\end{equation}
which is interpreted in large deviation theory \citep{Touchette2009} as the scaled cumulant generating function (SCGF) of the log-partition $A_N(\mathbf{y}) = \frac{1}{N} \ln Z_N(\mathbf{y})$. If the SCGF is well defined we can get the ensemble average log-partition as:
\begin{equation}
\label{SCGF_derivative}
\bar{A} = \left. \frac{d}{dn} A(n) \right \vert_{n=0}
\end{equation}
We can formally decompose Eq.~\eqref{SCGF} at finite $N$, before talking the $N \to \infty$ limit:
\begin{equation}
\label{A_decomposition}
A(n) = A_N^{(n)} - A_N^{(0)}, 
\quad A_N^{(n)} = \frac{1}{N} \ln Z_N^{(n)}, 
\quad A_N^{(0)} = \frac{1}{N} \ln Z_N^{(0)},
\end{equation}
where $Z_N^{(0)}$ is the partition function introduced in Eq.~\eqref{teacher} and $Z_N^{(n)}$ the partition function of the replicated system:
\begin{equation}
    \label{replicated_system}
    p^{(n)}(\{ \mathbf{x}^{(a)} \}_{a=0}^n, \mathbf{y}) = \frac{1}{Z_N^{(n)}} \prod_{k \in F} \left\{ 
    f_k^{(0)}(x_k^{(0)}; y_f) \prod_{a=1}^n f_k(x_k^{(a)}; y_k) 
    \right\}
\end{equation}
where $\mathbf{x}^{(a)}$ for $a=1\cdots n$ denote the $n$ replicas and $\mathbf{x}^{(0)}$ the ground truth. The replica free entropy is obtained by computing $Z_N^{(n)}$ as if $n\in \mathbb{N}$ in Eq.~\eqref{replicated_system}, but then letting $n \to 0$ in Eq.~\eqref{SCGF_derivative} as if $n$ was real \citep{mezard2009information}. The replica method is non-rigorous, however it has been successfully applied for decades and given numerous given exact results, some of which were later confirmed by rigorous methods.  There is therefore a very high level of trust in the replica method by the statistical physics community.

\paragraph{Overlaps} In statistical physics literature, the system is said to have a well defined thermodynamic limit $N \to \infty$ if the limiting ensemble average is fully characterized by a few scalar parameters (called order parameters). Here we will restrict our analysis to systems where these order parameters are the overlaps:
\begin{equation}
\label{overlaps}
 \phi(\mathbf{x}) = \left( \frac{x_i^{(a)}\cdot x_i^{(b)}}{N_i}\right)_{i \in V, 0\leq a \leq b \leq n}
\end{equation}
which due to the definition of the replicated system Eq.~\eqref{replicated_system} correspond to:
\begin{align}
    &\tau_i^{(0)} = \EE_{p^{(n)}} \frac{\Vert x_i^{(0)} \Vert^2}{N_i} 
    = \EE_{ p^{(0)}( \mathbf{x}^{(0)}) } \frac{\Vert x_i^{(0)} \Vert^2}{N_i} \\
    &\tau_i = \EE_{p^{(n)}} \frac{ \Vert x_i^{(a)} \Vert^2}{N_i} 
    = \EE_{ p^{(0)}(\mathbf{y} ) } \tau_i(\mathbf{y})
    \quad &\text{for all }1\leq a \leq n \\
    &m_i = \EE_{p^{(n)}} \frac{x_i^{(a)}\cdot x_i^{(0)}}{N_i} 
    = \EE_{ p^{(0)}( \mathbf{x}^{(0)}, \mathbf{y} ) } \frac{r_i(\mathbf{y}) \cdot x_i^{(0)}}{N_i} 
    \quad &\text{for all }1\leq a \leq n \\
    &q_i^{(ab)} = \EE_{p^{(n)}} \frac{x_i^{(a)} \cdot x_i^{(b)}}{N_i} 
    \quad &\text{for all }1\leq a \neq b \leq n
\end{align}
In the ensemble average, $m_i$ denotes the overlap with the ground truth, $\tau_i^{(0)}$ the teacher prior second moment, and $\tau_i$ is the student posterior second moment. 
It is difficult to tell under which conditions (on the high-dimensional factors and their arrangement in a tree graph) the ensemble average will be fully characterized by the overlaps, and we hope that future theoretical work could clarify this point.
Examples~\ref{example:glm_condition} and \ref{example:low_rank_condition} show however a few representative models and conditions. For instance in GLMs and network of GLMs
the weight matrices in the linear channels must come from an orthogonally invariant ensemble. The core of our heuristic derivation of the replica free entropy is to assume weak consistency on the overlaps Eq.~\eqref{overlaps}. We note that this weak consistency derivation could be extended by adapting the sufficient statistics Eq.~\eqref{overlaps} to include other order parameters, if those turn out to be relevant to describe the ensemble average.

\paragraph{Replica symmetry}
The system  is said to be \emph{replica symmetric} if the overlap $q_i^{(ab)}$ between two replicas 
concentrates to a single value $q_i$ which is then equal to the self-overlap
\begin{equation}
q_i = \EE_{p^{(0)}(\mathbf{y}) } q_i(\mathbf{y}) .
\end{equation}
In the Bayes-optimal setting, the system will always be replica symmetric \citep{nishimori2001, mezard2009information}.
In the mismatched setting, we expect the system to sometimes exhibit \emph{replica symmetry breaking}, 
where the overlap $q_i^{(ab)}$ between two replicas 
converges instead to a discrete distribution:
\begin{equation}
 P(q_i) = \sum_{r=0}^R \pi_r \delta(q_i - q_i^r), \quad q_i^0 \leq \ldots \leq q_i^R .
\end{equation}
This situation is called R-level symmetry breaking and the cumulative $s_r = \sum_{r'=0}^{r} \pi_{r'}$ which satisfy $0 \leq s_0 \leq \ldots \leq s_R = 1$ are called the Parisi parameters \citep{mezard2009information}. The system can also undergo full replica symmetry breaking
where the overlap distribution has a continuous part. In this manuscript, we focus on the replica symmetric solution, the replica symmetry breaking solution is deferred to a forthcoming publication.

\subsection{Teacher prior second moments}
\label{sec:teacher_second_moments}

The replica symmetric solution requires the teacher prior second moments $\tau_V = (\tau_i^{(0)})_{i \in V}$ which we explain how to compute in this section. 
The weak consistency approximation of the log-partition $A_N^{(0)} = \frac{1}{N} Z_N^{(0)}$ using the sufficient statistics
\begin{equation}
    \label{second_moments}
    \phi_i(x_i^{(0)}) = -\tfrac{1}{2} \Vert x_i^{(0)}\Vert^2 \quad \text{for all }i \in V
\end{equation}
is summarized by Proposition~\ref{prop:teacher_second_moments}. It yields an approximation for both the log-partition $A_N^{(0)}$ and second-moments $\tau_V^{(0)}$ that we assume to be asymptotically exact in the large $N$ limit. The full derivation follows exactly the same steps as Section~\ref{sec:EP} for the student posterior Eq.~\eqref{factor_graph_conditional} but here applied to the teacher prior
\begin{equation}
\label{teacher_prior}
    p^{(0)}(\mathbf{x}^{(0)}) = \frac{1}{Z_N^{(0)}} \prod_{k \in F} f_k^{(0)}(x_k^{(0)})
    \quad \text{with} \quad 
    f_k^{(0)}(x_k^{(0)}) = \int dy_k \, f_k^{(0)}(x_k^{(0)};y_k)
\end{equation}
obtained by marginalizing Eq.~\eqref{teacher} over $\mathbf{y}$.

\begin{proposition}[Weak consistency derivation of $A_N^{(0)}$]
\label{prop:teacher_second_moments}
Solving the relaxed Bethe variational problem, using the sufficient statistics Eq.~\eqref{second_moments}, leads to:
\begin{equation}
- A_N^{(0)} 
=
\min_{\tau_V^{(0)}} G^{(0)}[\tau_V^{(0)}]
   = \min \extr_{\hat{\tau}_E^{(0)}} - A^{(0)}[\hat{\tau}_E^{(0)}].
\end{equation} 
where the minimizer corresponds to the teacher prior second moments $\tau_V^{(0)} = (\tau_i^{(0)})_{i \in V}$ and 
$\hat{\tau}_E^{(0)} = (\hat{\tau}_{i \to k}^{(0)}, \hat{\tau}_{k \to i}^{(0)})_{(i,k) \in E}$ denotes the dual natural parameter messages. The potentials satisfy the tree decomposition:
\begin{align}
    &G^{(0)}[\tau_V^{(0)}] = 
    \sum_{k \in F} \alpha_k G_k^{(0)}[\tau_k^{(0)}] 
    + \sum_{i \in V} \alpha_i (1-n_i) G_i^{(0)}[\tau_i^{(0)}] 
    \quad \text{with} \quad
     \tau_k^{(0)} = (\tau_i^{(0)})_{i \in \partial k} \\
     \notag
    &A^{(0)}[\hat{\tau}_E^{(0)}] = 
    \sum_{k \in F} \alpha_k A_k^{(0)}[\hat{\tau}_k^{(0)}] 
    - \sum_{(i,k) \in E} \alpha_i A_i^{(0)}[\hat{\tau}_i^{k(0)}]
    + \sum_{i \in V} \alpha_i A_i^{(0)}[\hat{\tau}_i^{(0)}] \\
&\text{with} 
\quad
\hat{\tau}_k^{(0)} = (\hat{\tau}_{i \to k}^{(0)})_{i \in \partial k},
\quad
\hat{\tau}_i^{k(0)} =  \hat{\tau}_{i \to k}^{(0)} + \hat{\tau}_{k \to i}^{(0)},
\quad
\hat{\tau}_i^{(0)} =  \sum_{k \in \partial i}  \hat{\tau}_{k \to i}^{(0)}.
\end{align}
The scaled factor and variable log-partitions are given by:
\begin{align}
    \label{teacher_factor_log_partition}
    &A_k^{(0)}[\hat{\tau}_k^{(0)}] = \frac{1}{N_k} \ln \int dy_k dx_k^{(0)} \, f_k^{(0)}(x_k^{(0)} ; y_k) 
    \, e^{
    - \frac{1}{2} \hat{\tau}_k^{(0)}   \Vert x_k^{(0)} \Vert^2
    } \\
    \label{teacher_variable_log_partition}
    &A_i^{(0)}[\hat{\tau}_i^{(0)}] = \frac{1}{N_i} \ln \int dx_i^{(0)}
    \, e^{
    - \frac{1}{2} \hat{\tau}_i^{(0)}   \Vert x_i^{(0)} \Vert^2
    } = \frac{1}{2} \ln \frac{2\pi}{\hat{\tau}_i^{(0)}}
\end{align}
$A_k^{(0)} = \frac{1}{N_k} \ln Z_k^{(0)}$ is the log-partition of the exponential family distribution that approximates the teacher factor marginal $p^{(0)}(x_k^{(0)}, y_k)$
\begin{equation}
    \label{teacher_factor_marginal}
    p_k^{(0)}(x_k^{(0)}, y_k \mid \hat{\tau}_k^{(0)}) = \frac{1}{Z_k^{(0)}[\hat{\tau}_k^{(0)} ]} f_k^{(0)}(x_k^{(0)}; y_k) \, e^{-\frac{1}{2} \hat{\tau}_k^{(0)} \Vert x_k^{(0)} \Vert^2}  .
\end{equation}
Similarly $A_i^{(0)} = \frac{1}{N_i} \ln Z_i^{(0)}$ is the log-partition of the zero-mean Normal that approximates the teacher variable marginal $p^{(0)}(x_i^{(0)})$
\begin{equation}
\label{teacher_variable_marginal}
    p_i^{(0)}(x_i^{(0)} \mid \hat{\tau}_i^{(0)}) = \frac{1}{Z_i^{(0)}[\hat{\tau}_i^{(0)}]} e^{-\frac{1}{2} \hat{\tau}_i^{(0)} \Vert x_i^{(0)} \Vert^2} = \mathcal{N}(x_i^{(0)} \mid 0,  \, \tau_i^{(0)})
     \quad \text{with} \quad \tau_i^{(0)} = \frac{1}{\hat{\tau}_i^{(0)}} .
\end{equation}
The gradients of the factor and variable log-partitions give the dual mapping to the second moments:
\begin{align}
    &\tau_i^{k(0)}[\hat{\tau}_k^{(0)}] = 
    \EE_{p_k^{(0)}(x_k^{(0)}, y_k \mid \hat{\tau}_k^{(0)})} \frac{\Vert x_i^{(0)} \Vert^2}{N_i},
    \quad
    &-\frac{1}{2} \alpha_i^k \tau_i^{k(0)} =
    \partial_{\hat{\tau}_{i \to k}^{(0)}} A_k^{(0)}, \\
    &\tau_i^{(0)}[\hat{\tau}_i^{(0)}] = 
    \EE_{p_i^{(0)}(x_i^{(0)} \mid \hat{\tau}_i^{(0)})} \frac{\Vert x_i^{(0)} \Vert^2}{N_i} = \frac{1}{\hat{\tau}_i^{(0)}} , 
    \quad
    &- \frac{1}{2} \tau_i^{(0)} = 
    \partial_{\hat{\tau}_i^{(0)}} A_i^{(0)} .
\end{align}
$G_k^{(0)}$ and $G_i^{(0)}$ are the corresponding Legendre transforms.
Any stationary point of the potentials (not necessarily the global optima) is a fixed point:
\begin{equation}
    \tau_i^{k(0)} = \tau_i^{(0)}, \quad \hat{\tau}_i^{(0)} = \sum_{k \in \partial i} \hat{\tau}_{k \to i}^{(0)}.
\end{equation}
\end{proposition}

\paragraph{Bayesian network teacher}
When the teacher factor graph is a Bayesian network (Remark~\ref{remark:bayes_net}) the fixed point in Proposition~\ref{prop:teacher_second_moments} is particularly simple.
If $x_i$ is an input signal of the factor $f_k$:
\begin{equation}
   \hat{\tau}_{i \to k}^{(0)} = \hat{\tau}_i^{(0)} = \frac{1}{\tau_i^{(0)}} , \qquad
   \hat{\tau}_{k \to i}^{(0)} = 0 .
\end{equation}
If $x_i$ is an output signal of the factor $f_k$:
\begin{equation}
    \hat{\tau}_{k \to i}^{(0)} = \hat{\tau}_i^{(0)} = \frac{1}{\tau_i^{(0)}} , \qquad
    \hat{\tau}_{i \to k}^{(0)} = 0 .
\end{equation}
In other words the forward messages are equal to the precisions while the backward messages are null.
The factor log-partition is equal to:
\begin{equation}
    A_k^{(0)}[\hat{\tau}_f^{(0)}] = \sum_{i \in (\partial k)^{-}} \alpha_i^k A_i^{(0)}[\hat{\tau}_i^{(0)}] 
\end{equation}
where $(\partial k)^{-}$ denotes the input signals nodes of the factor node $k$.
In particular Proposition~\ref{prop:teacher_second_moments} gives consistently:
\begin{equation}
    A_N^{(0)} = \sum_{k \in F} \alpha_k A_k^{(0)}[\hat{\tau}_k^{(0)}] + \sum_{i \in V} \alpha_i (1-n_i) A_i^{(0)}[\hat{\tau}_i^{(0)}] = 0
\end{equation}
as it should, because $Z_N^{(0)} = 1$ and $A_N^{(0)} = 0$ for a Bayesian network.

\paragraph{Computing the fixed point}
In practice one can use the \tramp Algorithm~\ref{algo:teacher_second_moment} to find the fixed point, which will compute the second-moments $\tau_V^{(0)}$ as well as the messages $\hat{\tau}_E^{(0)}$. Note that in the usual state evolution algorithms, for example in the multi-layer GLM case \citep{Gabrie2018, Fletcher2018}, a first step  is always to compute the teacher prior second moments, often invoking the central limit theorem and using approximate isotropic Gaussian distributions along the way. These routines are exactly equivalent to Algorithm~\ref{algo:teacher_second_moment}, which indeed yields the fixed point in a single forward pass when the teacher factor graph is a Bayesian network. By contrast in factor graphs that are not Bayesian network the \tramp Algorithm~\ref{algo:teacher_second_moment} will be useful to find the fixed point which is no longer trivial and to compute the normalization constant $A_N^{(0)}$.

\vspace{1em}
\begin{algorithm}[H]
\label{algo:teacher_second_moment}
\DontPrintSemicolon
\Repeat{convergence}{
  \ForEach(\tcp*[f]{ forward and backward pass}){edge $e \in E_+ \cup E_-$}{
    \If{$e = k \to i$}{
      $ \hat{\tau}_i^{k(0)}  = 1 / \tau_i^{k(0)}[\hat{\tau}_k^{(0)}]$
      \\
      $\hat{\tau}_{k \to i}^{(0)\text{new}} = \hat{\tau}_i^{k(0)} - \hat{\tau}_{i \to k}^{(0)}$
      \tcp*[f]{message $f_k \to x_i$ }
    }
    \If{$e = i \to k$}{
      $\hat{\tau}_i^{(0)} = \sum_{k' \in \partial i} \hat{\tau}_{k' \to i}^{(0)}$
      \\
      $\hat{\tau}_{i \to k}^{(0)\text{new}} = \hat{\tau}_i^{(0)} - \hat{\tau}_{k \to i}^{(0)}$
      \tcp*[f]{message $x_i \to f_k$ }
    }
  }
}
\caption{\tramp algorithm for the teacher prior second moments}
\end{algorithm}

\subsection{Replica symmetric free entropy \label{sec:RS}}
The replica symmetric free entropy is derived by assuming weak consistency on the overlaps Eq.~\eqref{overlaps}. The full derivation is presented in Appendix~\ref{app:RS_free_entropy} and summarized in Proposition~\ref{prop:RS_free_entropy}. We assume that the teacher prior second moments $\tau_V^{(0)}$ and dual messages $\hat{\tau}_E^{(0)}$ are known thanks to Proposition~\ref{prop:teacher_second_moments} and should be now considered as fixed parameters.

\begin{proposition}[Replica symmetric $\bar{A}$] 
\label{prop:RS_free_entropy}
The replica symmetric (RS) $\bar{A}$ is given by:
\begin{equation}
- \bar{A} = \min_{m_V,q_V,\tau_V} \bar{A}^*[m_V,q_V,\tau_V]
   = \min \extr_{\hat{m}_E, \hat{q}_E, \hat{\tau}_E} - \bar{A}[\hat{m}_E,\hat{q}_E,\hat{\tau}_E].
\end{equation} 
where the minimizer corresponds to the overlaps $m_V = (m_i)_{i \in V}$ and $\hat{m}_E = (\hat{m}_{i \to k}, \hat{m}_{k \to i})_{(i,k) \in E}$ denotes the dual natural parameter messages (idem $q,\tau$).
The RS potentials satisfy the tree decomposition:
\begin{align}
\notag
    &\bar{A}^*[m_V,q_V,\tau_V] = \sum_{k \in F} \alpha_k \bar{A}_k^*[m_k,q_k,\tau_k] 
    + \sum_{i \in V} \alpha_i (1-n_i) \bar{A}_i^*[m_i, q_i, \tau_i] \\
    &\text{with} \quad
    m_k = (m_i)_{i \in \partial k} \quad (\text{idem }q,\tau), \\
     \notag
    &\bar{A}[\hat{m}_E, \hat{q}_E, \hat{\tau}_E] = 
    \sum_{k \in F} \alpha_k A_k[\hat{m}_k, \hat{q}_k, \hat{\tau}_k] 
    - \sum_{(i,k) \in E} \alpha_i A_i[\hat{m}_i^k, \hat{q}_i^k, \hat{\tau}_i^k]
    + \sum_{i \in V} \alpha_i A_i[\hat{m}_i,\hat{q}_i, \hat{\tau}_i] \\
&\text{with} 
\quad
\hat{m}_k = ( \hat{m}_{i \to k})_{i \in \partial k},
\quad
\hat{m}_i^k =  \hat{m}_{i \to k} + \hat{m}_{k \to i},
\quad
\hat{m}_i =  \sum_{k \in \partial i}  \hat{m}_{k \to i} \quad (\text{idem }\hat{q},\hat{\tau}).
\end{align}
The factor and variable RS potentials are given by:
\begin{align}
   \label{RS_potential_f}
    \bar{A}_k[\hat{m}_k, \hat{q}_k, \hat{\tau}_k] 
    &= \lim_{N_k \to \infty}\EE_{p_k^{(0)}(x_k^{(0)}, y_k, b_k)} A_k[a_k,b_k ; y_k] \\
    \label{RS_potential_x}
    \bar{A}_i[\hat{m}_i, \hat{q}_i, \hat{\tau}_i] 
    &= \lim_{N_i \to \infty}\EE_{p_i^{(0)}(x_i^{(0)}, b_i)} A_i[a_i,b_i] 
\end{align}
where $A_k[a_k,b_k ; y_k]$ and $A_i[a_i,b_i]$ are the scaled student EP log-partitions with isotropic Gaussian beliefs (Example~\ref{example:duality_isotropic}):
\begin{align}
    A_k[a_k,b_k;y_k] &= \frac{1}{N_k}  \ln \int dx_k \, f(x_k ; y_k) \, e^{-\frac{1}{2} a_k \Vert x_k \Vert^2 + b_k^\intercal x_k} \\
    A_i[a_i,b_i] &= \frac{1}{N_i} \ln \int dx_i \, e^{-\frac{1}{2} a_i \Vert x_i \Vert^2 + b_i^\intercal x_i} 
\end{align}
and the factor and variable ensemble averages are taken with:
\begin{align}
    \label{RS_ensemble_avg_f}
    p_k^{(0)}(x_k^{(0)}, y_k, b_k) &=
    \mathcal{N}(b_k \mid \hat{m}_k x_k^{(0)}, \hat{q}_k) \;
    p_k^{(0)}(x_k^{(0)}, y_k \mid \hat{\tau}_k^{(0)})
    \quad &\text{and} \quad
    a_k &= \hat{\tau}_k + \hat{q}_k, \\
    \label{RS_ensemble_avg_x}
    p_i^{(0)}(x_i^{(0)}, b_i) &=
    \mathcal{N}(b_i \mid \hat{m}_i x_i^{(0)}, \hat{q}_i) \;
    p_i^{(0)}(x_i^{(0)} \mid \hat{\tau}_i^{(0)})
    \quad &\text{and} \quad
     a_i &= \hat{\tau}_i + \hat{q}_i,
\end{align}
where $p_k^{(0)}(x_k^{(0)}, y_k \mid \hat{\tau}_k^{(0)})$ and $p_i^{(0)}(x_i^{(0)} \mid \hat{\tau}_i^{(0)})$
are the approximate teacher marginals defined in Eqs~\eqref{teacher_factor_marginal} and \eqref{teacher_variable_marginal}.
The gradient of the factor RS potential give the dual mapping to the overlaps:
\begin{align}
    \label{RS_overlap_f}
    &m_i^k[\hat{m}_k,\hat{q}_k,\hat{\tau}_k] = \EE_{p_k^{(0)}(x_k^{(0)}, y_k, b_k)} \frac{r_i^k[a_k, b_k ; y_k] \cdot x_i^{(0)}}{N_i},
    &\alpha_i^k m_i^k = 
    \partial_{\hat{m}_{i \to k}} \bar{A}_k, \\
    &q_i^k[\hat{m}_k,\hat{q}_k,\hat{\tau}_k] = \EE_{p_k^{(0)}(x_k^{(0)}, y_k, b_k)} \frac{ \Vert r_i^k[a_k, b_k ; y_k] \Vert^2 }{N_i}, 
    &-\tfrac{1}{2} \alpha_i^k q_i^k = 
    \partial_{\hat{q}_{i \to k}} \bar{A}_k, \\
    \label{RS_tau_f}
    &\tau_i^k[\hat{m}_k,\hat{q}_k,\hat{\tau}_k] = 
    \EE_{p_k^{(0)}(x_k^{(0)}, y_k, b_k)} \frac{ \Vert r_i^k[a_k, b_k ; y_k] \Vert^2 }{N_i} + v_i^k[a_k, b_k ; y_k] , 
    &-\tfrac{1}{2} \alpha_i^k \tau_i^k =
    \partial_{\hat{\tau}_{i \to k}} \bar{A}_k, 
\end{align}
where $r_i^k[a_k, b_k ; y_k]$ and $v_i^k[a_k , b_k ; y_k]$ are the posterior mean and isotropic variance Eqs~\eqref{factor_mean_isotropic} and \eqref{factor_variance_isotropic} as estimated by the student EP factor marginal Eq.~\eqref{factor_marginal_isotropic}.
The gradient of the variable RS potential give the dual mapping to the overlaps:
\begin{align}
    &m_i[\hat{m}_i,\hat{q}_i,\hat{\tau}_i] = \EE_{p_i^{(0)}(x_i^{(0)}, b_i)} \frac{r_i[a_i, b_i] \cdot x_i^{(0)}}{N_i}, \quad
    &m_i = 
    \partial_{\hat{m}_i} \bar{A}_i, \\
    &q_i[\hat{m}_i,\hat{q}_i,\hat{\tau}_i] = \EE_{p_i^{(0)}(x_i^{(0)}, b_i)} \frac{ \Vert r_i[a_i, b_i] \Vert^2 }{N_i}, \quad
    &-\tfrac{1}{2} q_i = 
    \partial_{\hat{q}_i} \bar{A}_i, \\
    &\tau_i[\hat{m}_i,\hat{q}_i,\hat{\tau}_i] = 
    \EE_{p_i^{(0)}(x_i^{(0)}, b_i)} \frac{ \Vert r_i[a_i, b_i] \Vert^2 }{N_i} + v_i[a_i, b_i]
    , \quad
    &-\tfrac{1}{2} \tau_i =
    \partial_{\hat{\tau}_i} \bar{A}_i,
\end{align}
where $r_i[a_i, b_i] = \frac{b_i}{a_i}$ and $v_i[a_i , b_i] = \frac{1}{a_i}$ are the posterior mean and isotropic variance as estimated by the student EP variable marginal Eq.~\eqref{variable_marginal_isotropic}.
$\bar{A}_k^*$ and $\bar{A}_i^*$ are the corresponding Legendre transforms. 
The ensemble average variances are given by:
\begin{align}
     &v_i^k[\hat{m}_k,\hat{q}_k,\hat{\tau}_k] = \EE_{p_k^{(0)}(x_k^{(0)}, y_k, b_k)} v_i^k[a_k, b_k ; y_k] 
     = \tau_i^k[\hat{m}_k,\hat{q}_k,\hat{\tau}_k] - q_i^k[\hat{m}_k,\hat{q}_k,\hat{\tau}_k], \\
     &v_i[\hat{m}_i,\hat{q}_i,\hat{\tau}_i] = \EE_{p_i^{(0)}(x_i^{(0)}, b_i)} v_i[a_i, b_i] 
     = \tau_i[\hat{m}_i,\hat{q}_i,\hat{\tau}_i] - q_i[\hat{m}_i,\hat{q}_i,\hat{\tau}_i] .
\end{align}
Any stationary point of the potentials (not necessarily the global optima) is a fixed point:
\begin{equation}
     m_i^k = m_i, \quad 
     \hat{m}_i = \sum_{k \in \partial i} \hat{m}_{k \to i}, \quad
     (\text{idem } q, \tau)
\end{equation}
\end{proposition}
 
\paragraph{RS potentials} 
The variable RS potential Eq.~\eqref{RS_potential_x} is explicitly given by:
 \begin{equation}
    \label{RS_potential_variable}
     \bar{A}_i[\hat{m}_i,\hat{q}_i,\hat{\tau}_i] = \frac{\hat{m}_i^2 \tau_i^{(0)} + \hat{q}_i}{2a_i} + \frac{1}{2} \ln \frac{2\pi}{a_i}
     \quad \text{with} \quad a_i = \hat{\tau}_i + \hat{q}_i, \\
 \end{equation}
 which yields the dual mapping:
\begin{equation}
     \label{RS_potential_mapping}
     m_i = \frac{\hat{m}_i \tau_i^{(0)}}{a_i}, \quad
     q_i = \frac{\hat{m}_i^2 \tau_i^{(0)} + \hat{q}_i}{a_i^2}, \quad
     \tau_i = q_i + v_i
    \quad \text{with} \quad  v_i = \frac{1}{a_i}.
 \end{equation}
Several factor RS potentials are given in Appendix~\ref{app:modules}.

\paragraph{Cross-entropy estimation and state evolution} 
%As usual in the free entropy formalism, there are two ways to interpret stationary points in Proposition~\ref{}, whether we are looking at a global vs local minimum. 
The ensemble average $\bar{A}$ which gives access to the cross-entropy through Eq.~\eqref{cross_entropy} is given by the global minimum according to Proposition~\ref{prop:RS_free_entropy}. The global minimizer gives access to the ensemble average overlaps $m_i, q_i, \tau_i$ (as well as $\mathrm{mse}_i = \tau_i^{(0)} -2m_i + q_i$ and $v_i = \tau_i - q_i$) for the student posterior Eq.~\eqref{factor_graph_conditional}. However the replica free entropy solution also appears as an ensemble average of the underlying EP Algorithm~\ref{algo:gaussian_EP} where the effective ensemble average is defined locally for each factor Eq.~\eqref{RS_ensemble_avg_f} and variable Eq.~\eqref{RS_ensemble_avg_x}. With this effective ensemble average interpretation in mind, we conjecture Algorithm~\ref{algo:SE_RS_mismatched} to give the state evolution of the EP Algorithm \ref{algo:gaussian_EP}.
Then the state evolution fixed point will give the overlaps $m_i,q_i,\tau_i$ (as well as $\mathrm{mse}_i = \tau_i^{(0)} -2m_i + q_i$ and $v_i = \tau_i - q_i$) corresponding to the student EP solution, which is in general only a local minimizer in Proposition~\ref{prop:RS_free_entropy}. 

\paragraph{Hard phase} When the SE fixed point happens to be the global minimizer, the EP algorithm is in a sense optimal as its solution achieves the same overlaps as the student posterior according to Proposition~\ref{prop:RS_free_entropy}. By contrast, when the SE fixed point fails to be a global minimizer, the EP algorithm is sub-optimal and is said to be in a computational \emph{hard phase}. Finally note that Algorithm~\ref{algo:SE_RS_mismatched} can be viewed more generally as an iterative routine to find stationary points of the replica free entropy potential. When initialized as in Algorithm~\ref{algo:SE_RS_mismatched} it leads to the SE fixed point, but if initialized in the right basin of attraction it leads to the global minimizer.

\paragraph{Prior work} We recover several results as particular cases. The state evolution for the ML-VAMP algorithm in the mismatched setting rigorously proven in \citep{Pandit2020} is equivalent to Algorithm~\ref{algo:SE_RS_mismatched} applied to the multi-layer GLM with orthogonally invariant weight matrices. 
%with matrix valued signals\footnote{The dimensions $N \times d$ of the signals are taken in the limit $N \to \infty$ with $d$ fixed. The overlaps $m,q,\tau$, as well as $\textrm{mse}$ and $v$ are then $d \times d$ matrices.}. 
We recover the replica symmetric free entropy in the mismatched setting derived for the GLM (with orthogonally invariant weight matrix) in \citep{kabashima2008} and the low rank factorization in \citep{Lesieur2017}.

\vspace{1em}
\begin{algorithm}[H]
\label{algo:SE_RS_mismatched}
\DontPrintSemicolon
\SetKw{Initialize}{initialize}
\Initialize{$\hat{m}_{i\to k},\hat{q}_{i\to k},\hat{\tau}_{i\to k},
             \hat{m}_{k\to i},\hat{q}_{k\to i},\hat{\tau}_{k\to i} = 0$}\\
\Repeat{convergence}{
  \ForEach(\tcp*[f]{ forward and backward pass}){edge $e \in E_+ \cup E_-$}{
    \If{$e = k \to i$}{
      $v_i^k =  v_i^k[\hat{m}_k, \hat{q}_k, \hat{\tau}_k], \quad 
      m_i^k = m_i^k[\hat{m}_k, \hat{q}_k, \hat{\tau}_k], \quad 
      q_i^k =  q_i^k[\hat{m}_k, \hat{q}_k, \hat{\tau}_k]$\\
      $a_i^k = \frac{1}{v_i^k}, \quad 
      \hat{m}_i^k = \frac{a_i^k m_i^k}{\tau_i^{(0)}}, \quad 
      \hat{q}_i^k = (a_i^k)^2 q_i^k - (\hat{m}_i^k)^2 \tau_i^{(0)} , \quad 
      \hat{\tau}_i^k = a_i^k - \hat{q}_i^k $ \\
      $\hat{m}_{k \to i}^\text{new}  = \hat{m}_i^k - \hat{m}_{i \to k}, \quad 
       \hat{q}_{k \to i}^\text{new}  = \hat{q}_i^k - \hat{q}_{i \to k}, \quad 
       \hat{\tau}_{k \to i}^\text{new}  = \hat{\tau}_i^k - \hat{\tau}_{i \to k}  $
    }
    \If{$e = i \to k$}{
      $\hat{m}_i = \sum_{k' \in \partial i} \hat{m}_{k' \to i}, \quad
      \hat{q}_i = \sum_{k' \in \partial i} \hat{q}_{k' \to i}, \quad
      \hat{\tau}_i = \sum_{k' \in \partial i} \hat{\tau}_{k' \to i}
      $\\
       $\hat{m}_{i \to k}^\text{new} = \hat{m}_i - \hat{m}_{k \to i},\quad
       \hat{q}_{i \to k}^\text{new} = \hat{q}_i - \hat{q}_{k \to i}, \quad
       \hat{\tau}_{i \to k}^\text{new} = \hat{\tau}_i - \hat{\tau}_{k \to i}$
    }
  }
}
\caption{\tramp State evolution (replica symmetric mismatched setting)}
\end{algorithm}

\subsection{Bayes-optimal setting \label{sec:bayes_optimal}}

In the Bayes-optimal setting, where all the student factors match the teacher factors $f_k(x_k,y_k) = f_k^{(0)}(x_k,y_k)$, the solution should be replica symmetric \citep{nishimori2001} and furthermore 
the ground truth $\mathbf{x}^{(0)}$ should behave as one the replicas $\mathbf{x}^{(a)}$ ($a=1\cdots n$) in Eq.~\eqref{replicated_system}. In particular:
\begin{equation}
    \tau_i = \tau_i^{(0)}, \quad m_i = q_i, \quad \mathrm{mse}_i = v_i = \tau_i^{(0)} - m_i .
\end{equation}
As the student and teacher generative models are identical 
$p(\mathbf{x}, \mathbf{y}) = p^{(0)}(\mathbf{x}, \mathbf{y}) $, 
the ensemble average $\bar{A}$ in Eq.~\eqref{cross_entropy} now gives access to the entropy
\begin{equation}
\label{entropy}
    - \bar{A}^{(0)} =  \lim_{N \to \infty} \frac{1}{N} H[p^{(0)}(\mathbf{y})] - A_N^{(0)}
\end{equation}
up to the constant $A_N^{(0)}$ that can be estimated through Proposition~\ref{prop:teacher_second_moments}.
With these simplifications, we get the following free entropy.

\begin{proposition}[Bayes-optimal setting $\bar{A}^{(0)}$] 
\label{prop:BO_free_entropy}
The Bayes-optimal (BO) setting $\bar{A}^{(0)}$ is given by:
\begin{equation}
- \bar{A}^{(0)}
=
\min_{m_V} \bar{A}^{(0)*}[m_V]
   = \min \extr_{\hat{m}_E} - \bar{A}^{(0)}[\hat{m}_E].
\end{equation} 
where the minimizer corresponds to the overlaps $m_V = (m_i)_{i \in V}$ and $\hat{m}_E = (\hat{m}_{i \to k}, \hat{m}_{k \to i})_{(i,k) \in E}$ denotes the dual natural parameter messages. The BO potentials satisfy the tree decomposition:
\begin{align}
    &\bar{A}^{(0)*}[m_V] = \sum_{k \in F} \alpha_k \bar{A}_k^{(0)*}[m_k] 
    + \sum_{i \in V} \alpha_i (1-n_i) \bar{A}_i^{(0)*}[m_i] 
    \qquad \text{with} \quad
    m_k = (m_i)_{i \in \partial k}, \\
    \notag
    &\bar{A}^{(0)}[\hat{m}_E] = 
    \sum_{k \in F} \alpha_k \bar{A}_k^{(0)}[\hat{m}_k] 
    - \sum_{(i,k) \in E} \alpha_i \bar{A}_i^{(0)}[\hat{m}_i^k]
    + \sum_{i \in V} \alpha_i \bar{A}_i^{(0)}[\hat{m}_i] \\
&\text{with} \quad
\hat{m}_k = (\hat{m}_{i \to k})_{i \in \partial k},
\quad
\hat{m}_i^k =  \hat{m}_{i \to k} + \hat{m}_{k \to i},
\quad
\hat{m}_i =  \sum_{k \in \partial i}  \hat{m}_{k \to i}.
\end{align}
The factor and variable BO potentials are given by:
\begin{align}
    \label{BO_potential_f}
    \bar{A}_k^{(0)}[\hat{m}_k] 
    &= \lim_{N_k \to \infty}\EE_{p_k^{(0)}(x_k^{(0)}, y_k, b_k)}  A_k^{(0)}[a_k, b_k ; y_k] \\
    \label{BO_potential_x}
    \bar{A}_i^{(0)}[\hat{m}_i]  
    &= \lim_{N_i \to \infty} \EE_{p_i^{(0)}(x_i^{(0)}, b_i)} A_i^{(0)}[a_i, b_i]
\end{align}
where $A_k^{(0)}[a_k,b_k ; y_k]$ and $A_i^{(0)}[a_i,b_i]$ are the scaled EP log-partitions with isotropic Gaussian beliefs (Example~\ref{example:duality_isotropic}):
\begin{align}
    A_k^{(0)}[a_k,b_k;y_k] &= \frac{1}{N_k} \ln \int dx_k \, f_k^{(0)}(x_k ; y_k) \, e^{-\frac{1}{2} a_k \Vert x_k \Vert^2 + b_k^\intercal x_k} \\
    A_i^{(0)}[a_i,b_i] &=  \frac{1}{N_i} \ln \int dx_i \, e^{-\frac{1}{2} a_i \Vert x_i \Vert^2 + b_i^\intercal x_i} 
\end{align}
and the factor and variable ensemble averages are taken with:
\begin{align}
    \label{BO_ensemble_avg_f}
    p_k^{(0)}(x_k^{(0)}, y_k, b_k) &=
    \mathcal{N}(b_k \mid \hat{m}_k x_k^{(0)}, \hat{m}_k) \;
    p_k^{(0)}(x_k^{(0)}, y_k \mid \hat{\tau}_k^{(0)})
    \quad &\text{and} \quad
    a_k = \hat{\tau}_k^{(0)} + \hat{m}_k,  \\
    \label{BO_ensemble_avg_x}
    p_i^{(0)}(x_i^{(0)}, b_i) &=
    \mathcal{N}(b_i \mid \hat{m}_i x_i^{(0)}, \hat{m}_i) \;
    p_i^{(0)}(x_i^{(0)} \mid \hat{\tau}_i^{(0)})
    \quad &\text{and} \quad
    a_i = \hat{\tau}_i^{(0)} + \hat{m}_i,  
\end{align} 
where $p_k^{(0)}(x_k^{(0)}, y_k \mid \hat{\tau}_k^{(0)})$ and $p_i^{(0)}(x_i^{(0)} \mid \hat{\tau}_i^{(0)})$
are the approximate teacher marginals defined in Eqs~\eqref{teacher_factor_marginal} and \eqref{teacher_variable_marginal}.
The gradient of the factor RS potential give the dual mapping to the overlap:
\begin{equation}
    \label{BO_overlap_f}
    m_i^k[\hat{m}_k] = \EE_{p_k^{(0)}(x_k^{(0)}, y_k, b_k)} \frac{r_i^k[a_k, b_k ; y_k] \cdot x_i^{(0)}}{N_i} 
    , \quad \tfrac{1}{2} \alpha_i^k m_i^k = \partial_{\hat{m}_{i \to k}} \bar{A}_k^{(0)},
\end{equation}
where $r_i^k[a_k, b_k ; y_k]$ and $v_i^k[a_k , b_k ; y_k]$ are the posterior mean and isotropic variance Eqs~\eqref{factor_mean_isotropic} and \eqref{factor_variance_isotropic} as estimated by the EP factor marginal Eq.~\eqref{factor_marginal_isotropic}.
The gradient of the variable RS potential give the dual mapping to the overlap:
\begin{equation}
    m_i[\hat{m}_i] = \EE_{p_i^{(0)}(x_i^{(0)}, b_i)} \frac{r_i[a_i, b_i] \cdot x_i^{(0)}}{N_i} 
    , \quad \tfrac{1}{2} m_i = \partial_{\hat{m}_i} \bar{A}_i^{(0)},
\end{equation}
where $r_i[a_i, b_i] = \frac{b_i}{a_i}$ and $v_i[a_i , b_i] = \frac{1}{a_i}$ are the posterior mean and isotropic variance as estimated by the EP variable marginal Eq.~\eqref{variable_marginal_isotropic}.
$\bar{A}_k^{(0)*}$ and $\bar{A}_i^{(0)*}$ are the corresponding Legendre transforms. 
The ensemble average variances are given by:
\begin{align}
     &v_i^k[\hat{m}_k] = \EE_{p_k^{(0)}(x_k^{(0)}, y_k, b_k)} v_i^k[a_k, b_k ; y_k] 
     =  \tau_i^{(0)} - m_i^k[\hat{m}_k],\\
     &v_i[\hat{m}_i] = \EE_{p_i^{(0)}(x_i^{(0)}, b_i)} v_i[a_i, b_i] 
     = \tau_i^{(0)} - m_i[\hat{m}_i].
\end{align}
Any stationary point of the potentials (not necessarily the global optima) is a fixed point:
\begin{equation}
     m_i^k = m_i, \quad 
     \hat{m}_i = \sum_{k \in \partial i} \hat{m}_{k \to i}.
\end{equation}
\end{proposition}

\paragraph{BO potentials}
The variable BO potential Eq.~\eqref{BO_potential_x} is explicitly given by:
 \begin{equation}
    \label{BO_potential_variable}
     \bar{A}_i^{(0)}[\hat{m}_i] = \frac{\hat{m}_i\tau_i^{(0)}}{2} + \frac{1}{2} \ln \frac{2\pi}{a_i}
    \quad \text{with} \quad a_i = \hat{\tau}_i^{(0)} + \hat{m}_i
 \end{equation}
 which yields the dual mapping:
  \begin{equation}
    \label{BO_dual_mapping}
     m_i =\tau_i^{(0)} - v_i, \quad v_i = \frac{1}{a_i}.
 \end{equation}
 Several factor BO potentials are given in Appendix~\ref{app:modules}.

\paragraph{Entropy estimation and state evolution} 
The ensemble average $\bar{A}^{(0)}$ which gives access to the entropy through Eq.~\eqref{entropy} is given by the global minimum according to Proposition~\ref{prop:BO_free_entropy}. The global minimizer gives access to the ensemble average overlap $m_i$  as well as $\mathrm{mse}_i = \tau_i^{(0)} -m_i = v_i$ for the student posterior Eq.~\eqref{factor_graph_conditional}. As the student posterior is Bayes-optimal $\mathrm{mse_i} = \mathrm{mmse}_i$ is the MMSE. 
However the replica free entropy solution also appears as an ensemble average of the underlying EP Algorithm~\ref{algo:gaussian_EP} where the effective ensemble average is defined locally for each factor Eq.~\eqref{BO_ensemble_avg_f} and variable Eq.~\eqref{BO_ensemble_avg_x}. With this effective ensemble average interpretation in mind we conjecture Algorithm~\ref{algo:SE_RS_matched} to give the state evolution of the EP Algorithm \ref{algo:gaussian_EP} in the Bayes-optimal setting. 
Then the state evolution fixed point will give the overlap $m_i$ as well as $\mathrm{mse}_i = \tau_i^{(0)} -m_i = v_i$ corresponding to the EP student solution, which is in general only a local minimizer in Proposition~\ref{prop:BO_free_entropy}. 

\paragraph{Hard phase} 
When the SE fixed point happens to be the global minimizer, the EP algorithm is in a sense optimal as its solution achieves the same overlap as the Bayes-optimal posterior according to Proposition~\ref{prop:BO_free_entropy}, in particular $\mathrm{mse}_i = \mathrm{mmse}_i$. By contrast, when the SE fixed point fails to be a global minimizer, the EP algorithm is sub-optimal  $\mathrm{mse}_i > \mathrm{mmse}_i$ and is said to be in a computational \emph{hard phase}. Finally note that Algorithm~\ref{algo:SE_RS_matched} can be viewed more generally as an iterative routine to find stationary points of the replica free entropy potential. When initialized as in Algorithm~\ref{algo:SE_RS_matched} it leads to the SE fixed point, but if initialized in the right basin of attraction it leads to the global minimizer. 

\paragraph{Prior work} 
We recover several results as particular cases. The state evolution for the ML-VAMP algorithm in the matched setting, rigorously proven in \citep{Fletcher2018}, is equivalent to Algorithm~\ref{algo:SE_RS_matched} applied to the multi-layer GLM with orthogonally invariant weight matrices. We recover the corresponding replica symmetric free entropy derived in \citep{Gabrie2018}. The \cite{Reeves2017} formalism, developed for tree networks of GLMs and expressed in term of mutual information potentials, is shown to be equivalent to Proposition~\ref{prop:BO_free_entropy} in Section~\ref{sec:information_theory}.

\vspace{1em}
\begin{algorithm}[H]
\label{algo:SE_RS_matched}
\DontPrintSemicolon
\SetKw{Initialize}{initialize}
\Initialize{$\hat{m}_{i\to k}, \hat{m}_{k\to i}= 0$}\\
\Repeat{convergence}{
  \ForEach(\tcp*[f]{ forward and backward pass}){edge $e \in E_+ \cup E_-$}{
    \If{$e = k \to i$}{
      $a_i^k =  1 / v_i^k[\hat{m}_k], \quad \hat{m}_i^k = a_i^k - \hat{\tau}_i^{k(0)}$ \tcp*[f]{variance-matching}\\
      $\hat{m}_{k \to i}^\text{new} = \hat{m}_i^k - \hat{m}_{i \to k}$
      \tcp*[f]{message $f_k \to x_i$ }
    }
    \If{$e = i \to k$}{
      $\hat{m}_i = \sum_{k' \in \partial i} \hat{m}_{k' \to i}$
      \tcp*[f]{precision }\\
       $\hat{m}_{i \to k}^\text{new} = \hat{m}_i^k - \hat{m}_{k \to i}$
      \tcp*[f]{message $x_i \to f_k$ }
    }
  }
}
\caption{\tramp State evolution (Bayes-optimal setting)}
\end{algorithm}

\subsection{Information theoretic expressions \label{sec:information_theory}}

\paragraph{Local teacher-student scenario}
The RS factor potential Eq.~\eqref{RS_potential_f}
 is given by a local ensemble average of the EP log-partition. The effective ensemble average in Eq.~\eqref{RS_ensemble_avg_f} can be interpreted as a local teacher-student scenario. The local teacher generative model is given by:
 \begin{equation}
    \label{local_teacher}
    p_k^{(0)}(x_k^{(0)}, y_k, b_k) = \mathcal{N}(b_k \mid \hat{m}_k x_k^{(0)}, \hat{q}_k) 
    \underbrace{ 
        f_k^{(0)}(x_k^{(0)}; y_k) \,
        e^{-\frac{1}{2}\hat{\tau}_k^{(0)} \Vert x_k^{(0)}\Vert^2 
        - N_k A_k^{(0)}[\hat{\tau}_k^{(0)}] }
    }_{ p_k^{(0)}(x_k^{(0)}, y_k \mid \hat{\tau}_k^{(0)})}
\end{equation}
while the local student generative model is given by:
\begin{equation}
    \label{local_student}
    p_k(x_k, y_k, b_k) = \mathcal{N}(b_k \mid \hat{q}_k x_k, \hat{q}_k)  
    \underbrace{
        f_k(x_k; y_k) \,
        e^{-\frac{1}{2}\hat{\tau}_k \Vert x_k\Vert^2 
        - N_k A_k[\hat{\tau}_k] }
    }_{  p_k(x_k, y_k \mid \hat{\tau}_k) }
\end{equation}
 The local student posterior is then given by:
\begin{equation}
    p_k(x_k \mid y_k, b_k) = f_k(x_k ; y_k) \, e^{- \frac{1}{2} a_k \Vert x_k \Vert^2 + b_k^\intercal x_k - N_k A_k[a_k,b_k;y_k]} \quad \text{with} \quad a_k = \hat{\tau}_k + \hat{q}_k
\end{equation}
which we recognize as the EP factor marginal Eq.~\eqref{factor_marginal_isotropic}.
The rescaled messages $x_k^\sharp = b_k / \sqrt{\hat{q}_k} $ act as pseudo-measurements of the signals $x_k$ corrupted by Gaussian noise. The local teacher Eq.~\eqref{local_teacher} actually generates $x_k^\sharp$ according to:
\begin{equation}
    x_k^\sharp = \sqrt{\hat{m}_k^{(0)}} x_k + \xi_k,  \quad 
    \xi_k \sim \mathcal{N}(0,1), \quad 
    \hat{m}_k^{(0)} = \frac{\hat{m}_k^2}{\hat{q}_k}
\end{equation}
with signal-to-noise ratios (SNR) $\hat{m}_k^{(0)}$ 
while the student Eq.~\eqref{local_student} believes that the $x_k^\sharp$ are generated with SNR $\hat{q}_k$.

\paragraph{Bayes-optimal setting}
The teacher and student factors are matched $f_k = f_k^{(0)}$ and the local teacher and student generative models are identical:
\begin{equation}
    \label{local_BO}
    p_k^{(0)}(x_k^{(0)},y_k, b_k) = \mathcal{N}(b_k \mid \hat{m}_k x_k^{(0)}, \hat{m}_k) 
     \underbrace{ 
        f_k^{(0)}(x_k^{(0)}; y_k) \,
        e^{-\frac{1}{2}\hat{\tau}_k^{(0)} \Vert x_k^{(0)}\Vert^2 - A_k^{(0)}[\hat{\tau}_k^{(0)}] }
    }_{ p_k^{(0)}(x_k^{(0)}, y_k \mid \hat{\tau}_k^{(0)})}
\end{equation}
We recall that in the Bayes-optimal setting $\hat{m}_k = \hat{q}_k$ and $\hat{\tau}_k = \hat{\tau}_k^{(0)}$
and consistently the teacher and student SNR are matched $\hat{m}_k^{(0)} = \hat{q}_k = \hat{m}_k$.
The local student posterior is the EP factor marginal Eq.~\eqref{factor_marginal_isotropic}
\begin{equation}
    p_k^{(0)}(x_k \mid y_k, b_k) = f_k^{(0)}(x_k ; y_k) e^{- \frac{1}{2} a_k \Vert x_k \Vert^2 + b_k^\intercal x_k - N_k A_k^{(0)}[a_k,b_k;y_k]} \quad \text{with} \quad a_k = \hat{\tau}_k^{(0)} + \hat{m}_k
\end{equation}
and is Bayes-optimal. In particular the posterior mean $r_i^k[a_k, b_k; y_k]$ is the MMSE estimator of the signal $x_i$ and the posterior variance $v_i^k[a_k, b_k; y_k]$ gives the MMSE.

\paragraph{Decomposition of the RS and BO factor potentials} 
With these local teacher and student generative models, we can give the following information theoretic interpretation of the RS factor potential Eq.~\eqref{RS_potential_f} and BO factor potential Eq.~\eqref{BO_potential_f}, see Appendix~\ref{app:proof_IT} for a proof. The RS potential differs from the BO potential by a KL divergence term:
\begin{align}
\notag
&\bar{A}_k[\hat{m}_k,\hat{q}_k,\hat{\tau}_k] - A_k[\hat{\tau}_k] = \bar{A}_k^{(0)}[\hat{m}_k^{(0)}] - A_k^{(0)}[\hat{\tau}_k^{(0)}]  - K_k[\hat{m}_k,\hat{q}_k,\hat{\tau}_k] \\
\label{RS_potential_decomposition}
&\text{with} \quad 
K_k[\hat{m}_k,\hat{q}_k,\hat{\tau}_k] = \lim_{N_k \to \infty} \frac{1}{N_k} \mathrm{KL}[p_k^{(0)}(y_k, b_k) \Vert p_k(y_k, b_k)]  \quad \text{and} \quad \hat{m}_k^{(0)} = \frac{\hat{m}_k^2}{\hat{q}_k}.
\end{align}
$K_k$ is the KL divergence between the local teacher evidence $p_k^{(0)}(y_k, b_k)$ and the local student evidence $p_k(y_k, b_k)$. The BO potential is directly related to an entropic term:
\begin{align}
\notag
&\bar{A}_k^{(0)}[\hat{m}_k] - A_k^{(0)}[\hat{\tau}_k^{(0)}] =  \sum_{i \in \partial k} \frac{\alpha_i^k \hat{m}_{i \to k} \tau_i^{(0)}}{2} - H_k[\hat{m}_k] \\
\label{BO_potential_decomposition}
&\text{with} \quad 
H_k[\hat{m}_k] = \lim_{N_k \to \infty} \frac{1}{N_k} H[y_k, b_k] - \frac{1}{N_k} H[b_k \mid x_k^{(0)}],
\end{align}
where the entropies are defined over the random variables $x_k^{(0)}, y_k, b_k$ distributed according to Eq.~\eqref{local_BO}.

\paragraph{Reeves formalism} 
The entropic expression Eq.~\eqref{BO_potential_decomposition} in the Bayes-optimal setting allows to recover the \citep{Reeves2017} formalism developed  for a tree network of GLMs (Example~\ref{example:glm_condition}) and extends it to other tree-structured factor graphs.
For a non-likelihood factor, that is $y_k = \emptyset$, the entropic potential $H_k$ reduced to the mutual information between the signals $x_k^{(0)}$ and the pseudo-measurements $x_k^\sharp$:
\begin{equation}
    H_k[\hat{m}_k] = I_k[\hat{m}_k] = \lim_{N_k \to \infty} \frac{1}{N_k} I[x_k^{(0)} ; x_k^\sharp] .
\end{equation}
For a likelihood factor $y_k \neq \emptyset$, the entropic potential $H_k$ reduced to the mutual information between the signals $x_k^{(0)}$ and the pair of measurements $y_k$ and pseudo-measurements $x_k^\sharp$ plus an entropic noise term:
\begin{align}
    \notag
    &H_k[\hat{m}_k] = I_k[\hat{m}_k] + E_k , \\
    &I_k[\hat{m}_k] = \lim_{N_k \to \infty} \frac{1}{N_k} I[x_k^{(0)} ; (y_k, x_k^\sharp)], \quad
    E_k = \lim_{N_k \to \infty} \frac{1}{N_k} H[y_k \mid x_k^{(0)}] .
\end{align}
For a noiseless output channel (deterministic relationship between $y_k$ and $x_k^{(0)}$) the mutual information $I_k$ and the entropic noise $E_k$ can be ill-defined and the more general relation Eq.~\eqref{BO_potential_decomposition} should be preferred.
The mutual information potential $I_k$ and the entropic potential $H_k$ are functions of the SNR $\hat{m}_k$ . Their gradients give the the dual mapping with the variances:
\begin{equation}
    \frac{1}{2} \alpha_i^k v_i^k = \partial_{\hat{m}_{i \to k}} I_k[\hat{m}_k] =  \partial_{\hat{m}_{i \to k}} H_k[\hat{m}_k]
\end{equation}
known as the I-MMSE relationship \citep{Guo2005} as $v_i^k$ gives the MMSE. Then Proposition~\ref{prop:BO_free_entropy} is exactly equivalent to the \cite{Reeves2017} formalism, where the BO potentials $\bar{A}_k^{(0)}$ are replaced by the mutual information potentials $I_k$ and the overlaps $m_k$ by the variances $v_k$.

\paragraph{RS and BO variable potentials} 
The RS variable potential Eq.~\eqref{RS_potential_x}
 is given by a local ensemble average of the EP log-partition. The effective ensemble average in Eq.~\eqref{RS_ensemble_avg_x} can be interpreted as a local teacher-student scenario.
The local teacher and student generative models are the Gaussians:
\begin{align}
    &p_i^{(0)}(x_i^{(0)}, b_i ) = \mathcal{N}(b_i \mid \hat{m}_i x_i^{(0)}, \hat{q}_i) \, \mathcal{N}(x_i^{(0)} \mid 0, 1/\hat{\tau}_i^{(0)} ), \\
    &p_i(x_i, b_i ) = \mathcal{N}(b_i \mid \hat{q}_i x_i, \hat{q}_i) \, \mathcal{N}(x_i \mid 0, 1/\hat{\tau}_i).
\end{align}
In the Bayes-optimal setting the local teacher and student generative models are identical.
The  RS and BO variable potentials follows the same decomposition Eqs~\eqref{RS_potential_decomposition}-\eqref{BO_potential_decomposition} 
as the factor potentials. 
But in that case the decomposition can be straightforwardly checked from the explicit expressions:
\begin{align}
    &\bar{A}_i[\hat{m}_i,\hat{q}_i,\hat{\tau}_i] = 
    \frac{\hat{m}_i^2 \tau_i^{(0)} + \hat{q}_i}{2a_i} + \frac{1}{2} \ln \frac{2\pi}{a_i}
    \quad  \text{with} \quad 
    a_i = \hat{\tau}_i + \hat{q}_i, \quad
    A_i[\hat{\tau}_i] = \frac{1}{2} \ln \frac{2\pi}{\hat{\tau}_i}, \\
    &K_i[\hat{m}_i,\hat{q}_i,\hat{\tau}_i] = \frac{1}{2} \left(
     \ln \frac{v_S}{v_T} + \frac{v_T}{v_S} - 1
     \right)
     \quad \text{with} \quad
     v_T = 1 +  \frac{\hat{m}_i^{(0)}}{\hat{\tau}_i^{(0)}}, \quad 
     v_S = 1 +  \frac{\hat{q}_i}{\hat{\tau}_i} ,\\
    &\bar{A}_i^{(0)}[\hat{m}_i^{(0)}] = \frac{\hat{m}_i^{(0)} \tau_i^{(0)}}{2} + \frac{1}{2} \ln \frac{2\pi}{a_i^{(0)}}
     \quad \text{with} \quad
     a_i^{(0)} = \hat{\tau}_i^{(0)} + \hat{m}_i^{(0)} , \;
    A_i^{(0)}[\hat{\tau}_i^{(0)}] =  \frac{1}{2} \ln \frac{2\pi}{\hat{\tau}_i^{(0)}}, \\
     &H_i[\hat{m}_i^{(0)}] = I_i[\hat{m}_i^{(0)}] = \frac{1}{2} \ln a_i^{(0)} \tau_i^{(0)}.
\end{align}
$K_i$ is the KL divergence between the local teacher evidence $p_i^{(0)}(x_i^\sharp) = \mathcal{N}(x_i^\sharp \mid 0, v_T)$ and the local student evidence $p_i(x_i^\sharp) = \mathcal{N}(x_i^\sharp \mid 0, v_S)$ of the pseudo-measurement $x_i^\sharp = b_i / \sqrt{\hat{q}_i}$.

\subsection{State evolution modules}

For each EP factor module with isotropic Gaussian beliefs (Section~\ref{sec:EP_modules}) one can easily implement the corresponding free entropy / state evolution module by taking the ensemble average Eq.~\eqref{RS_ensemble_avg_f} in the replica symmetric (RS) setting or Eq.~\eqref{BO_ensemble_avg_f} in the Bayes-optimal (BO) setting. In practice the RS module must be able to compute the RS potential $\bar{A}_k[\hat{m}_k,\hat{q}_k, \hat{\tau}_k]$ given in Eq.~\eqref{RS_potential_f} and the associated overlaps $m_k[\hat{m}_k,\hat{q}_k, \hat{\tau}_k]$,  $q_k[\hat{m}_k,\hat{q}_k, \hat{\tau}_k]$ and  $\tau_k[\hat{m}_k,\hat{q}_k, \hat{\tau}_k]$ given in Eqs~\eqref{RS_overlap_f}-\eqref{RS_tau_f}.
Similarly the BO module must be able to compute the BO potential $\bar{A}_k^{(0)}[\hat{m}_k]$ given in Eq.~\eqref{BO_potential_f} and the associated overlap $m_k[\hat{m}_k]$ given in Eq.~\eqref{BO_overlap_f}.
We have closed-form expressions for the linear channel in the RS (Appendix~\ref{module:linear_rs}) and BO (Appendix~\ref{module:linear_bo}) cases.
For separable factors the RS and BO potentials and associated overlaps can be analytically obtained through a low dimensional integration: see Appendix~\ref{module:prior_rs}-\ref{module:prior_bo} for a separable prior, 
Appendix~\ref{module:likelihood_rs}-\ref{module:likelihood_bo} for a separable likelihood, and
Appendix~\ref{module:channel_rs}-\ref{module:channel_bo} for a separable channel.

\section{Examples \label{sec:Examples}}
 This section is dedicated to illustrating the \tramp package. We first point out that its reconstruction performances asymptotically reaches the Bayes optimal limit out of the hard phase and that its fast execution speed often exceeds competing algorithms. Moreover we stress that the cornerstone of \tramp is its modularity, which allows it to handle a wide range of inference tasks. To appreciate its great flexibility, we illustrate its performance on various tree-structured models. Finally, the last section depicts the ability of \tramp to predict its own state evolution performance on two simple GLMs: compressed sensing and sparse phase retrieval.
The codes corresponding to the examples presented in this section can be found in the documentation gallery\footnote{\href{https://sphinxteam.github.io/tramp.docs/0.1/html/gallery/index.html}{https://sphinxteam.github.io/tramp.docs/0.1/html/gallery/index.html}} or in the repository\footnote{\href{https://github.com/sphinxteam/tramp/tree/master/examples/figures}{https://github.com/sphinxteam/tramp/tree/master/examples/figures}}.

\subsection{Benchmark on sparse linear regression \label{sec:sparse_regression}}
\begin{figure}[H]
\centering
    \begin{tikzpicture}
        \node[latent,label=$\bbR^{N}$]  (x)  {$x$} ;
        \node[latent, right=of x,label=$\bbR^{M}$] (z) {$z$} ;
        \node[obs, right=of z, label=$\bbR^{M}$] (y) {$y$} ;
        \factor[left=of x] {pF} {$\mN_\rho$} {} {x} ;
        \factor[left=of y] {p} {$\Delta$} {z} {y} ;
        \factor[left=of z] {pF} {$A$} {x} {z} ;
    \end{tikzpicture}
    \caption{Sparse linear regression factor graph.}
    \label{fig:PGM_sparse_regression}
\end{figure}
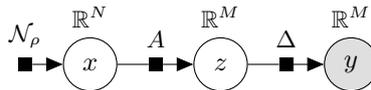

 Let us consider a sparse signal $x\in \bbR^{N}$, $\iid$ drawn according to $x \sim \prod_{n=1}^N \mathcal{N}_\rho(x_n)$ where $\mathcal{N}_\rho = [1-\rho] \delta + \rho \mathcal{N}$ is the Gauss-Bernoulli prior and $\mathcal{N}$ the normal distribution. The inference task is to reconstruct the signal $x$ from noisy observations $y \in \bbR^{M}$ generated according to
 \begin{align}
            y = Ax + \xi
            \label{eq:sparse_reg}
\end{align}
where $A\in\bbR^{M \times N}$ is the sensing matrix with $\iid$ Gaussian entries $A_{mn} \sim \mN(0,1/N)$ and $\xi$ is a $\iid$ Gaussian noise $\xi \sim \mN(0,\Delta)$. We define $\alpha=M/N$ the aspect ratio of the matrix $A$. The corresponding factor graph is depicted in Figure~\ref{fig:PGM_sparse_regression}.

The sparse linear regression problem can be easily solved with the \tramp package. We simply need to import the necessary modules, declare the model, and run the Expectation Propagation algorithm:
\begin{lstlisting}[language=Python]
# import modules
from tramp.base import Variable as V
from tramp.priors import GaussBernoulliPrior
from tramp.likelihoods import GaussianLikelihood
from tramp.channels import LinearChannel
# declare sparse linear regression model
model = (
    GaussBernoulliPrior(rho=rho, size=N) @ V('x') @
    LinearChannel(A) @ V('z') @ 
    GaussianLikelihood(var=Delta, y=y)
).to_model()
# run EP
from tramp.algos import ExpectationPropagation
ep = ExpectationPropagation(model)
ep.iterate(max_iter=200)
\end{lstlisting}

\begin{figure}[t]
    \centering
    \includegraphics[width=0.95\linewidth]{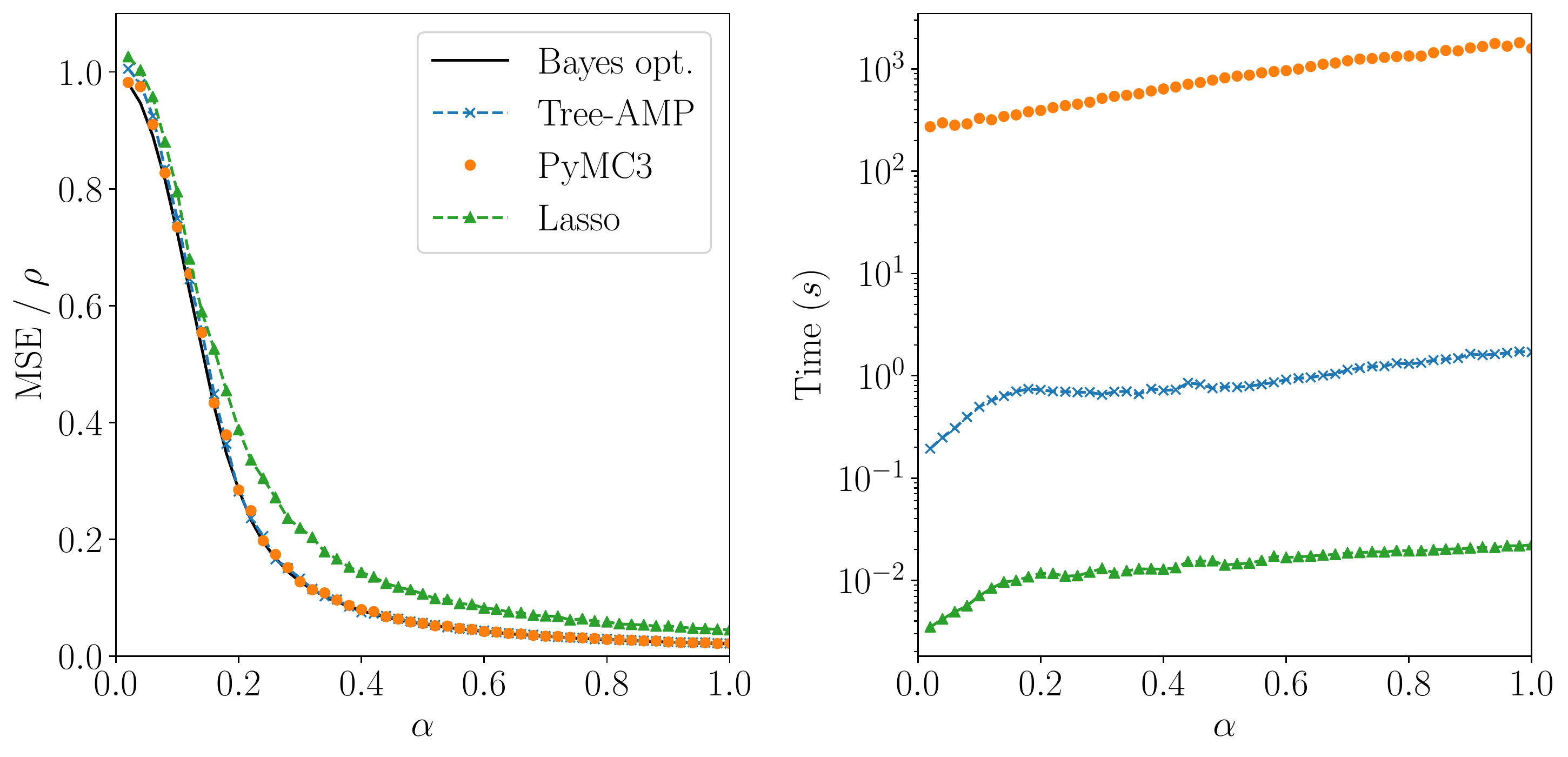}
    \caption{Benchmark on a sparse linear regression task: rescaled MSE as a function of $\alpha=M/N$. The MSE achieved by \tramp (blue) is compared to the Bayes-optimal MMSE (black), Hamiltonian Monte-Carlo (orange) from \pymc (with $n_{\rm s}=1000$ distribution samples and NUTS sampler) and Lasso (green) from \scikit (with the optimal regularization parameter obtained beforehand by simulation). The above experiments have been performed with parameters $(N, \rho, \Delta)=(1000, 0.05, 0.01)$ and have been averaged over 100 samples.}
    \label{fig:benchmark_sparse}
\end{figure}

We compare the \tramp performance on this inference task to the Bayes optimal theoretical prediction from \citep{Barbier2019} to two state of the art algorithms for this task: Hamiltonian Monte-Carlo from the \pymc package \citep{pymc3} and Lasso (L1-regularized linear regression) from the \scikit package \citep{scikit-learn}. Note that to perform our experimental benchmark in Figure~\ref{fig:benchmark_sparse} the \tramp and \pymc algorithms had access to the ground-truth parameters $(\rho,\Delta)$ used to generate the observations. In order to make the benchmark as fair as possible, we use the optimal regularization parameter for the Lasso, obtained beforehand by simulation. 

We observe in Figure~\ref{fig:benchmark_sparse} (\textbf{left}) that for this model \tramp is Bayes-optimal and reaches the MMSE, up to finite size fluctuations, just as \pymc. They naturally both outperform Lasso from \scikit  that never achieves the Bayes-optimal MMSE for the full range of aspect ratio $\alpha$ under investigation. This is expected and unfair to Lasso as the two Bayesian methods have full knowledge of the exact generating distribution in our toy model, but this is rarely the case in real applications.

Whereas the Hamiltonian Monte-Carlo algorithm requires to draw a large number of  samples ($n_{\rm s}= 10^{3}$) to reach a given threshold of precision, \tramp is an iterative algorithm that converges in a few iterations varying broadly speaking between $[10^0;10^2]$. It leads interestingly to an execution time smaller by two orders of magnitude with respect to \pymc as illustrated in Figure~\ref{fig:benchmark_sparse} (\textbf{right}). Hence the fast convergence and execution time of \tramp is certainly a deep asset over \pymc, or similar Markov Chain Monte-Carlo packages.

\subsection{Depicting  \tramp modularity}
In order to show the adaptability and modularity of \tramp to handle various inference tasks, we present here different examples where the prior distributions are modified flexibly. In particular, we consider first Gaussian denoising of synthetic data with either sparse discrete Fourier transform (DFT) or sparse gradient, and second the denoising and inpainting of real images drawn from the MNIST data set, using a trained Variational Auto-Encoder (VAE) as a prior. 

\subsubsection{Sparse DFT/gradient denoising}

Let us consider a signal $x\in\bbR^{N}$ corrupted by a Gaussian noise $\xi\sim \mN(0,\Delta)$, that leads to the observation $y\in\bbR^{N}$ according to 
\begin{align}
    y = x + \xi
    \label{eq:denoising_gaussian_channel}
\end{align}

\begin{figure}[H]
          \centering
            \begin{tikzpicture}
                \node[latent,label=$\bbR^{N}$]  (x)  {$x$} ;
                \node[obs, right=of x, yshift=0.8cm,label=$\bbR^{N}$] (y) {$y$} ;
                \node[latent, right=of x, yshift=-0.8cm] (z) {$z$} ;
                \factor[left=of y] {p} {$\Delta$} {x} {y} ;
                \factor[left=of z] {pF} {$\Omega$} {x} {z} ;
                \factor[right=of z] {p} {right:$\mN_\rho$} {} {z} ;
                \factor[left=of x] {p} {left:$\mN$} {} {x} ;
            \end{tikzpicture}
            \caption{Factor graph for sparse $\Omega$ denoising, where $\Omega$ represents either the DFT or the gradient operator.}
            \label{fig:PGM_sparse_operator}
\end{figure}
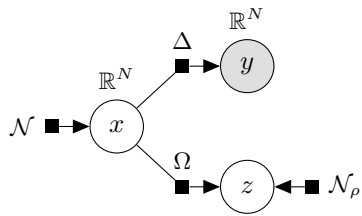

In contrast to the first section in which we considered the signal $x$ to be sparse, we assume here that the signal is dense but that a linear transformation of the signal is sparse. In other words let us define the variable $z=\Omega x$ that we assume to be sparse, where $\Omega$ denotes a linear operator acting on the signal. The factor graph associated to this model is depicted in Figure~\ref{fig:PGM_sparse_operator}. 
          
As a matter of clarity, we focus on two toy one-dimensional signals:
\begin{enumerate}
            \item $x\in \bbR^{N}$ such that $\forall n\in[1:N], x_n=\cos(t_n) + \sin(2 t_n)$, with $t_n=2\pi(-1 + \frac{2 n}{N} )$. The signal is sparse in the Fourier basis with only two spikes, that leads us to consider a sparse DFT prior: $\Omega$ is the discrete Fourier transform,
            \item $x\in \bbR^{N}$ such that it is randomly drawn constant by pieces. Its gradient contains a lot of zeros and therefore inference with a sparse gradient prior is appropriate: $\Omega$ is the gradient operator.
\end{enumerate}

\begin{figure}[h]
    \centering
    \includegraphics[width=0.90\linewidth]{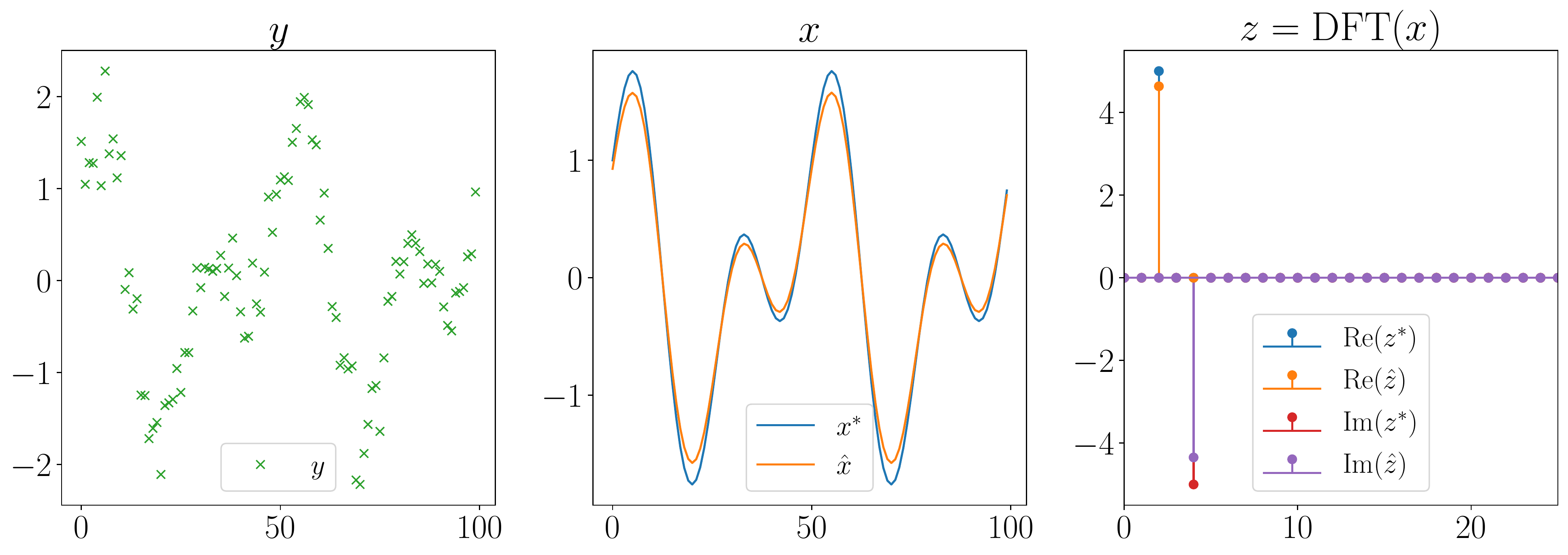}
    \includegraphics[width=0.90\linewidth]{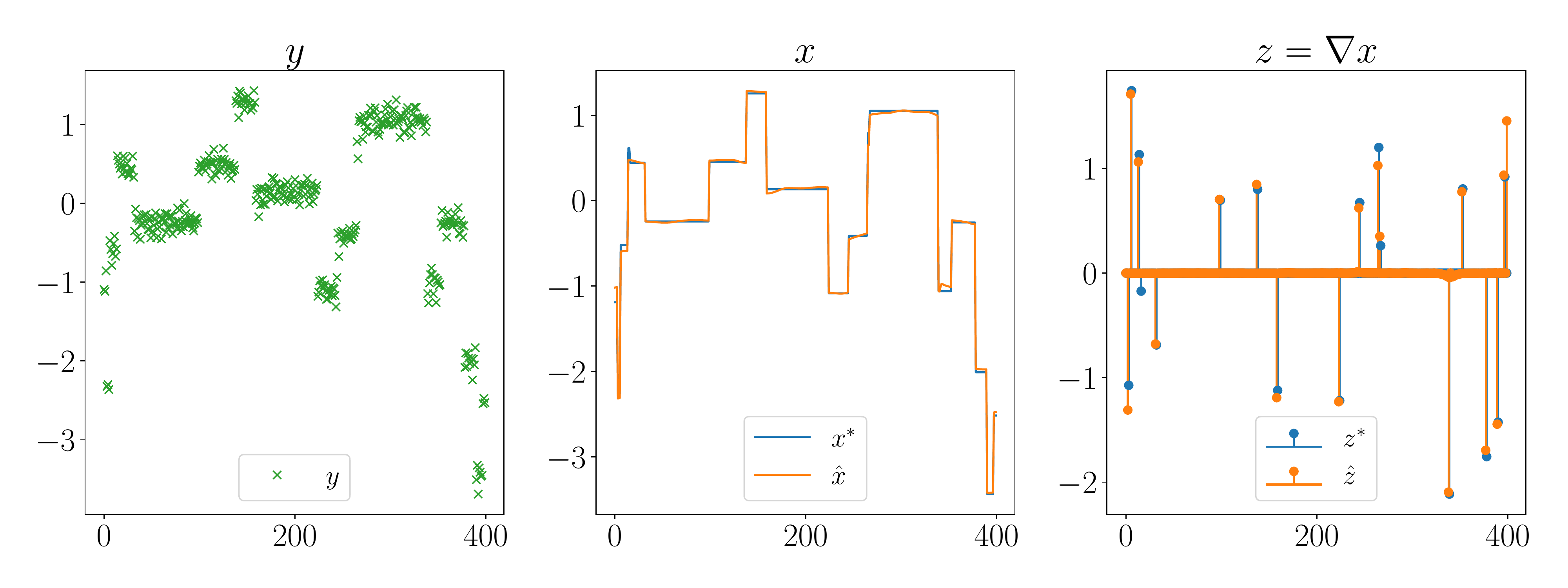}
    \caption{Sparse FFT/gradient denoising: \textbf{(left)} noisy observation $y$, \textbf{(middle)} ground truth signal $x^*$ and predicted $\hat{x}$ and \textbf{(right)} ground truth linear transform $z^*$ and predicted $\hat{z}$ for \textbf{(upper)} sparse DFT denoising with $(N,\rho,\Delta)=(100,0.02,0.1)$ and \textbf{(lower)} sparse gradient denoising with $(N,\rho,\Delta)=(400,0.04,0.01)$. }
    \label{fig:sparse_fft_gradient}
\end{figure}    

After importing the relevant modules, declaring the model in the \tramp package is simple,
for instance for the sparse gradient model:
\begin{lstlisting}[language=Python]
# sparse gradient denoising
model = (
  GaussianPrior(size=N) @ V('x', n_prev=1, n_next=2) @ (
    GaussianLikelihood(var=Delta, y=y) + (
      GradientChannel() + 
      GaussBernoulliPrior(rho=rho, size=(1,N))
    ) @ V('z', n_prev=2, n_next=0)
  )
).to_model()
\end{lstlisting}
For the sparse DFT model, one just needs to replace {\ttfamily GradientChannel} by {\ttfamily DFTChannel}. Numerical experiments are shown in Figure~\ref{fig:sparse_fft_gradient}. The left panel shows the observation $y$, while the middle and right panels illustrate the \tramp reconstruction of the signal $\hat{x}$ and of its linear transform $\hat{z}$ compared to the ground truth $x^*$ and $z^*$. The \tramp reconstruction approaches closely the ground truth signal and leads to MSE $\sim 10^{-2}/10^{-3}$ for signals $1/2$.

\subsubsection{Variational Auto-Encoder on MNIST}

Let us consider a signal $x\in \bbR^{N}$ (with $N=784$) drawn from the MNIST data set. We want to reconstruct the original image from a corrupted observation $y = \varphi(x) \in\bbR^{N}$, where $\varphi:\bbR^{N}\to\bbR^{N}$ represents a noisy channel. In the following the noisy channel represents either a Gaussian additive channel or an inpainting channel, that erases some pixels of the input image. 

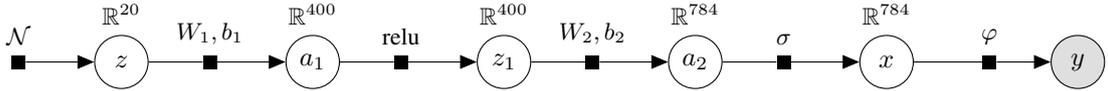
\begin{figure}[H]
\centering
  \begin{tikzpicture}
    \node[latent, label=$\bbR^{20}$]  (z)  {$z$} ;
    \node[latent, right=1.8 of z, label=$\bbR^{400}$]  (a1)  {$a_1$} ;
    \node[latent, right=1.8 of a1, label=$\bbR^{400}$] (z1) {$z_1$} ;
    \node[latent, right=1.8 of z1, label=$\bbR^{784}$] (a2) {$a_2$} ;
    \node[latent, right=1.8 of a2, label=$\bbR^{784}$] (x) {$x$} ;
    \node[obs, right=1.8 of x] (y) {$y$} ;

    \factor[left=0.9 of z] {p} {$\mN$} {} {z} ;
    \factor[left=0.9 of a1] {p} {$W_1, b_1$} {z} {a1} ;
    \factor[left=0.9 of z1] {p} {$\text{relu}$} {a1} {z1} ;
    \factor[left=0.9 of a2] {p} {$W_2, b_2$} {z1} {a2} ;
    \factor[left=0.9 of x] {p} {$\sigma$} {a2} {x} ;
    \factor[right=0.9 of x] {p} {$\varphi$} {x} {y} ;
  \end{tikzpicture}
 \caption{Denoising/inpaiting a MNIST image with a VAE prior. The weights $W_1, W_2$ and biases $b_1, b_2$ were learned beforehand on the MNIST data set.}
 \label{fig:factorgraph_vae}
\end{figure}

In order to reconstruct correctly the MNIST image, we investigated the possibility of using a generative prior such as a Variational Auto-Encoder (VAE) along the lines of \citep{bora2017compressed,Fletcher2018}. The information theoretical and approximate message passing properties of reconstruction of a low rank or GLM channel, using a dense feed-forward neural network generative prior with $\iid$ weights has been studied in particular in \citep{Aubin2019,aubin2019exact}. The VAE architecture is summarized in Figure~\ref{fig:factorgraph_vae} and the training procedure on the MNIST data set follows closely the canonical one detailed in \citep{keras_vae}. We considered two common inference tasks: denoising and inpainting.
%However, neither information theoretical or algorithmic perspective was investigated to handle a \emph{trained} generative prior with non-$\iid$ weights.

\paragraph{Denoising:}
In that case, the corrupted channel $\varphi_{\rm den, \Delta}$ adds a Gaussian noise and corresponds to the noisy channel 
$$\varphi_{\rm den, \Delta}(x) = x + \xi \quad \textrm{with} \quad \xi \sim \mN(0,\Delta)\,.$$ 
\paragraph{Inpaiting:}
The corrupted channel erases a few pixels of the input image and corresponds formally to
$$\varphi_{\rm inp, I_\alpha}(x) = \begin{cases} 0 \textrm{ if } i\in \rm{I}_{\alpha}\,,\\ x_i \textrm{ otherwise}\,.\end{cases}$$
where $\rm{I}_{\alpha}$ denotes the set of erased indexes of size $\left\lfloor \alpha N \right\rfloor$ for some $\alpha \in [0;1]$. As an illustration, we consider two different manners of generating the erased interval $I_\alpha$:
\begin{enumerate}
    \item A central horizontal band of width $\lfloor\alpha N\rfloor$: $\rm{I}_{\alpha}^{\rm band}=[\lfloor\frac{N}{2}(1 -\alpha)\rfloor ; \lfloor\frac{N}{2}(1 + \alpha)\rfloor]$
    \item Indices drawn uniformly at random $\lfloor\alpha N\rfloor$ : $\rm{I}_{\alpha}^{\rm uni} \sim U([1,N];\lfloor\alpha N \rfloor)$
\end{enumerate}

\begin{figure}[t]
\centering
\begin{minipage}[c]{0.45\linewidth}
  \includegraphics[scale=0.38]{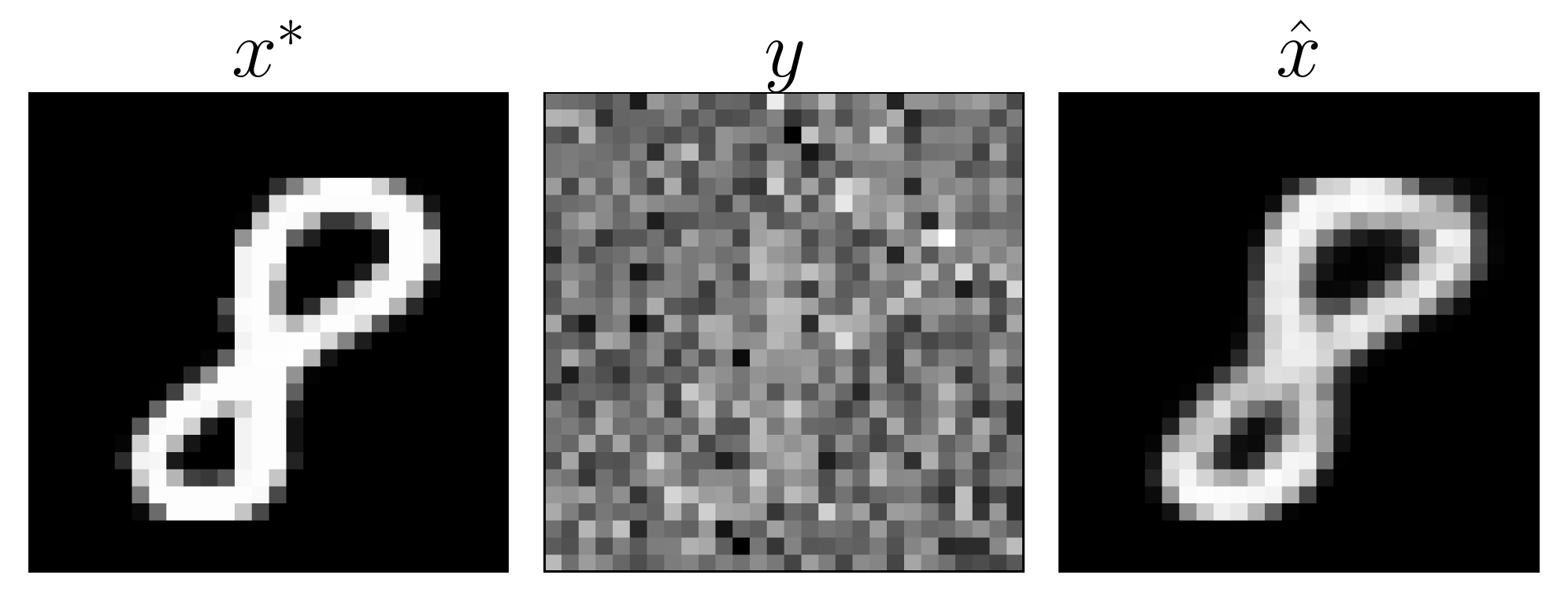}
  \includegraphics[scale=0.38]{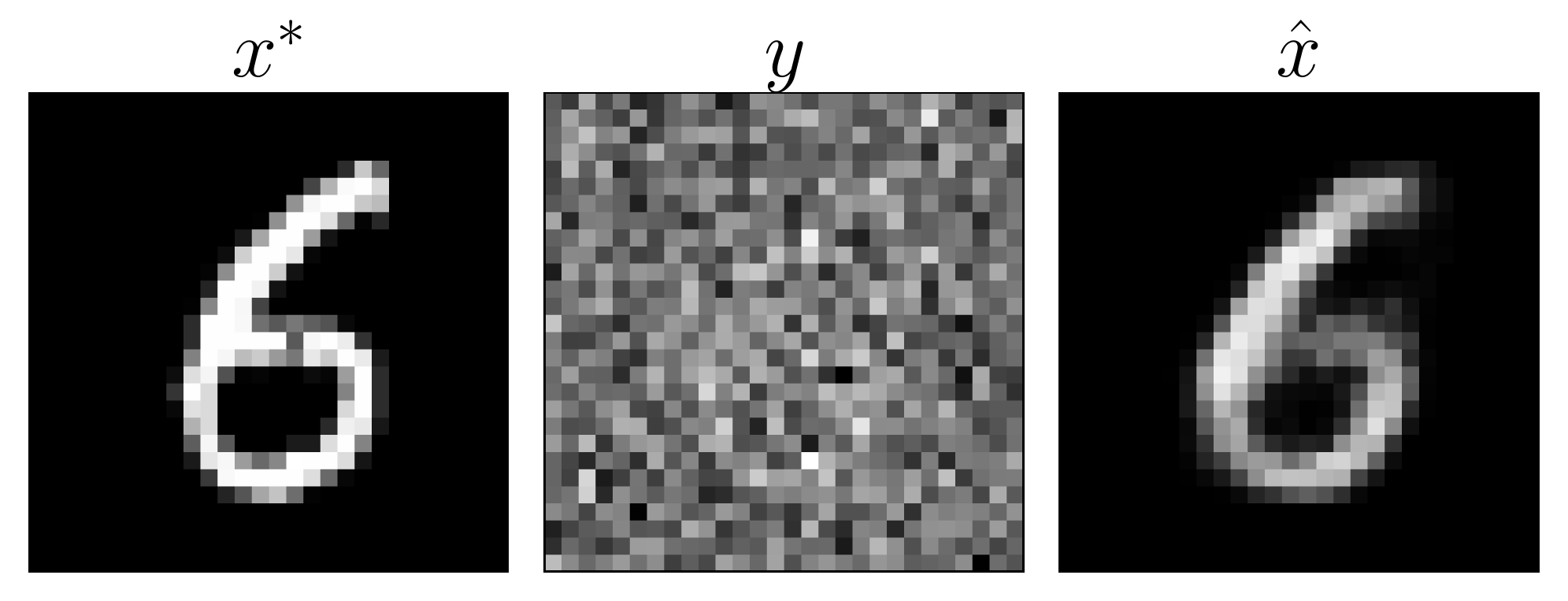}
\end{minipage}
\hspace{0.6cm}
\begin{minipage}[c]{0.45\linewidth}
  \includegraphics[scale=0.37]{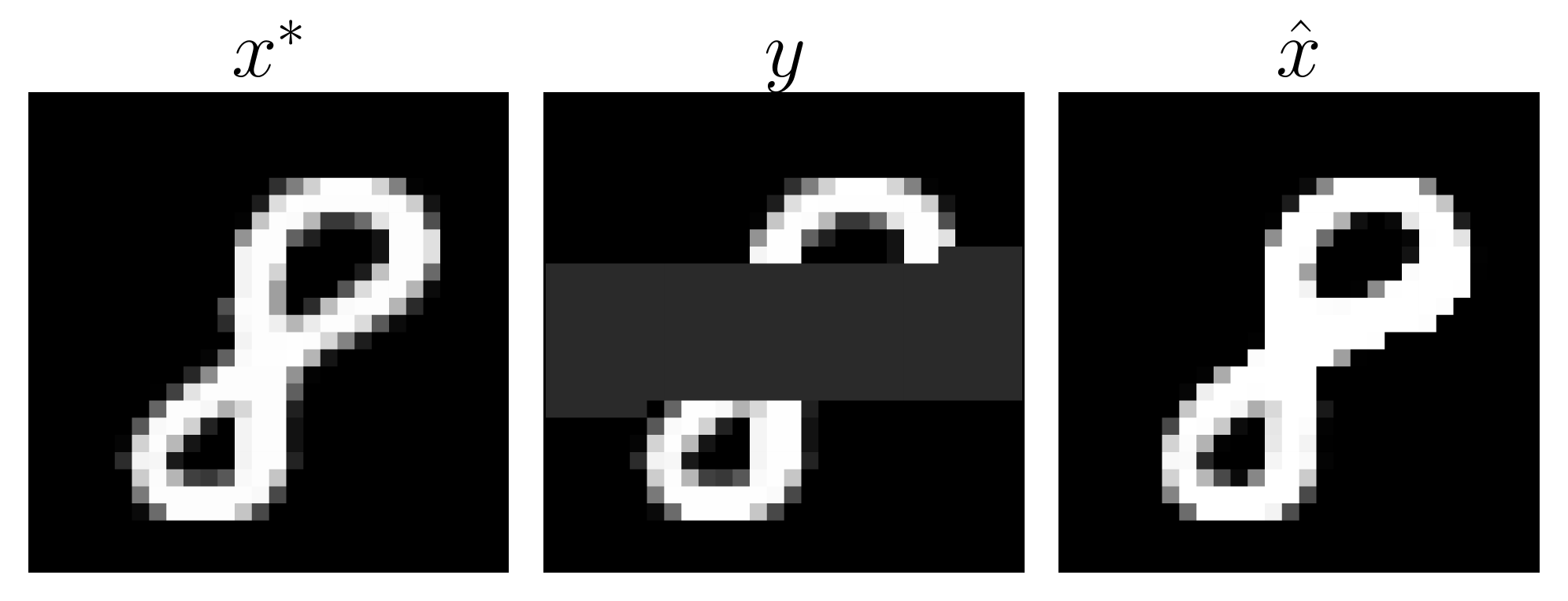}
  \includegraphics[scale=0.37]{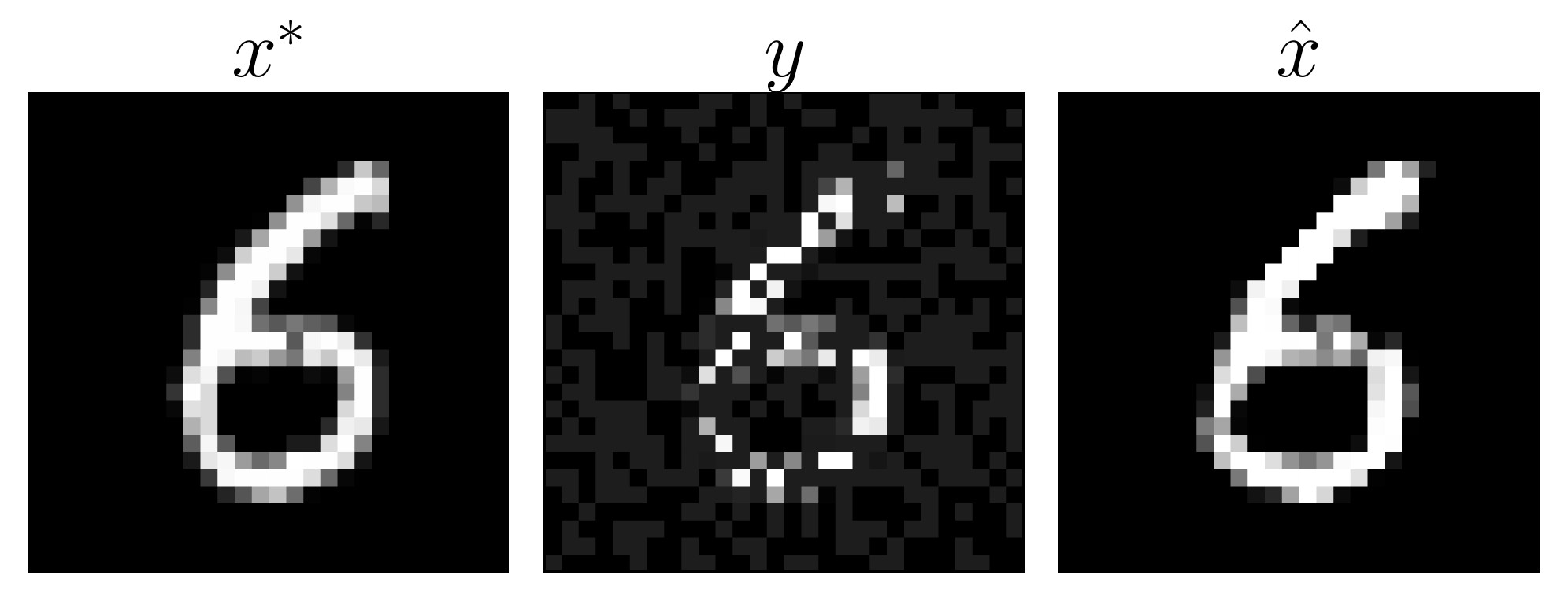}
\end{minipage}
    \caption{Illustration of the \tramp prediction $\hat{x}$ using a VAE prior from observation $y = \varphi(x^*)$ with $x^*$ a MNIST sample.
    \textbf{(left)} Denoising $\varphi = \varphi_{\rm den, \Delta}$ with $\Delta =4$.   
    \textbf{(right-upper)} Band-inpainting $\varphi_{\rm inp, I_\alpha^{\rm band}}$ with $\alpha=0.3$
    \textbf{(right-lower)} Uniform-inpainting $\varphi_{\rm inp, I_\alpha^{\rm uni}}$ with $\alpha=0.5$.}
    \label{fig:vae_denoising}
\end{figure}

Solving these inference tasks in \tramp is straightforward: first declare the
model Figure~\ref{fig:factorgraph_vae} and then run Expectation Propagation as exemplified in Section~\ref{sec:sparse_regression} for the sparse regression case. A few MNIST samples $x^*$ compared to the noisy observations $y$ and \tramp reconstructions $\hat{x}$ are presented in Figure~\ref{fig:vae_denoising}, that suggest that \tramp is able to use the trained VAE prior information to either denoise very noisy observations or reconstruct missing pixels.

\subsection{Theoretical prediction of performance}

Previous sections were devoted to applications of the Expectation Propagation (EP) Algorithm~\ref{algo:gaussian_EP} implemented in \tramp. Moreover the state evolution (SE) Algorithm~\ref{algo:SE_RS_matched} has also been implemented in the package. This two-in-one package makes it easier to obtain performances of the EP algorithm on finite size instances as well as the infinite size limit behavior predicted by the state evolution. 

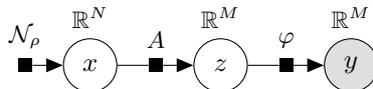
\begin{figure}[H]
  \centering
  \begin{tikzpicture}
    \node[latent, label=$\bbR^{N}$] (x)  {$x$} ;
    \node[latent, right=of x, label=$\bbR^{M}$] (z) {$z$} ;
    \node[obs, right=of z, label=$\bbR^{M}$] (y) {$y$} ;
    \factor[left=of x] {p} {$\mN_\rho$} {} {x} ;
    \factor[left=of z] {p} {$A$} {x} {z} ;
    \factor[left=of y] {p} {$\varphi$} {z} {y} ;
  \end{tikzpicture}
  \caption{Graphical model representing the compressed sensing ($\varphi(x)=x)$) and phase retrieval ($\varphi(x)=|x|)$). We denote $\alpha=M/N$ the aspect ratio of the sensing matrix $A$.}
  \label{fig:graph_cs_pr_sparse}
\end{figure}

We illustrate this on two generalized linear models: \emph{compressed sensing} and \emph{sparse phase retrieval}, whose common factor graph is represented in Figure~\ref{fig:graph_cs_pr_sparse}. Briefly, we consider a sparse $x \in \bbR^{N}$ $\iid$ drawn from a Gauss-Bernoulli distribution $\mathcal{N}_\rho$. We observe $y\in \bbR^{M}=\varphi(Ax)$ with $A\in\bbR^{M\times N}$ a Gaussian $\iid$ matrix, and the noiseless channel is $\varphi(x)=x$ in the compressed sensing case and $\varphi(x)=|x|$ in the phase retrieval one.

\begin{figure}[t]
            \includegraphics[width=0.45\linewidth]{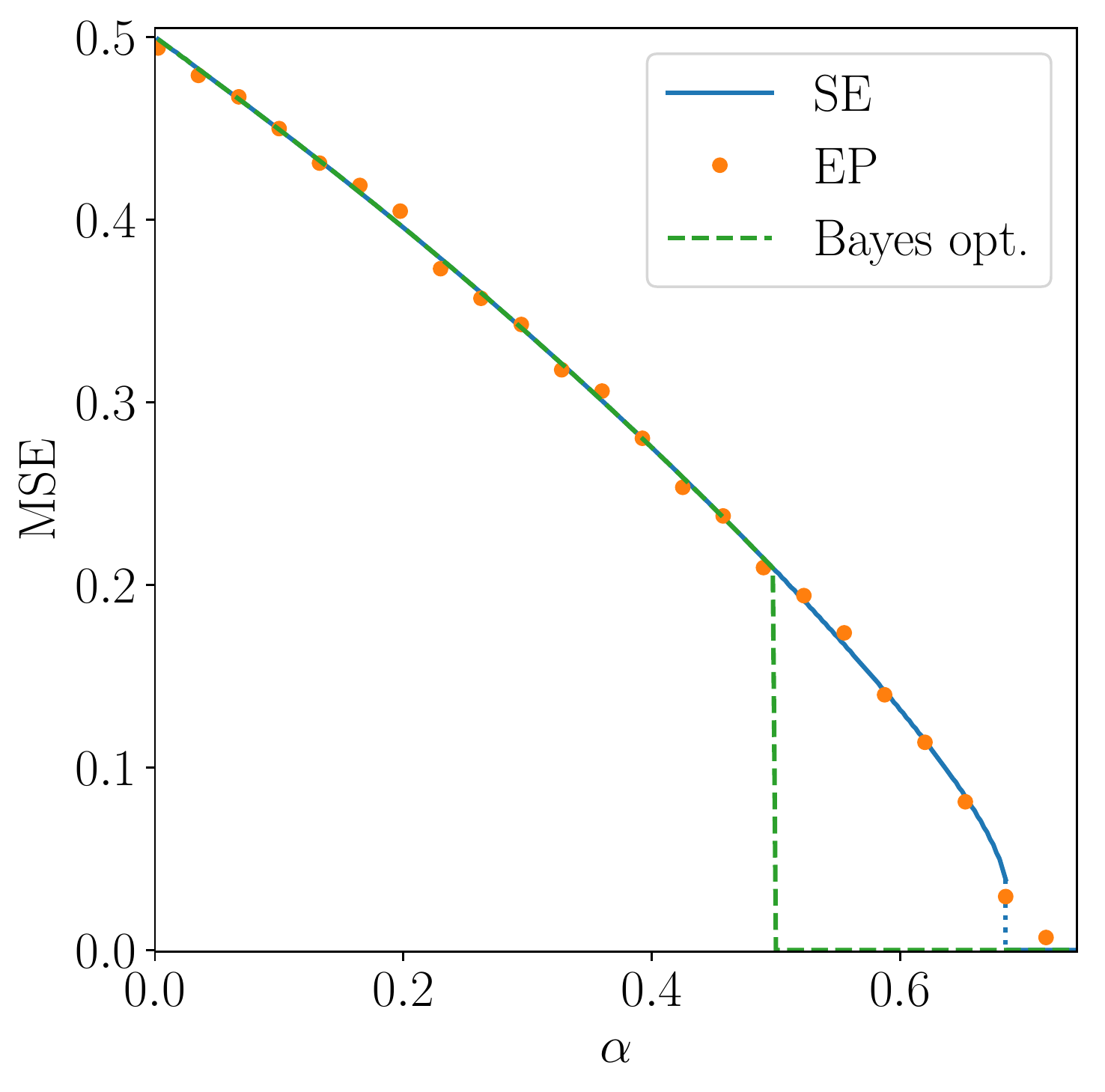}
            \includegraphics[width=0.45\linewidth]{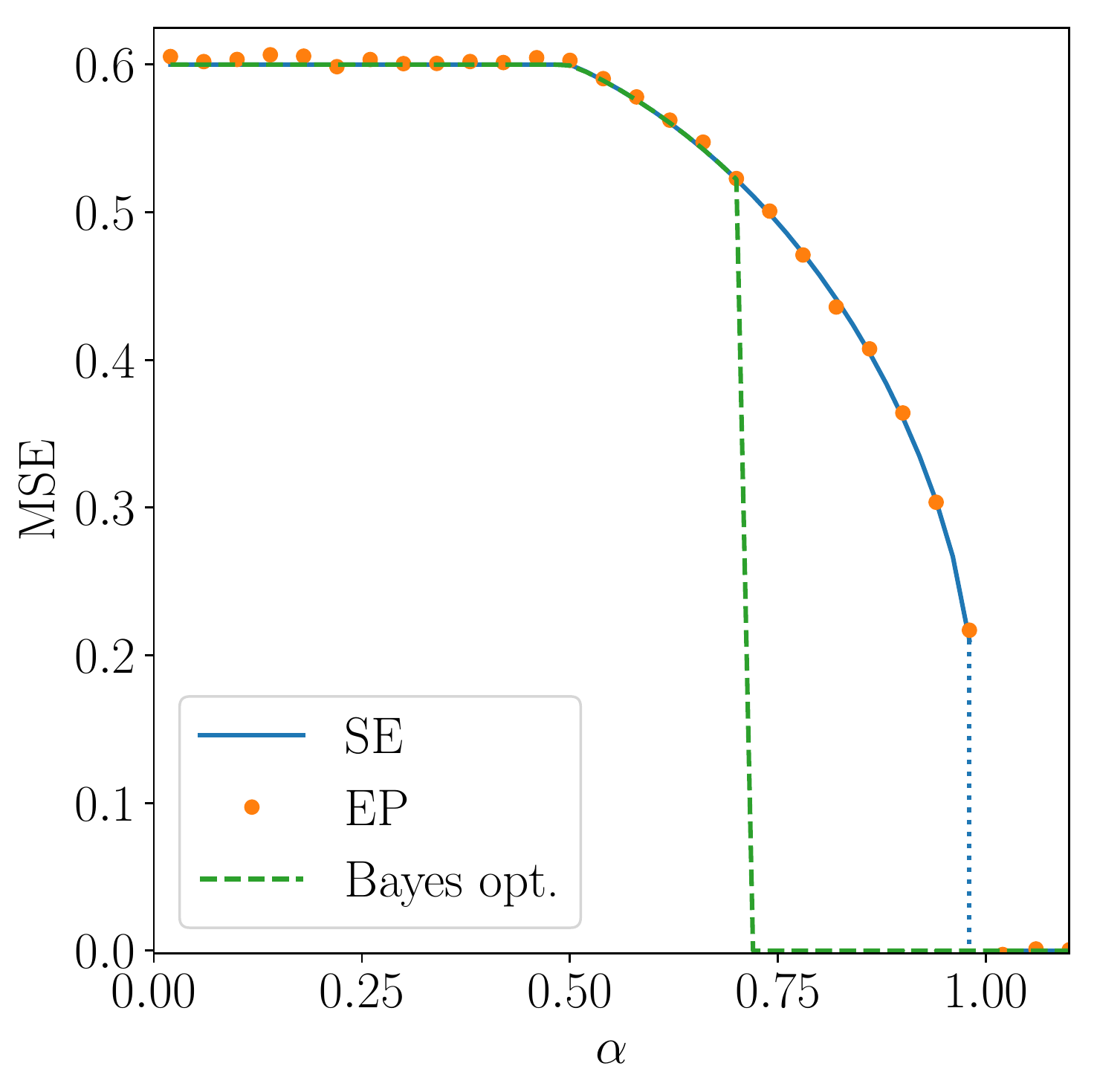}
            \caption{MSE as a function of $\alpha =  M/N$ for (\textbf{left}) Compressed sensing ($\rho=0.5$), (\textbf{right}) Sparse phase retrieval ($\rho=0.6$).}
            \label{fig:cs_pr_sparse}
\end{figure}

Getting the MSE predicted by state evolution is straightforward in the \tramp package. After importing the relevant modules, one just needs to declare the model and run the SE algorithm. For instance for the 
sparse phase retrieval model:

\begin{minipage}[c]{\linewidth}
\begin{lstlisting}[language=Python]
# declare sparse phase retrieval model
model = (
    GaussBernoulliPrior(rho=rho, size=N) @ V('x') @
    LinearChannel(A) @ V('z') @ 
    AbsLikelihood(y=y)
).to_model()
# run SE
from tramp.algos import StateEvolution
se = StateEvolution(model)
se.iterate(max_iter=200)
\end{lstlisting}
\end{minipage}
In Figure~\ref{fig:cs_pr_sparse}, we compare the MSE theoretically predicted by state evolution and the MSE obtained on high-dimensional ($N=2000$) instances of EP.
Notably up to finite size effects, the MSE averaged over 25 instances of EP match perfectly the MSE predicted by SE. Also the MSE is equal to the Bayes optimal MMSE proven in \citep{Barbier2019}, except for a region of $\alpha$ values known as the hard phase. In that phase, there is a significant gap between the MMSE that is information-theoretically achievable and the MSE actually achieved by EP. 

Note that the MMSE is also a state evolution fixed point (Section~\ref{sec:bayes_optimal}) and can thus be obtained by initializing the SE Algorithm~\ref{algo:SE_RS_matched} in the right basin of attraction. For the two models discussed here, we found that initializing the incoming prior message $\hat{m}_{x \rightarrow \mathcal{N}_\rho} \gg 1$ was sufficient to converge towards the MMSE and obtain the Bayes optimal curve.

\section{Discussion}

\paragraph{Modularity} 
The \tramp package aims to solve compositional inference tasks, which can be broken down into local inference problems. As long as the underlying factor graph is tree-structured, the global inference task can be solved by message passing using Algorithm~\ref{algo:gaussian_EP}, which is just a particular instance of EP. Note that Algorithm~\ref{algo:gaussian_EP} is generic, meaning that it can be implemented independently of the probabilistic graphical model under consideration.
The main strength of the presented approach is therefore its modularity. In the \tramp package, each module corresponds to a local inference problem given by a factor and associated beliefs on its variables. As long as the module is implemented (which means computing the log-partition $A_f[\lambda_f]$ and the moment function $\mu_f[\lambda_f]$), it can be composed at will with other modules to solve complex inference tasks. Several popular machine learning tasks can be reformulated that way as illustrated in Figure~\ref{fig:tree_models}. We hope that the \tramp  package offers a unifying framework to run these models, as well as study them theoretically using the state evolution and free entropy formalism. Below, we review some shortcomings of the \tramp package and possible ways to overcome them. 

\paragraph{Hyper-parameter learning} In principle, it should be straightforward to learn hyper-parameters. As usually done in hierarchical Bayesian modelling, one simply needs to add the hyper-parameters as scalar variables in the graphical model with associated hyper priors. In term of the \tramp package, one would simply need to implement the corresponding module (where the set of variables of the factor now includes the hyper-parameters to learn). In the typical use case, where the signals are high dimensional but the hyper-parameters are just scalars, Algorithm~\ref{algo:gaussian_EP} will likely be equivalent to the Expectation-Maximization \citep{Dempster1977} learning of hyper-parameters, as usually done in AMP algorithms \citep{Krzakala2012}.

\paragraph{Generic belief} While the message passing Algorithm~\ref{algo:gaussian_EP} is formulated for any kind of beliefs, the current \tramp implementation only supports isotropic Gaussian beliefs.  However we could consider more generic beliefs to deal with more complicated types of variable, such as Gaussian process beliefs \citep{Rasmussen2006} for functions or harmonic exponential family beliefs \citep{Cohen2015} for elements of compact groups. Maybe one can recover algorithms similar to \citep{Opper2000} for Gaussian process classification or \citep{Perry2018} for synchronisation problems over compact groups, and reformulate them in a more modular way. Even if we restrict ourselves to Gaussian beliefs, it may be beneficial to go beyond the isotropic case and consider diagonal or full covariance beliefs \citep{Opper2005JMLR}, or any kind of prescribed covariance structure.
As exemplified in the committee machine keeping a covariance between experts leads to a more accurate algorithm \citep{Aubin2018}.

\paragraph{Beyond trees} By design, the \tramp package can only handle tree-structured factor graphs. To overcome this fundamental limitation and extend to generic factor graphs, 
one could use the Kikuchi free energy \citep{Yedidia2001} in place of the Bethe free energy as a starting point. Similar to the Bethe free energy which is exact for tree-structured factor graphs, the Kikuchi free energy will be exact if the graphical model admits a hyper-tree factorization \citep{Wainwright2008}. The minimization of the Kikuchi free energy under weak consistency constraints \citep{Zoeter2005} could be used to implement a generalization of Algorithm~\ref{algo:gaussian_EP}, however the message passing will be more challenging than in the tree case. An equivalent of Proposition~\ref{prop:free_energy} will likely hold, where the tree decomposition Eqs~\eqref{gibbs_free_energy}-\eqref{minka_free_energy} found for the EC Gibbs and EP free energies will be replaced by the hyper-tree factorization.  

\paragraph{Convergence} The message passing Algorithm~\ref{algo:gaussian_EP}, like other EP algorithms, is not guaranteed to converge. This is a major drawback, and indeed on some instances the naive application of Algorithm~\ref{algo:gaussian_EP} will  
diverge. Double loop algorithms like \citep{Heskes2002} will ensure convergence, but are  unfortunately very slow. In practice, damping the updates is often sufficient to converge towards a fixed point. In the \tramp package the amount of damping has to be chosen by the user. It will be therefore interesting to generalize the adaptive damping scheme \citep{Vila2015} in order to tune this damping automatically.

\paragraph{Proofs} In Section~\ref{sec:SE}, we heuristically derive the free entropy using weak consistency on the overlaps and conjecture the corresponding state evolution, but did not provide any rigorous proof. We nonetheless recovered earlier derivations of these results for specific models. For instance in the multi-layer model with orthogonally invariant weight matrices,
the state evolution was rigorously proven by \cite{Fletcher2018} while the free potential was heuristically derived using the replica method \citep{Gabrie2018}. When the weight matrices are Gaussian, the replica free entropy can further be shown to be rigorous \citep{Reeves2016, Barbier2019}. One extension of our work would be to generalize existing proofs to arbitrary tree-structured models. In particular it would be beneficial to determine under which conditions the overlaps are the relevant order parameters and if our weak consistency derivation is indeed asymptotically exact.

\acks{
This work is supported by the ERC under the European Union’s Horizon
2020 Research and Innovation Program 714608-SMiLe, as well as by the
French Agence Nationale de la Recherche under grant
ANR-17-CE23-0023-01 PAIL. 
Additional funding is acknowledged by AB from `Chaire de recherche sur les mod\`eles et sciences des donn\'ees', Fondation CFM pour la Recherche-ENS.
}

\newpage
\appendix
\section{Proof of Proposition~\ref{prop:free_energy} \label{app:free_energy}}

\subsection{Minimization of $G[\mu_V]$}

Let's first minimize the Bethe free energy at fixed moments $\mu_V =  ( \mu_i )_{i \in V}$:
\begin{equation}
  F_{\mu_V} = \min_{ (\tilde{p}_F,\tilde{p}_V) \in \mathcal{M}_{\mu_V}} \mathcal{F}_\text{Bethe}[\tilde{p}_F,\tilde{p}_V]
\end{equation}
where $\mathcal{M}_{\mu_V}$ is the set of factor and variable marginals
at fixed moment:
\begin{equation}
  \mathcal{M}_{\mu_V} = \left\{
   (\tilde{p}_F,\tilde{p}_V) \quad : \quad
    \forall (i,k) \in E, \quad
    \mathbb{E}_{\tilde{p}_i} \phi_i(x_i) =
    \mathbb{E}_{\tilde{p}_k} \phi_i(x_i) =
    \mu_i
  \right\}.
\end{equation}
The solution will be a stationary point of the Lagrangian
\begin{equation}
  \begin{split}
  &\mathcal{L}[\tilde{p}_F, \tilde{p}_V, \lambda_F, \lambda_V] =
  \mathcal{F}_\text{Bethe}[\tilde{p}_F,\tilde{p}_V] \\
  &+ \sum_{k \in F} \sum_{i \in \partial k}
  \langle \lambda_{i \to k} , \mu_i - \mathbb{E}_{\tilde{p}_k}\phi_i(x_i) \rangle
  +\sum_{i \in V} (1-n_i)
  \langle \lambda_i ,  \mu_i - \mathbb{E}_{\tilde{p}_i}\phi_i(x_i) \rangle
\end{split}
\end{equation}
with Lagrange multipliers $\lambda_{i \to k}$ and $\lambda_i$
associated to the moment constraint $\mathbb{E}_{\tilde{p}_k} \phi_i(x_i) = \mu_i$ and
$\mathbb{E}_{\tilde{p}_i} \phi_i(x_i) = \mu_i$.
Then
$0 = \delta_{\tilde{p}_i} \mathcal{L} = \delta_{\tilde{p}_k} \mathcal{L} = \partial_{\lambda_{i \to k}} \mathcal{L} = \partial_{\lambda_i} \mathcal{L}$
leads to the solution:
\begin{align}
    \label{exp_fam_factor}    
  &\tilde{p}_k(x_k) = p_k(x_k \mid \lambda_k) = f(x_k;y_k)\, e^{\langle \lambda_k,  \phi_k(x_k) \rangle - A_k[\lambda_k]} , \\
   \label{exp_fam_variable}    
  &\tilde{p}_i(x_i) = p_i(x_i \mid \lambda_i) = e^{\langle \lambda_i,  \phi_i(x_i) \rangle - A_i[\lambda_i]} , \\
  &\mu_i = \mu_i[\lambda_i] = \mu_i^k[\lambda_k].
\end{align}
Besides the minimal $F_{\mu_V}$ is equal to:
\begin{align*}
  F_{\mu_V} &= \sum_{k \in F} F_k[\tilde{p}_k] + \sum_{i \in V} (1-n_i) F_i[\tilde{p}_i]\\
      &= \sum_{k \in F} \KL [\tilde{p}_k \Vert f_k] + \sum_{i \in V} (1-n_i) (-\mathrm{H}[\tilde{p}_i])\\
      &= \sum_{k \in F} \KL [p_k(x_k \mid \lambda_k)  \Vert f_k(x_k;y_k)] + \sum_{i \in V} (1-n_i) (-\mathrm{H}[p_i(x_i|\lambda_i)])\\
      &= \sum_{k \in F} G[\mu_k] + \sum_{i \in V} (1-n_i) G_i[\mu_i] \\
      &= G[\mu_V],
\end{align*}
due to the definition Eq.~\eqref{bethe_free_energy} of the Bethe free energy, the fact that the solutions $\tilde{p}_k$ and $\tilde{p}_i$ belong to the exponential families Eqs~\eqref{exp_fam_factor}-\eqref{exp_fam_variable}, and the definitions Eqs~\eqref{factor_legendre}-\eqref{variable_legendre} of $G_k$ and $G_i$. But then the relaxed Bethe variational problem Eq.~\eqref{relaxed_bethe_variational_pb} can be written as:
\begin{equation}
  F_\phi(\mathbf{y}) = \min_{\mu_V} \min_{ (\tilde{p}_F,\tilde{p}_V) \in \mathcal{M}_{\mu_V}} \mathcal{F}_\text{Bethe}[\tilde{p}_F,\tilde{p}_V]
  = \min_{\mu_V} F_{\mu_V} = \min_{\mu_V} G[\mu_V] .
\end{equation}
Besides minimizing $G[\mu_V]$ leads to:
\begin{equation}
  0 = \partial_{\mu_i} G[\mu_V]
    = \sum_{k \in \partial i} \lambda_{i \to k} + (1-n_i) \lambda_i
\end{equation}
which is the natural parameter constraint. The solution is therefore the same as
the EP fixed point Eq.~\eqref{ep_fixed_point}.

\subsection{Stationary point of  $A[\lambda_V, \lambda_F]$}
Using the
duality (Section~\ref{sec:exp_family_duality}) between natural parameters and moments:
\begin{align}
  \notag
  \min_{\mu_V} G[\mu_V] 
  &= \min_{\mu_V} \sum_{k \in F} G_k[\mu_k] + \sum_{i \in V} (1-n_i) G_i[\mu_i] \\
  \notag    
  &= \min_{\mu_V} \sum_{k \in F} \max_{\lambda_k} \left\{
  \langle \lambda_k , \mu_k \rangle - A_k[\lambda_k]
  \right\}
  + \sum_{i \in V} \underbrace{(1-n_i)}_{\leq 0 } \max_{\lambda_i} \left\{
  \langle \lambda_i ,  \mu_i \rangle - A_i[\lambda_i]
  \right\} \\
  \notag 
  &= \min_{\mu_V} \min_{\lambda_V} \max_{\lambda_F}  - \sum_{k \in F} A_k[\lambda_k]
  - \sum_{i \in V} (1-n_i) A_i[\lambda_i] \\
  \notag
  &+ \sum_{i \in V} \langle 
     \sum_{k \in \partial i} \lambda_{i \to k} + (1-n_i) \lambda_i, 
     \; \mu_i \rangle \\
  \label{minka_dual_pb}
  &= \min_{\lambda_V} \max_{\lambda_F} - A[\lambda_V, \lambda_F]
  \quad\text{s.t}\quad \forall i\in V:
   (n_i - 1) \lambda_i = \sum_{k \in \partial i} \lambda_{i \to k} .
\end{align}
As a consistency check, let's directly derive the solution to Eq.~\eqref{minka_dual_pb}. Consider the Lagrangian
\begin{equation}
  \mathcal{L}[\lambda_V, \lambda_F, \mu_V] =
  A[\lambda_V, \lambda_F] +
  \sum_{i \in V} \langle
    (n_i - 1) \lambda_i - \sum_{k \in \partial i} \lambda_{i \to k},
    \;\mu_i \rangle
\end{equation}
with Lagrangian multiplier $\mu_i$ associated to the 
constraint $(n_i - 1) \lambda_i = \sum_{k \in \partial i} \lambda_{i \to k}$.
At a stationary point:
\begin{align}
  \partial_{\mu_i} \mathcal{L} = 0 &\implies
  (n_i - 1) \lambda_i = \sum_{k \in \partial i} \lambda_{i \to k} \\
  \partial_{\lambda_{i \to k}} \mathcal{L} = 0 &\implies
  \mu_i = \mu_i^k[\lambda_k] \\
  \partial_{\lambda_i} \mathcal{L} = 0 &\implies
  \mu_i = \mu_i[\lambda_i]
\end{align}
which is exactly the same as the EP fixed point
Eq.~\eqref{ep_fixed_point}. Furthermore
the Lagrangian multiplier is the posterior moment $\mu_i$.

\subsection{Stationary point of  $A[\lambda_E]$}
Finally, the optimization
under constraint of $A[\lambda_V, \lambda_F]$ is equivalent to finding a stationary
point of $A[\lambda_E]$ without any constraint. Indeed:
\begin{align}
  &\partial_{\lambda_{k \to i}} A[\lambda_E] = 0 \implies
  \mu_i[\lambda_i^k] = \mu_i[\lambda_i] \implies
  \lambda_i^k = \lambda_i \\
  &\partial_{\lambda_{i \to k}} A[\lambda_E] = 0 \implies
  \mu_i^k[\lambda_k] = \mu_i[\lambda_i^k]
\end{align}
which implies the natural parameter constraint
\begin{equation}
  \sum_{k \in \partial i} \lambda_{i \to k}
  = \sum_{k \in \partial i} (\lambda_i^k -  \lambda_{k \to i} )
  = n_i \lambda_i - \lambda_i = (n_i - 1) \lambda_i
\end{equation}
and the moment matching $\mu_i^k[\lambda_k] = \mu_i[\lambda_i]$
defining the EP fixed point
Eq.~\eqref{ep_fixed_point}.

\newpage
\section{MAP estimation \label{app:map_estimation}}
           
To yield the maximum at posterior (MAP) estimate, we follow the derivation given in Appendix A.1 of \citep{Manoel2018} in the context of the TV-VAMP algorithm and generalize it to tree-structured models. As usual in statistical physics literature \citep{mezard2009information}  we introduce an inverse temperature $\beta$ and consider the distribution:
\begin{equation}
    p^{(\beta)}(\mathbf{x} \mid \mathbf{y}) = \frac{1}{Z^{(\beta)}(\mathbf{y})} \prod_{k \in F} f_k(x_k ; y_k)^\beta = \frac{1}{Z^{(\beta)}(\mathbf{y})} e^{ - \beta \sum_{k \in F} E_k(x_k ; y_k)}
\end{equation}
and take the limit $\beta \to \infty$ to make the distribution concentrate around its mode. Indeed by the Laplace method:
\begin{equation}
    F^{(\beta)}(\mathbf{y}) 
    = - \frac{1}{\beta} \ln Z^{(\beta)}(\mathbf{y}) 
    = - \frac{1}{\beta} \int d\mathbf{x} e^{-\beta E(\mathbf{y})} 
    \xrightarrow{\beta \to \infty} 
    F^{(\infty)}(\mathbf{y}) = \min_{\mathbf{x}} E(\mathbf{x}, \mathbf{y})
\end{equation}
where $E$ is the total energy of the system:
\begin{equation}
    E(\mathbf{x}, \mathbf{y}) = \sum_{k \in F} E_k(x_k ; y_k).
\end{equation}
The inference problem becomes an energy minimization problem, the minimizer $\mathbf{x}^* = \argmin_{\mathbf{x}} E(\mathbf{x}, \mathbf{y})$ is the MAP estimate and called the ground state in statistical physics \citep{mezard2009information}.  

\paragraph{Limit of the weak consistency derivation} At fixed $\beta$, the weak consistency approximation Proposition~\ref{prop:free_energy}, here expressed for isotopic Gaussian beliefs, gives:
\begin{equation}
    F_\phi^{(\beta)}( \mathbf{y}) = \min_{r_V^{(\beta)}, \tau_V^{(\beta)}} G^{(\beta)}[r_V^{(\beta)}, \tau_V^{(\beta)}] = \min \extr_{a_E^{(\beta)}, b_E^{(\beta)}} - A^{(\beta)}[a_E^{(\beta)}. b_E^{(\beta)}] 
\end{equation}
The limiting behavior can easily be found by scaling the natural parameters $a^{(\beta)} = \beta a$ and 
 $b^{(\beta)} = \beta b$. By the Laplace method, the factor log-partitions are given by:
\begin{align*}
A_k^{(\beta)}[a_k^{(\beta)}, b_k^{(\beta)}]  
&= \frac{1}{\beta} \ln \int dx_k \, f_k(x_k ; y_k)^{\beta} 
e^{- \frac{a_k^{(\beta)}}{2} \Vert x_k \Vert^2 + b_k^{(\beta)\intercal} x_k } \\ 
&= \frac{1}{\beta} \ln \int dx_k \, 
e^{- \beta \{ E_k(x_k ; y_k) + \frac{a_k}{2} \Vert x_k \Vert^2 - b_k^\intercal x_k \} } \\
&\xrightarrow{\beta \to \infty} A_k^{(\infty)}[a_k, b_k] = - \min_{x_k} E_k(x_k ; y_k , a_k , b_k) 
\end{align*}
where we introduce the tilted energy function:
\begin{equation}
E_k(x_k ; y_k, a_k, b_k) = E_k(x_k ; y_k) + \frac{a_k}{2} \Vert x_k \Vert^2 - b_k^\intercal x_k.
\end{equation}
The limiting factor log-partition is thus given by:
\begin{equation}
    \label{factor_A_map}
    A_k^{(\infty)}[a_k, b_k] = - E_k(x_k^* ; y_k, a_k, b_k)   
    \quad \text{with} \quad x_k^* = \argmin_{x_k} E_k(x_k^* ; y_k, a_k, b_k)  
\end{equation}
which can be alternatively expressed as:
\begin{equation}
    A_k^{(\infty)}[a_k, b_k] =  \frac{\Vert b_k\Vert^2}{2a_k} 
    - \mathcal{M}_{\tfrac{1}{a_k}E_k(. ; y_k)}\left( \frac{b_k}{a_k} \right)  ,
    \quad 
    x_k^* = \mathrm{prox}_{\tfrac{1}{a_k}E_k(.;y_k)}\left( \frac{b_k}{a_k} \right) 
\end{equation}
using the Moreau envelop 
$\mathcal{M}_g(y) = \min_x \{g(x) + \tfrac{1}{2}\Vert x - y \Vert^2\}$ 
and proximal operator
$\mathrm{prox}_g(y) = \argmin_x \{g(x) + \tfrac{1}{2}\Vert x - y \Vert^2\} $.
The moment / natural parameter duality becomes:
\begin{align*}
    r_i^{k(\beta)}[a_k^{(\beta)}, b_k^{(\beta)}] &= 
    \partial_{b_{i \to k}^{(\beta)}} \beta A_k^{(\beta)}[a_k^{(\beta)}, b_k^{(\beta)}]
    \quad \xrightarrow{\beta \to \infty}
     & r_i^{k(\infty)}[a_k, b_k] &=  \partial_{b_{i \to k}} A_k^{(\infty)}[a_k, b_k] \\
     -\frac{N_i}{2} \tau_i^{k(\beta)}[a_k^{(\beta)}, b_k^{(\beta)}] &= 
     \partial_{a_{i \to k}^{(\beta)}} \beta A_k^{(\beta)}[a_k^{(\beta)}, b_k^{(\beta)}] 
    \quad \xrightarrow{\beta \to \infty}
     &-\frac{N_i}{2} \tau_i^{k(\infty)}[a_k, b_k] &= \partial_{a_{i \to k}} A_k^{(\beta)}[a_k, b_k]
\end{align*}
As the precisions are scaled as $a^{(\beta)} = \beta a$, the variances must be scaled as $\beta v^{(\beta)} = v^{(\infty)}$ and:
\begin{equation*}
     \beta v_i^{k(\beta)}[a_k^{(\beta)}, b_k^{(\beta)}] = 
     \beta  \langle 
     \partial^2_{b_{i \to k}^{(\beta)}} \beta A_k^{(\beta)}[a_k^{(\beta)}, b_k^{(\beta)}]
     \rangle
    \quad \xrightarrow{\beta \to \infty} \quad
     v_i^{k\infty}[a_k, b_k] =  \langle \partial^2_{b_{i \to k}} A_k^{(\infty)}[a_k, b_k] \rangle.
\end{equation*}
where $\langle \cdot \rangle$ denotes the average over components of $x_i$.
From Eq.~\eqref{factor_A_map}, we then get the posterior mean and second moment:
\begin{equation*}
    r_i^{k(\infty)}[a_k, b_k]  = \partial_{b_{i \to k}} A_k^{(\infty)}[a_k, b_k] = x_i^{k*}, \quad \tau_i^{k(\infty)}[a_k, b_k] = -\frac{2}{N_i} \partial_{a_{i \to k}} A_k^{(\infty)}[a_k, b_k]
    = \frac{\Vert x_i^{k*} \Vert^2 }{N_i}
\end{equation*}
in agreement with the concentration at the mode. Note that in the zero temperature limit the second-moment is the self-overlap $\tau_i^{(\infty)} = q_i^{(\infty)} = \frac{\Vert r_i^{(\infty)} \Vert^2}{N_i} \neq q_i^{(\infty)} + v_i^{(\infty)}$. Indeed $ \beta v_i^{(\beta)} \xrightarrow{\beta \to \infty} v_i^{(\infty)} $  is the \emph{scaled} variance, the actual variance $v_i^{(\beta)} \xrightarrow{\beta \to \infty} 0$. The variable log-partition corresponds to the $E_i(x_i) = 0$ special case. The variable log-partition, mean and (scaled) variance are explicitly given by: 
\begin{equation}
    A_i^{(\infty)}[a_i, b_i] = \frac{\Vert b_i \Vert^2}{2a_i}, \quad
    r_i^{(\infty)}[a_i, b_i] = \partial_{b_i} A_i^{(\infty)} = \frac{b_i}{a_i}, \quad
    v_i^{(\infty)}[a_i, b_i] = \langle \partial^2_{b_i} A_i^{(\infty)} \rangle = \frac{1}{a_i}.
\end{equation}

\paragraph{Limit of the EP algorithm} Let us denote $\text{EP}^{(\beta)}$ the instance of the EP Algorithm~\ref{algo:gaussian_EP} running with $a^{(\beta)}, b^{(\beta)}$ and  $r^{(\beta)}, v^{(\beta)}$, which can be used to search a stationary point of $A^{(\beta)}[a_E^{(\beta)}, b_E^{(\beta)}]$. Then 
we have the well defined limit $\text{EP}^{(\beta)}  \xrightarrow{\beta \to \infty} \text{EP}^{(\infty)}$ where $\text{EP}^{(\infty)}$ is the instance of the EP Algorithm~\ref{algo:gaussian_EP} running with $a, b$ and  $r^{(\infty)}, v^{(\infty)}$, which can be used to search a stationary point of $A^{(\infty)}[a_E, b_E]$. At an EP fixed point we will have 
\begin{equation}
    x_i^* = r_i = r_i^k, \quad v_i = v_i^k, \quad
     (n_i - 1) a_i = \sum_{k \in \partial i} a_{i \to k}, \quad 
     (n_i - 1) b_i = \sum_{k \in \partial i} b_{i \to k}, 
\end{equation}
and therefore:
\begin{align*}
    - A^{(\infty)}[a_E, b_E] 
    &= - \sum_k A_k^{(\infty)}[a_k, b_k] - \sum_{i \in V} (1 - n_i) A_i^{(\infty)}[a_i, b_i] \\
    &= \sum_k E_k(x_k^* ; y_k, a_k, b_k) + \sum_{i \in V} (1 - n_i) E_i(x_i^* ; a_i, b_i) \\
    &= \sum_k E_k(x_k^* ; y_k) \\ 
    &+ \sum_{i \in V} 
    -\frac{\Vert x_i^* \Vert^2}{2} 
    \underbrace{\left( \sum_{k \in \partial i} a_{i \to k} + (1 - n_i) a_i \right)}_{0}
    +x_i^{*\intercal} 
      \underbrace{\left( \sum_{k \in \partial i} b_{i \to k} + (1 - n_i) b_i \right)}_{0} \\
     &= E(\mathbf{x^*}, \mathbf{y}).
\end{align*}

\paragraph{Summary} To recap, the energy minimization / MAP estimation problem can be formulated as:
\begin{equation}
    F_\phi^{(\infty)}( \mathbf{y}) = \min_{\mathbf{x}} E(\mathbf{x}, \mathbf{y}) = \min \extr_{a_E, b_E} - A^{(\infty)}[a_E, b_E] 
\end{equation}
which is the $\beta \to \infty$ limit of Proposition~\ref{prop:free_energy}. A stationary point of $A^{(\infty)}[a_E, b_E]$ can be searched by the $\text{EP}^{(\infty)}$ Algorithm, which is just Algorithm~\ref{algo:gaussian_EP} using the MAP modules:
\begin{align}
    \notag
    &A_k^{(\infty)}[a_k, b_k] =  \frac{\Vert b_k\Vert^2}{2a_k} 
    - \mathcal{M}_{\tfrac{1}{a_k}E_k(. ; y_k)}\left( \frac{b_k}{a_k} \right) , \\
    &r_k^{(\infty)}[a_k, b_k] = \mathrm{prox}_{\tfrac{1}{a_k}E_k(.;y_k)}\left( \frac{b_k}{a_k} \right) , \quad
    v_k^{(\infty)}[a_k, b_k] = \langle \partial_{b_k} r_k^{(\infty)}[a_k, b_k] \rangle , \\
    &A_i^{(\infty)}[a_i, b_i] = \frac{\Vert b_i \Vert^2}{2a_i}, \quad
    r_i^{(\infty)} = \partial_{b_i} A_i^{(\infty)} = \frac{b_i}{a_i}, \quad
    v_i^{(\infty)} = \partial^2_{b_i} A_i^{(\infty)} = \frac{1}{a_i}.
\end{align}

\newpage
\section{Proof of Proposition~\ref{prop:RS_free_entropy} \label{app:RS_free_entropy}}

\subsection{Decomposition of $A(n)$}

We recall from Eq.~\eqref{A_decomposition} that we can formally decompose the SCGF $A(n)$ as:
\begin{equation}
\label{eq:A_decomposition}
A(n) = A_N^{(n)} - A_N^{(0)},
\quad A_N^{(n)} = \frac{1}{N} \ln Z_N^{(n)},
\quad A_N^{(0)} = \frac{1}{N} \ln Z_N^{(0)},
\end{equation}
where $Z_N^{(0)}$ is the partition function of the teacher generative model:
\begin{equation}
\label{eq:teacher}
 p^{(0)}(\mathbf{x}^{(0)} , \mathbf{y}) =
 \frac{1}{Z_N^{(0)}} \prod_{k \in F} f_k^{(0)}(x_k^{(0)} ; y_k),
\end{equation}
and $Z_N^{(n)}$ is the partition function of the replicated system:
\begin{equation}
    p^{(n)}(\{ \mathbf{x}^{(a)} \}_{a=0}^n, \mathbf{y}) = \frac{1}{Z_N^{(n)}}
    \prod_{k \in F} \left\{
    f_k^{(0)}(x_k^{(0)}; y_k) \prod_{a=1}^n f_k(x_k^{(a)}; y_k)
    \right\}
\end{equation}
where $x^{(a)}$ for $a=1\cdots n$ denote the $n$ replicas and $x^{(0)}$ the ground truth. Now the key observation is that both $Z_N^{(0)}$ and $Z_N^{(n)}$ are partition functions associated to a tree-structured model (and this is actually the same tree). In that case we know that the log-partition can be obtained exactly from  the Bethe variational problem which is unfortunately intractable. However following \cite{Heskes2005} we can solve the relaxed Bethe variational problem by enforcing weak consistency (moment matching) instead of the full consistency of the marginals,  as reviewed in Section~\ref{sec:EP} for the derivation of the EP algorithm.

We must nonetheless specify for which sufficient statistics we want to enforce moment-matching, which in that setting we interpret as identifying the relevant order parameters for the thermodynamic limit. In the following, to make the connection with previously derived replica formulas, we will assume that the relevant order parameters are the overlaps:
\begin{equation}
\label{eq:overlaps}
 \phi(\mathbf{x}) = \left( \frac{x_i^{(a)}\cdot x_i^{(b)}}{N_i}\right)_{i \in V, 0\leq a \leq b \leq n}
\end{equation}
In some settings other order parameters could be relevant, fortunately the following derivation could be easily extended to these cases.

\subsection{Weak consistency derivation of $A_N^{(n)}$}

Solving the relaxed Bethe variational problem applied to the log-partition $A_N^{(n)} = \frac{1}{N} \ln Z_N^{(n)}$, using the overlaps Eq.~\eqref{eq:overlaps} as sufficient statistics, leads to:
\begin{equation}
- A_N^{(n)}
=
\min_{Q_V} G^{(n)}[Q_V]
   = \min \extr_{\hat{Q}_E} - A^{(n)}[\hat{Q}_E].
\end{equation}
where the minimizer corresponds to the overlaps:
\begin{equation}
Q_V = ( Q_i^{(ab)} )_{i \in V, 0\leq a \leq b \leq n}, \quad
Q_i^{(ab)} = \EE \frac{x_i^{(a)}\cdot x_i^{(b)}}{N_i}
\end{equation}
with corresponding dual natural parameter messages: 
\begin{equation}
    \hat{Q}_E =
    (\hat{Q}_{i \to k}^{(ab)}, \hat{Q}_{k \to i}^{(ab)})_{(i,k)  \in E, 0\leq a \leq b \leq n}
\end{equation}
The potentials satisfy the tree decomposition:
\begin{align}
    &G^{(n)}[Q_V] = \sum_{k \in F} \alpha_k G_k^{(n)}[Q_k]
    + \sum_{i \in V} \alpha_i (1-n_i) G_i^{(n)}[Q_i] 
    \quad \text{with} \quad
    Q_k =  ( Q_i )_{i \in \partial k} ,\\
      \notag
    &A^{(n)}[\hat{Q}_E] =
    \sum_{k \in F} \alpha_k A_k^{(n)}[\hat{Q}_k]
    - \sum_{(i,k) \in E} \alpha_i A_i^{(n)}[\hat{Q}_i^{k}]
    + \sum_{i \in V} \alpha_i A_i^{(n)}[\hat{Q}_i] \\
    &\text{with} \quad
    \hat{Q}_k = ( \hat{Q}_{i \to k} )_{i \in \partial k}
    \quad
    \hat{Q}_i^{k} =  \hat{Q}_{i \to k} + \hat{Q}_{k \to i},
    \quad
    \hat{Q}_i =  \sum_{k \in \partial i}  \hat{Q}_{k \to i}.
\end{align}
The factor and variable log-partition are given by:
\begin{align}
    \notag
    &A_k^{(n)}[\hat{Q}_k] \\
    &= \frac{1}{N_k} \ln \int dy_k dx_k^{(0)}
    f_k^{(0)}(x_k^{(0)}; y_k) \left[ \prod_{a=1}^n dx_k^{(a)} f_k(x_k^{(a)}; y_k) \right]
    e^{
    \sum_{i \in \partial k}
    \sum_{0\leq a\leq b \leq n} \hat{Q}_{i \to k}^{(ab)}   x_i^{(a)} \cdot x_i^{(b)}
    } \\
    &A_i^{(n)}[\hat{Q}_i] = \frac{1}{N_i} \ln \int dx_i^{(0)} \left [\prod_{a=1}^n dx_i^{(a)} \right]
    \, e^{
    \sum_{0\leq a\leq b \leq n} \hat{Q}_i^{(ab)}   x_i^{(a)} \cdot x_i^{(b)}
    }
\end{align}
and their gradients give the dual mapping to the overlaps:
\begin{equation}
    \alpha_i^k Q_i^{k(ab)} =
    \partial_{\hat{Q}_{i \to k}^{(ab)}} A_k^{(n)},
    \quad
    Q_i^{(ab)} = \partial_{\hat{Q}_i^{(ab)}} A_i^{(n)} .
\end{equation}
$G_k$ and $G_i$ are the corresponding Legendre transforms. Any stationary point of the potentials (not necessarily the global optima) is a fixed point:
\begin{equation}
    Q_i^{k(ab)} = Q_i^{(ab)}, \quad \hat{Q}_i^{(ab)} = \sum_{k \in \partial i} \hat{Q}_{k \to i}^{(ab)}.
\end{equation}

\subsection{Weak consistency derivation of $A_N^{(0)}$}

The weak consistency derivation of $A_N^{(0)}$ is given in Proposition~\ref{prop:teacher_second_moments}
and consistently corresponds to the $n=0$ case (ignoring the replicas and only keeping the ground truth) of $A_N^{(n)}$. The teacher second moments and dual messages in $A_N^{(0)}$ correspond to the $a=b=0$ overlaps and dual messages in $A_N^{(n)}$:
\begin{equation}
    \tau_i^{(0)} =  Q_i^{(00)} = \EE \frac{\Vert x_i^{(0)} \Vert^2}{N_i} , \quad
    - \frac{1}{2} \hat{\tau}_{i \to k}^{(0)} =\hat{Q}_{i \to k}^{(00)}, \quad
    - \frac{1}{2} \hat{\tau}_{k \to i}^{(0)} =\hat{Q}_{k \to i}^{(00)}.
\end{equation}

\subsection{Weak consistency derivation of $A(n)$}

Now we can finally express the SCGF $A(n)$ in Eq.~\eqref{eq:A_decomposition} by subtracting the weak consistency derivation  $A_N^{(0)}$ to the weak consistency derivation of $A_N^{(n)}$. We assume that the teacher second moments $\tau_V^{(0)} = Q_V^{(00)}$ are known (Section~\ref{sec:teacher_second_moments}) and are now considered as fixed parameters. The SCGF obtained by weak consistency on the overlaps is given by:
\begin{equation}
- A(n)
=
\min_{Q_V^\ast} G(n)[Q_V^\ast]
   = \min \extr_{\hat{Q}_E^\ast} - A(n)[\hat{Q}_E^\ast].
\end{equation}
where the minimizer corresponds to the overlaps (with the $ab=00$ overlap omitted):
\begin{equation}
Q_V^\ast = ( Q_i^{(ab)} )_{i \in V, 0\leq a \leq b \leq n, ab\neq 00}, \quad
Q_i^{(ab)} = \EE \frac{x_i^{(a)}\cdot x_i^{(b)}}{N_i}
\end{equation}
with corresponding dual natural parameter messages: 
\begin{equation}
    \hat{Q}_E^\ast =
    (\hat{Q}_{i \to k}^{(ab)}, \hat{Q}_{k \to i}^{(ab)})_{(i,k)  \in E, 0\leq a \leq b \leq n, ab\neq 00}
\end{equation}
The potentials satisfy the tree decomposition:
\begin{align}
    \label{tree_decomposition_scgf}
    &G(n)[Q_V^\ast] = \sum_{k \in F} \alpha_k G_k(n)[Q_k^\ast]
    + \sum_{i \in V} \alpha_i (1-n_i) G_i(n)[Q_i^\ast] 
    \quad \text{with} \quad
    Q_k^\ast =  ( Q_i^\ast )_{i \in \partial k} ,\\
    \notag
    &A(n)[\hat{Q}_E^\ast] =
    \sum_{k \in F} \alpha_k A_k(n)[\hat{Q}_k^\ast]
    - \sum_{(i,k) \in E} \alpha_i A_i(n)[\hat{Q}_i^{k\ast}]
    + \sum_{i \in V} \alpha_i A_i(n)[\hat{Q}_i^\ast] \\
    &\text{with}
    \quad
    \hat{Q}_k^\ast = ( \hat{Q}_{i \to k}^\ast )_{i \in \partial k}
    \quad
    \hat{Q}_i^{k\ast} =  \hat{Q}_{i \to k}^\ast + \hat{Q}_{k \to i}^\ast,
    \quad
    \hat{Q}_i^\ast =  \sum_{k \in \partial i}  \hat{Q}_{k \to i}^\ast.
\end{align}
The factor and variable log-partition are now given by:
\begin{align}
    \notag
    &A_k(n)[\hat{Q}_k^\ast] 
    = \frac{1}{N_k} \ln \int dy_k dx_k^{(0)}  \\
    \label{factor_A_n}
    &p_k^{(0)}(x_k^{(0)}, y_k \mid \hat{\tau}_k^{(0)}) \left[ \prod_{a=1}^n dx_k^{(a)} f_k(x_k^{(a)}; y_k) \right]
    e^{
    \sum_{i \in \partial k}
    \sum_{0\leq a\leq b \leq n, ab \neq 00} \hat{Q}_{i \to k}^{(ab)}   x_i^{(a)} \cdot x_i^{(b)}
    } \\
    &A_i(n)[\hat{Q}_i^\ast] = \frac{1}{N_i} \ln \int dx_i^{(0)}
    p_i^{(0)}(x_i^{(0)} \mid \hat{\tau}_i^{(0)}) 
    \left[ \prod_{a=1}^n dx_i^{(a)} \right]
    e^{
    \sum_{0\leq a\leq b \leq n, ab \neq 00} \hat{Q}_i^{(ab)}   x_i^{(a)} \cdot x_i^{(b)}
    }
\end{align}
and their gradients give the dual mapping to the overlaps:
\begin{equation}
    \alpha_i^k Q_i^{k(ab)} =
    \partial_{\hat{Q}_{i \to k}^{(ab)}} A_k^{(n)},
    \quad
    Q_i^{(ab)} = \partial_{\hat{Q}_i^{(ab)}} A_i^{(n)}.
\end{equation}
$G_k$ and $G_i$ are the corresponding Legendre transforms. Any stationary point of the potentials (not necessarily the global optima) is a fixed point:
\begin{equation}
    Q_i^{k(ab)} = Q_i^{(ab)}, \quad \hat{Q}_i^{(ab)} = \sum_{k \in \partial i} \hat{Q}_{k \to i}^{(ab)}.
\end{equation}

\begin{proof}
It follows straightforwardly from $\hat{Q}_k^{(00)} = - \frac{1}{2}\hat{\tau}_k^{(0)}$ and Eq.~\eqref{teacher_factor_marginal}:
\begin{align*}
 &A_k^{(n)}[\hat{Q}_k] - A_k^{(0)}[\hat{\tau}_k^{(0)}] 
 \\
&= \frac{1}{N_k} \ln \int dy_k dx_k^{(0)} 
\,
\underbrace{
\tfrac{1}{Z_k^{(0)}[\hat{\tau}_k^{(0)}]}
f_k^{(0)}(x_k^{(0)}; y_k)
e^{-\frac{1}{2} \sum_{i \in \partial k} \hat{\tau}_{i \to k}^{0}
\Vert x_i^{(0)} \Vert^2}
}_{
p_k^{(0)}(x_k^{(0)}, y_k \mid \hat{\tau}_k^{(0)})
} 
\left[ \prod_{a=1}^n dx_k^{(a)} f_k(x_k^{(a)}; y_k) \right] \\
&e^{
\sum_{i \in \partial k}
\sum_{0\leq a\leq b \leq n, ab\neq 00} \hat{Q}_{i \to k}^{(ab)}   x_i^{(a)} \cdot x_i^{(b)}
}\\
&= A_k(n)[\hat{Q}_k^\ast]
\end{align*}
Similarly it follows straightforwardly from $\hat{Q}_i^{(00)} = - \frac{1}{2}\hat{\tau}_i^{(0)}$ and Eq.~\eqref{teacher_variable_marginal}:
\begin{align*}
 &A_i^{(n)}[\hat{Q}_i] - A_i^{(0)}[\hat{\tau}_i^{(0)}] \\
 &= \frac{1}{N_i} \ln \int dx_i^{(0)}
\underbrace{
\tfrac{1}{Z_i^{(0)}[\hat{\tau}_i^{(0)}]}
e^{-\frac{1}{2} \hat{\tau}_i^{0}
\Vert x_i^{(0)} \Vert^2}
}_{
p_i^{(0)}(x_i^{(0)} \mid \hat{\tau}_i^{0})
}
\left[ \prod_{a=1}^n dx_i^{(a)} \right]
e^{
\sum_{0\leq a\leq b \leq n, ab\neq 00} \hat{Q}_i^{(ab)}   x_i^{(a)} \cdot x_i^{(b)}
} \\
&= A_i(n)[\hat{Q}_i^\ast]
\end{align*}
\end{proof}

 \subsection{Replica symmetric $A(n)$}

We take the replica symmetric ansatz \citep{mezard1987spin} for the overlaps $(1\leq a < b \leq n)$:
 \begin{equation}
     \tau = Q^{(aa)},
     \;
     m = Q^{(0a)},
     \;
     q = Q^{(ab)},
     \quad
     -\frac{1}{2}\hat{\tau} = \hat{Q}^{(aa)},
      \;
     \hat{m} = \hat{Q}^{(0a)},
     \;
     \hat{q} = \hat{Q}^{(ab)}.
 \end{equation}
The replica symmetric SCGF $A(n)$ is given by:
\begin{equation}
- A(n)
=
\min_{m_V, q_V, \tau_V} G(n)[m_V, q_V, \tau_V]
   = \min \extr_{\hat{m}_E, \hat{q}_E, \hat{\tau}_E} - A(n)[\hat{m}_E, \hat{q}_E, \hat{\tau}_E].
\end{equation}
The potentials satisfy the tree decomposition:
\begin{align}
    \notag
    &G(n)[m_V, q_V, \tau_V] = \sum_{k \in F} \alpha_k G_k(n)[m_k, q_k, \tau_k]
    + \sum_{i \in V} \alpha_i (1-n_i) G_i(n)[m_i, q_i, \tau_i] \\
    &\text{with} \quad m_k = (m_i)_{i \in \partial k} 
    \quad (\text{idem }q,\tau), \\
    \notag
    &A(n)[\hat{m}_E, \hat{q}_E, \hat{\tau}_E] \\
    \notag
    &= \sum_{k \in F} \alpha_k A_k(n)[\hat{m}_k, \hat{q}_k, \hat{\tau}_k]
    - \sum_{(i,k) \in E} \alpha_i A_i(n)[\hat{m}_i^k, \hat{q}_i^k, \hat{\tau}_i^k]
    + \sum_{i \in V} \alpha_i A_i(n)[\hat{m}_i, \hat{q}_i, \hat{\tau}_i] \\
    &\text{with} \quad
    \hat{m}_k = (\hat{m}_{i \to k})_{i \in \partial k},
    \quad
    \hat{m}_i^k =  \hat{m}_{i \to k} + \hat{m}_{k \to i},
    \quad
    \hat{m}_i =  \sum_{k \in \partial i}  \hat{m}_{k \to i} \quad (\text{idem }q,\tau).
\end{align}
The factor and variable log-partition are given by:
\begin{align}
    &A_k(n)[\hat{m}_k, \hat{q}_k, \hat{\tau}_k]
    = \frac{1}{N_k} \EE_{p_k^{(0)}(x_k^{(0)}, y_k, b_k)} e^{N_k n A_k[a_k, b_k ; y_k]} \\
    &A_i(n)[\hat{m}_i, \hat{q}_i, \hat{\tau}_i]
    = \frac{1}{N_i} \EE_{p_i^{(0)}(x_i^{(0)}, b_i)} e^{N_i n A_i[a_i, b_i]}
\end{align}
taken with:
\begin{align}
    p_k^{(0)}(x_k^{(0)}, y_k, b_k) &=
    \mathcal{N}(b_k \mid \hat{m}_k x_k^{(0)}, \hat{q}_k) \;
    p_k^{(0)}(x_k^{(0)}, y_k ; \hat{\tau}_k^{(0)})
    \quad &\text{and} \quad
    a_k &= \hat{\tau}_k + \hat{q}_k, \\
    p_i^{(0)}(x_i^{(0)}, b_i) &=
    \mathcal{N}(b_i \mid \hat{m}_i x_i^{(0)}, \hat{q}_i) \;
    p_i^{(0)}(x_i^{(0)} ; \hat{\tau}_i^{(0)})
    \quad &\text{and} \quad
     a_i &= \hat{\tau}_i + \hat{q}_i,
\end{align}
and where $A_k[a_k, b_k ; y_k]$ and $A_i[a_i, b_i]$ are the scaled EP log-partitions:
\begin{align}
    A_k[a_k,b_k;y_k] &= \frac{1}{N_k} \ln \int dx_k \, f(x_k ; y_k) \, e^{-\frac{1}{2} a_k \Vert x_k \Vert^2 + b_k^\intercal x_k} \\
    A_i[a_i,b_i] &= \frac{1}{N_i} \ln \int dx_i \, e^{-\frac{1}{2} a_i \Vert x_i \Vert^2 + b_i^\intercal x_i} .
\end{align}
We have therefore the following large deviation theory interpretation:
\begin{align}
    \label{LDT_An}
    A(n) &= \text{SCGF of }A_N(\mathbf{y})\\
    \label{LDT_Ak}
    A_k(n) &= \text{SCGF of }A_k[a_k, b_k; y_k] \\
    \label{LDT_Ai}
    A_i(n) &= \text{SCGF of }A_i[a_i, b_i]
\end{align}

\begin{proof}
The replica symmetric ansatz gives:
\begin{align*}
  &\left. \sum_{0\leq a\leq b \leq n, ab \neq 00} \hat{Q}_{i \to k}^{(ab)}   x_i^{(a)} \cdot x_i^{(b)} \right\vert_\text{RS}\\
 &=
\hat{m}_{i \to k} x_i^{(0)} \cdot \sum_{a=1}^n x_i^{(a)} + 
\hat{q}_{i \to k} \sum_{1\leq a < b \leq n} x_i^{(a)} \cdot x_i^{(b)} 
-\frac{\hat{\tau}_{i \to k}}{2} \sum_{a=1}^n \Vert x_i^{(a)} \Vert^2 \\
&=  
\hat{m}_{i \to k} x_i^{(0)} \cdot \sum_{a=1}^n x_i^{(a)}
+ \frac{\hat{q}_{i \to k}}{2} \Vert \sum_{a=1}^n x_i^{(a)} \Vert^2 
- \frac{\hat{\tau}_{i \to k} + \hat{q}_{i \to k}}{2}  \sum_{a=1}^n 
\Vert x_i^{(a)} \Vert ^2
\end{align*}
Using the Gaussian identity:
\begin{equation*}
e^{\frac{\hat{q}_{i \to k}}{2} \Vert \sum_{a=1}^n x_i^{(a)} \Vert^2} = \int_{\mathbb{R}^{N_i}} d\xi_{i \to k} \mathcal{N}(\xi_{i \to k}) e^{\sqrt{\hat{q}_{i \to k}} \xi_{i \to k} \cdot \sum_{a=1}^n x_i^{(a)}}
\end{equation*}
we can express the replica symmetric $A_k(n)[\hat{Q}_k^\ast]$ as: 
\begin{align*}
&\left. A_k(n)[\hat{Q}_k^\ast] \right\vert_\text{RS}\\
    &=\frac{1}{N_k} \ln \int dy_k dx_k^{(0)} p_k^{(0)}(x_k^{(0)}, y_k \mid \hat{\tau}_k^{(0)}) 
    \left[ \prod_{i \in \partial k} d\xi_{i \to k} \mathcal{N}(\xi_{i \to k}) \right]
    \left[ \prod_{a=1}^n dx_k^{(a)} f_k(x_k^{(a)}; y_k) \right] \\ 
    &\prod_{a=1}^n e^{
    \sum_{i \in \partial k}
    [\hat{m}_{i \to k} x_i^{(0)} + \sqrt{\hat{q}_{i \to k}} \xi_{i \to k}] \cdot x_i^{(a)} 
    -\frac{\hat{\tau}_{i \to k} + \hat{q}_{i \to k}}{2} \Vert x_i^{(a)} \Vert^2 
    } \\
    &=\frac{1}{N_k} \ln \int dy_k dx_k^{(0)} p_k^{(0)}(x_k^{(0)}, y_k \mid \hat{\tau}_k^{(0)}) 
    \left[ \prod_{i \in \partial k} db_{i \to k} \mathcal{N}(b_{i \to k} \mid \hat{m}_{i \to k} x_i^{(0)}, \hat{q}_{i \to k}) \right] \\
    &\left[ \prod_{a=1}^n dx_k^{(a)} f_k(x_k^{(a)}; y_k) 
    e^{\sum_{i \in \partial k} 
    -\frac{a_{i \to k}}{2} \Vert x_i^{(a)} \Vert^2 
    + b_{i \to k} \cdot x_i^{(a)} 
    }
    \right] \\
    &\text{with} \quad
    b_{i \to k} = \hat{m}_{i \to k} x_i^{(0)} + \sqrt{\hat{q}_{i\to k}} \xi_{i \to k} , \quad 
    a_{i \to k} = \hat{\tau}_{i \to k} + \hat{q}_{i \to k} .
\end{align*}
The integration over $x_k^{(a)}$ gives the EP partition function
\begin{equation}
    Z_k[a_k, b_k ; y_k] = \int dx_k \, f(x_k ; y_k) \, e^{\sum_{i \in \partial k} -\frac{1}{2} a_{i \to k} \Vert x_i \Vert^2 + b_{i \to k} \cdot x_k}
\end{equation}
therefore
\begin{align*}
    &\left. A_k(n)[\hat{Q}_k^\ast] \right\vert_\text{RS} \\
    &=\frac{1}{N_k} \ln \int dy_k dx_k^{(0)} db_k 
    \mathcal{N}(b_k \mid \hat{m}_k x_k^{(0)}, \hat{q}_k) 
    p_k^{(0)}(x_k^{(0)}, y_k \mid \hat{\tau}_k^{(0)}) 
    \prod_{a=1}^n Z_k[a_k, b_k ; y_k] \\
    &=\frac{1}{N_k} \ln \EE_{p_k^{(0)}(x_k^{(0)},y_k,b_k)} 
    Z_k[a_k, b_k ; y_k]^n \\
    &=\frac{1}{N_k} \ln \EE_{p_k^{(0)}(x_k^{(0)},y_k,b_k)} 
    e^{N_k n A_k[a_k, b_k ; y_k]} 
    \quad \text{with} \quad A_k[a_k, b_k ; y_k] = \frac{1}{N_k} \ln Z_k[a_k,b_k;y_k] \\
    &= A_k(n)[\hat{m}_k, \hat{q}_k, \hat{\tau}_k]
\end{align*}
The proof is identical for the replica symmetric $A_i(n)[\hat{Q}_i^\ast]$.
\end{proof}

\subsection{Replica symmetric $\bar{A}$}

We recall from Eq.~\eqref{SCGF_derivative} that the ensemble average $\bar{A}$ can be obtained by:
\begin{equation}
\label{eq:SCGF_derivative}
\bar{A} = \left. \frac{d}{dn} A(n) \right \vert_{n=0}
\end{equation}
Taking the derivative Eq.~\eqref{eq:SCGF_derivative} of the replica symmetric $A(n)$ we get Proposition~\ref{prop:RS_free_entropy} with:
\begin{align*}
 &\left. \frac{d}{dn} A_k(n)[\hat{m}_k, \hat{q}_k, \hat{\tau_k}] \right \vert_{n=0} = \EE_{p_k^{(0)}(x_k^{(0)},y_k,b_k)} A_k[a_k, b_k; y_k] = \bar{A}_k[\hat{m}_k, \hat{q}_k, \hat{\tau_k}] \\
 &\left. \frac{d}{dn} A_i(n)[\hat{m}_i, \hat{q}_i, \hat{\tau_i}] \right \vert_{n=0} = \EE_{p_i^{(0)}(x_i^{(0)},b_i)} A_i[a_i, b_i] = \bar{A}_i[\hat{m}_i, \hat{q}_i, \hat{\tau_i}] \\
\end{align*}
as expected from the large deviation theory interpretation Eqs~\eqref{LDT_Ak}-\eqref{LDT_Ai}.
\newpage
\section{Proof of the information theoretic decomposition \label{app:proof_IT}}

We recall that the local teacher generative model Eq.~\eqref{local_teacher} is given by:
\begin{equation*}
    p_k^{(0)}(x_k^{(0)}, y_k, b_k) = \mathcal{N}(b_k \mid \hat{m}_k x_k^{(0)}, \hat{q}_k) 
    \, p_k^{(0)}(x_k^{(0)}, y_k \mid \hat{\tau}_k^{(0)}),
\end{equation*}
while the local student generative model Eq.~\eqref{local_student} is given by:
\begin{equation*}
    p_k(x_k, y_k, b_k) = \mathcal{N}(b_k \mid \hat{q}_k x_k, \hat{q}_k)  
   \, p_k(x_k, y_k \mid \hat{\tau}_k).
\end{equation*}
Interestingly by rescaling $b_k^{(0)} = \frac{\hat{m}_k}{\hat{q}_k} b_k$ the local teacher generative model is equal to the Bayes-optimal setting generative model Eq.~\eqref{local_BO}:
\begin{align*}
    p_k^{(0)}(x_k^{(0)}, y_k, b_k^{(0)}) = \mathcal{N}(b_k^{(0)} \mid \hat{m}_k^{(0)} x_k^{(0)}, \hat{m}_k^{(0)})
    \, p_k^{(0)}(x_k^{(0)}, y_k \mid \hat{\tau}_k^{(0)})
    \quad \text{with} \quad 
    \hat{m}_k^{(0)} = \frac{\hat{m}_k^2}{\hat{q}_k}.
\end{align*}
The RS potential Eq.~\eqref{RS_potential_f} is related to the cross-entropy between the local teacher evidence $p_k^{(0)}( y_k, b_k)$ and the local student evidence  $p_k( y_k, b_k)$:
\begin{align}
    \notag
    &\bar{A}_k[\hat{m}_k, \hat{q}_k, \hat{\tau}_k] - A_k[\hat{\tau}_k]  \\
    \label{RS_decompostion1}
    &= - \frac{1}{N_k} H[p_k^{(0)}(y_k, b_k) , p_k(y_k, b_k)] 
       + \sum_{i \in \partial k} \frac{\alpha_i^k \hat{m}_{i \to k}^{(0)} \tau_i^{(0)}}{2} 
       + \frac{\alpha_i^k}{2} \ln 2 \pi e \hat{q}_{i \to k} .
\end{align}
Similarly, the BO potential Eq.~\eqref{BO_potential_f} is related to the entropy of the local Bayes-optimal evidence $p_k^{(0)}( y_k, b_k^{(0)})$:
\begin{equation}
    \label{BO_decomposition1}
 \bar{A}_k^{(0)}[\hat{m}_k^{(0)}] - A_k^{(0)}[\hat{\tau}_k^{(0)}]  \\
   = - \frac{1}{N_k} H[p_k^{(0)}(y_k, b_k^{(0)})] 
       + \sum_{i \in \partial k} \frac{\alpha_i^k \hat{m}_{i \to k}^{(0)} \tau_i^{(0)}}{2} 
       + \frac{\alpha_i^k}{2} \ln 2 \pi e \hat{m}_{i \to k}^{(0)}.
\end{equation}

\begin{proof}
The local student evidence is equal to:
\begin{align*}
   p_k(y_k, b_k) 
    &= \int dx_k p_k(x_k, y_k, b_k) 
    = \int dx_k  p_k(x_k , y_k \mid \hat{\tau_k} ) \mathcal{N}(b_k \mid \hat{q}_k x_k, \hat{q}_k) \\
    &= \int dx_k f_k(x_k;y_k) e^{-\frac{1}{2}\hat{\tau}_k \Vert x_k \Vert^2 - N_k A_k[\hat{\tau_k}]}  e^{-\frac{1}{2}\hat{q}_k \Vert x_k \Vert^2 + b_k^\intercal x_k } \mathcal{N}(b_k \mid 0, \hat{q}_k) \\
    &= \int dx_k f_k(x_k;y_k) e^{-\frac{1}{2}a_k \Vert x_k \Vert^2 +  b_k^\intercal x_k -  N_k A_k[\hat{\tau_k}]} \mathcal{N}(b_k \mid 0, \hat{q}_k)  
    \quad \text{with} \quad a_k = \hat{\tau}_k + \hat{q}_k \\
    &= e^{N_k A_k[a_k,b_k ; y_k] - N_k A_k[\hat{\tau}_k] }  \mathcal{N}(b_k \mid 0, \hat{q}_k) 
\end{align*}
Then:
\begin{align*}
    &\bar{A}_k[\hat{m}_k, \hat{q}_k, \hat{\tau}_k] - A_k[\hat{\tau}_k]
    =\EE_{p_k^{(0)}(x_k^{(0)},y_k,b_k)} A_k[a_k,b_k ; y_k] - A_k[\hat{\tau}_k] \\
    &=\frac{1}{N_k} \EE_{p_k^{(0)}(x_k^{(0)},y_k,b_k)} \ln \frac{p_k(y_k, b_k) }{\mathcal{N}(b_k \mid 0, \hat{q}_k)} \\
    &= - \frac{1}{N_k} H[p_k^{(0)}(y_k, b_k) , p_k(y_k, b_k)] 
    + \frac{1}{N_k} \EE_{p_k^{(0)}(x_k^{(0)},y_k,b_k)} \sum_{i \in \partial k} \frac{\Vert b_{i \to k}\Vert^2}{2\hat{q}_{i \to k}} + \frac{N_i}{2} \ln 2\pi\hat{q}_{i \to k} \\
    &= - \frac{1}{N_k} H[p_k^{(0)}(y_k, b_k) , p_k(y_k, b_k)] 
    + \frac{1}{N_k} \sum_{i \in \partial k} \frac{N_i(
    \hat{m}_{i \to k}^2 \tau_i^{(0)} + \hat{q}_{i \to k}
    )}{2\hat{q}_{i \to k}} + \frac{N_i}{2} \ln 2\pi\hat{q}_{i \to k} \\
    &= - \frac{1}{N_k} H[p_k^{(0)}(y_k, b_k) , p_k(y_k, b_k)] 
    + \sum_{i \in \partial k} \frac{\alpha_i^k \hat{m}_{i \to k}^{(0)} \tau_i^{(0)}}{2} + \frac{\alpha_i^k}{2} \ln 2 \pi e \hat{q}_{i \to k}
\end{align*}
with $\hat{m}_{i \to k}^{(0)} = \frac{\hat{m}_{i \to k}^2}{\hat{q}_{i \to k}}$ and $\alpha_i^k = \frac{N_i}{N_k}$ which proves Eq.~\eqref{RS_decompostion1}.
The proof for the BO potential Eq.~\eqref{BO_decomposition1} is identical to the RS case
with the simplifications $\hat{m}_k^{(0)} = \hat{q}_k = \hat{m}_k$ and $\hat{\tau}_k = \hat{\tau}_k^{(0)}$.
\end{proof}

From Eq.~\eqref{RS_decompostion1} we recover the 
decomposition Eq.~\eqref{RS_potential_decomposition} for the RS potential:
\begin{equation*}
 \bar{A}_k[\hat{m}_k, \hat{q}_k, \hat{\tau}_k] - A_k[\hat{\tau}_k] 
 = \bar{A}_k^{(0)}[\hat{m}_k^{(0)}]  - A_k^{(0)}[\hat{\tau}_k^{(0)}]  
  - \frac{1}{N_k} \mathrm{KL}[p_k^{(0)}(y_k, b_k) \Vert p_k(y_k, b_k)]
\end{equation*}
From Eq.~\eqref{BO_decomposition1} we recover the decomposition Eq.~\eqref{BO_potential_decomposition} for the BO potential:
\begin{equation*}
\bar{A}_k^{(0)}[\hat{m}_k^{(0)}] - A_k^{(0)}[\hat{\tau}_k^{(0)}] =  \sum_{i \in \partial k} \frac{\alpha_i^k \hat{m}_{i \to k}^{(0)} \tau_i^{(0)}}{2} + \frac{1}{N_k} H[p_k^{(0)}(b_k^{(0)} \mid x_k^{(0)})]  - \frac{1}{N_k} H[p_k^{(0)}(y_k, b_k^{(0)})] .
\end{equation*}

\begin{proof}
The proof for the BO case is straightforward:
\begin{equation*}
p_k^{(0)}(b_k^{(0)} \mid x_k^{(0)}) =  \mathcal{N}(b_k^{(0)} \mid \hat{m}_k^{(0)} x_k^{(0)}, \hat{m}_k^{(0)})
\implies
\frac{1}{N_k} H[p_k^{(0)}(b_k^{(0)} \mid x_k^{(0)})] = \sum_{i \in \partial k} \frac{\alpha_i^k}{2} \ln 2\pi e \hat{m}_{i \to k}^{(0)}.
\end{equation*}
In the RS case, due to the rescaling $b_k^{(0)} = \frac{\hat{m}_k}{\hat{q}_k} b_k$ we have for the densities:
\begin{equation*}
    \frac{1}{N_k} \ln p_k^{(0)}(y_k, b_k^{(0)}) = \frac{1}{N_k} \ln p_k^{(0)}(y_k, b_k) + 
    \sum_{i \in \partial k} \alpha_i^k \ln \frac{\hat{q}_{i \to k}}{\hat{m}_{i \to k}} 
\end{equation*}
which gives for the entropies:
\begin{equation*}
    - \frac{1}{N_k} H[p_k^{(0)}(y_k, b_k^{(0)})]  = - \frac{1}{N_k} H[p_k^{(0)}(y_k, b_k)] 
    + \sum_{i \in \partial k} \alpha_i^k \ln \frac{\hat{q}_{i \to k}}{\hat{m}_{i \to k}} 
\end{equation*}
Then Eq.~\eqref{BO_decomposition1} at $\hat{m}_k^{(0)} = \frac{\hat{m}_k^2}{\hat{q}_k}$ can be written:
\begin{align*}
    &\bar{A}_k^{(0)}[\hat{m}_k^{(0)}]  - A_k^{(0)}[\hat{\tau}_k^{(0)}] \\
     &= - \frac{1}{N_k} H[p_k^{(0)}(y_k, b_k)] 
     + \sum_{i \in \partial k} \alpha_i^k \ln \frac{\hat{q}_{i \to k}}{\hat{m}_{i \to k}} 
     + \sum_{i \in \partial k} \frac{\alpha_i^k \hat{m}_{i \to k}^{(0)} \tau_i^{(0)}}{2} 
     + \frac{\alpha_i^k}{2} \ln 2 \pi e \frac{\hat{m}_{i \to k}^2}{\hat{q}_{i \to k}} \\
     &= - \frac{1}{N_k} H[p_k^{(0)}(y_k, b_k)] 
     + \sum_{i \in \partial k} \frac{\alpha_i^k \hat{m}_{i \to k}^{(0)} \tau_i^{(0)}}{2} 
     + \frac{\alpha_i^k}{2} \ln 2 \pi e \hat{q}_{i \to k}
\end{align*}
By using Eq.~\eqref{RS_decompostion1} we finally get:
\begin{align*}
     &\bar{A}_k[\hat{m}_k, \hat{q}_k, \hat{\tau}_k] - A_k[\hat{\tau}_k] \\
     &=  \bar{A}_k^{(0)}[\hat{m}_k^{(0)}]  - A_k^{(0)}[\hat{\tau}_k^{(0)}]  
    - \frac{1}{N_k} H[p_k^{(0)}(y_k, b_k) , p_k(y_k, b_k)]
        + \frac{1}{N_k} H[p_k^{(0)}(y_k, b_k)] \\
    &= \bar{A}_k^{(0)}[\hat{m}_k^{(0)}]  - A_k^{(0)}[\hat{\tau}_k^{(0)}]   - \frac{1}{N_k} \mathrm{KL}[p_k^{(0)}(y_k, b_k) \Vert p_k(y_k, b_k)]
\end{align*}
\end{proof}

\newpage
\section{\tramp modules \label{app:modules}}

\subsection{Variable \label{module:variable}}

The \tramp package only implements isotropic Gaussian beliefs, but
the variable log-partitions presented here will be useful to derive the factor
modules. 

\subsubsection{General variable}

An approximate belief, which we may as well call a variable type, is specified by 
the base space $X$ as well as the chosen set of sufficient statistics $\phi(x)$. Any variable type defines an exponential family distribution 
\begin{equation}
    p(x \mid \lambda) = e^{\langle \lambda ,  \phi(x) \rangle - A[\lambda]}
\end{equation}
indexed by the natural parameter $\lambda$. The family can be alternatively indexed by the moments $\mu = \mathbb{E}_{p(x \mid \lambda)} \phi(x)$. The log-partition 
\begin{equation}
    A[\lambda] = \ln \int_X dx e^{\lambda^\intercal  \phi(x)} 
\end{equation}
provides the bijective mapping between the natural parameters and the moments:
\begin{equation}
    \mu[\lambda] = \partial_\lambda A[\lambda].
\end{equation}
For all the variable types considered below, we will always have $x \in \phi(x)$ in the set of sufficient statistics. Its associated natural parameter $b \in \lambda$ is thus dual to the mean $r \in \mu$. The mean and variance are given by:
\begin{equation}
    r[\lambda] = \partial_b A[\lambda], \qquad
    v[\lambda] = \partial_b^2 A[\lambda],
\end{equation}
We list below the log-partition, mean and variance for several variable types, which correspond to well known exponential family distributions.

\subsubsection{Isotropic Gaussian variable}
For $x \in \mathbb{R}^N$, sufficient statistics $\phi(x) = (x, -\frac{1}{2}x^\intercal  x )$, natural parameters $\lambda = (b,a)$ with $b \in \mathbb{R}^N$ and scalar precision $a \in \mathbb{R}$.
\begin{align}
  &A[a,b] = \ln \int dx\, e^{-\frac{1}{2} ax^\intercal x + b^\intercal  x}
  = \frac{\Vert b \Vert^2}{2a} + \frac{N}{2} \ln\frac{2\pi}{a}, \\
  &r[a,b] =\frac{b}{a}, \qquad v[a,b] = \frac{1}{a} \in \mathbb{R}.
\end{align}
The corresponding exponential family is the isotropic multivariate Normal:
\begin{equation}
  p(x\mid a,b) = \mathcal{N}(x\mid r,v)  
  \quad\text{with}\quad 
  r=\frac{b}{a}, \quad  v=\frac{1}{a}.
\end{equation}

\subsubsection{Diagonal Gaussian variable}
For $x \in \mathbb{R}^N$, sufficient statistics 
$\phi(x) = (x, -\frac{1}{2} x^2 )$, natural parameters $\lambda = (b,a)$ with $b \in \mathbb{R}^N$ and diagonal precision $a \in \mathbb{R}^N$.
\begin{align}
  &A[a,b] = \ln \int dx\, e^{-\frac{1}{2} x^\intercal  a x + b^\intercal  x}
  = \sum_{n=1}^N \frac{b_n^2}{2a_n} + \frac{1}{2} \ln\frac{2\pi}{a_n}, \\
  &r[a,b] =\frac{b}{a}, \qquad v[a,b] = \frac{1}{a} \in \mathbb{R}^N.
\end{align}
The corresponding exponential family is the diagonal multivariate Normal:
\begin{equation}
  p(x\mid a,b) = \mathcal{N}(x\mid r,v)  
 \quad\text{with}\quad 
  r=\frac{b}{a}, \quad  v=\frac{1}{a}.
\end{equation}

\subsubsection{Full covariance Gaussian variable}
For $x \in \mathbb{R}^N$, sufficient statistics 
$\phi(x) = (x, -\frac{1}{2} xx^\intercal )$, natural parameters $\lambda = (b,a)$ with $b \in \mathbb{R}^N$ and matrix precision $a \in \mathbb{R}^{N\times N}$.
\begin{align}
  &A[a,b] = \ln \int dx\, e^{-\frac{1}{2} x^\intercal  a x  + b^\intercal  x}
  =\frac{1}{2} b^\intercal  a^{-1} b + \frac{1}{2} \ln \det 2\pi a^{-1}, \\
  &r[a,b] =\frac{b}{a}, \qquad \Sigma[a,b] = a^{-1} \in \mathbb{R}^{N\times N},
\end{align}
The corresponding exponential family is the full covariance multivariate Normal:
\begin{equation}
  p(x\mid a,b) = \mathcal{N}(x\mid r,\Sigma)  
  \quad\text{with}\quad 
  r=\frac{b}{a}, \quad  \Sigma=a^{-1}.
\end{equation}

\subsubsection{Real variable \label{module:real_variable}} 
For $x \in \mathbb{R}$, sufficient statistics $\phi(x) = ( x, -\frac{1}{2} x^2 )$, natural parameters $\lambda = (b,a)$.
\begin{align}
  &A[a,b] = \ln \int dx\, e^{-\frac{1}{2} ax^2 + bx}
  = \frac{b^2}{2a} + \frac{1}{2} \ln\frac{2\pi}{a}, \\
  &r[a,b] =\frac{b}{a}, \qquad v[a,b] = \frac{1}{a}.
\end{align}
The corresponding exponential family is the Normal:
\begin{equation}
  p(x\mid a,b) = \mathcal{N}(x\mid r,v)  
  \quad\text{with}\quad 
  r=\frac{b}{a}, \quad  v=\frac{1}{a}.
\end{equation}

\subsubsection{Binary variable \label{module:binary_variable}}  
For $x \in \pm$, sufficient statistics $\phi(x) = x $, natural parameter $\lambda = b$.
\begin{align}
  &A[b] = \ln \sum_{x = \pm} e^{bx} = \ln (e^{+b} + e^{-b}), \\
  &r[b] = \tanh(b), \qquad v[b] = \frac{1}{\cosh(b)^2}.
\end{align}
The corresponding exponential family is the Bernoulli (over $\pm$):
\begin{equation}
  p(x\mid b) = p_+ \delta_{+1}(x) + p_- \delta_{-1}(x)
\end{equation}
where the natural parameter $b = \frac{1}{2} \ln \frac{p_+}{p_-}$ is the log-odds.

\subsubsection{Sparse variable \label{module:sparse_variable}}  
 For $x \in \mathbb{R}\cup\{ 0\}$, 
sufficient statistics $\phi(x) =( x, -\frac{1}{2} x^2, \delta(x) )$, natural parameters $\lambda = (b, a, \eta)$. There is a finite probability that $x=0$. 
The natural parameter $\eta$ corresponding to the sufficient statistic $\delta(x)$ is dual to the fraction of zero elements $\kappa = \mathbb{E} \delta(x) = p(x=0)$. 
The sparsity $\rho = 1 - \kappa =  p(x \neq 0)$ is the fraction of non-zero elements.
 \begin{align}
   &A[a,b,\eta] = \ln\left[e^\eta + \int dx e^{-\frac{1}{2} ax^2 + bx}\right]
   = \eta + \ln(1 + e^\xi)  \quad \text{with} \quad \xi = A[a,b] - \eta,\\
   &r[a,b,\eta] = \frac{b}{a} \sigma(\xi), \qquad
   v[a,b,\eta] = \frac{1}{a} \sigma(\xi) + \frac{b^2}{a^2}  \sigma(\xi) \sigma(-\xi),
   \qquad
   \rho[a,b,\eta] = \sigma(\xi),
 \end{align}
where $\sigma$ is the sigmoid function and the parameter $\xi$ is the sparsity log-odds:
\begin{equation}
  \xi = A[a,b] - \eta = \ln \frac{\sigma(\xi)}{\sigma(-\xi)} = \ln \frac{\rho}{1-\rho}.
\end{equation}
The corresponding exponential family is the Gauss-Bernoulli:
\begin{equation}
  p(x\mid a,b,\eta) = [1-\rho]\delta(x)+\rho\mathcal{N}(x\mid r,v)
  \quad\text{with}\quad 
  r=\frac{b}{a}, \quad  v=\frac{1}{a}, \quad \rho = \rho[a,b,\eta].
\end{equation}

\subsubsection{Interval variable \label{module:interval_variable}} 
For $x \in X$, sufficient statistics $\phi(x) =( x, -\frac{1}{2} x^2 )$, natural parameters $\lambda = (b, a)$, where $X  = [x_\text{min} , x_\text{max} ]\subset \mathbb{R}$ is a real interval.
The probability that $x$ belongs to $X$ is equal to:
\begin{align}
  &p_X[a,b] = \int_X dx \mathcal{N}(x\mid r,v) = \Phi(z_\text{max}) - \Phi(z_\text{min}), \\
  &z_\text{min} = \frac{x_\text{min} - r}{\sqrt{v}} = \frac{ax_\text{min} -b}{\sqrt{a}}, \quad
  z_\text{max} = \frac{x_\text{max} - r}{\sqrt{v}}= \frac{ax_\text{max} -b}{\sqrt{a}},
\end{align}
where $\Phi$ is the cumulative Normal distribution and $z_\text{min}$ and $z_\text{max}$
are the z-scores of $x_\text{min}$ and $x_\text{max}$ for the Normal of mean $r=\frac{b}{a}$ and variance $v=\frac{1}{a}$. Then:
\begin{align}
  &A_X[a,b] = \ln \int_X dx\, e^{-\frac{1}{2} ax^2 + bx} = A[a,b] + \ln p_X[a,b], \\
  &r_X[a,b] = \frac{b}{a} - \frac{1}{\sqrt{a}}
  \frac{\mathcal{N}(z_\text{max})-\mathcal{N}(z_\text{min})}{\Phi(z_\text{max})-\Phi(z_\text{min})}, \\
  &v_X[a,b] = \frac{1}{a} \left\{
    1 -
    \frac{z_\text{max}\mathcal{N}(z_\text{max})-z_\text{min}\mathcal{N}(z_\text{min})}{\Phi(z_\text{max})-\Phi(z_\text{min})}
    - \left[
    \frac{\mathcal{N}(z_\text{max})-\mathcal{N}(z_\text{min})}{\Phi(z_\text{max})-\Phi(z_\text{min})}
    \right]^2
  \right\}.
\end{align}
The corresponding exponential family is the truncated Normal distribution:
\begin{equation}
  p_X(x\mid a,b) = \frac{1}{p_X[a,b]} \mathcal{N}(x\mid r,v) \mathbf{1}_X(x)
  \quad\text{with}\quad 
  r=\frac{b}{a}, \quad  v=\frac{1}{a}.
\end{equation}

\subsubsection{Positive/negative variable \label{module:positive_variable}} 
For $x \in \mathbb{R}_\pm$, sufficient statistics $\phi(x) =( x, -\frac{1}{2} x^2 )$, natural parameters $\lambda = (b, a)$.
It's a particular case of the interval variable with $X=\mathbb{R}_\pm$.
\begin{align}
  &p_\pm[a,b] = \int_{\mathbb{R}_\pm} dx \mathcal{N}(x\mid a,b) = \Phi(z_\pm)
  \quad \text{with} \quad
  z_\pm = \pm \frac{b}{\sqrt{a}}, \\
  &A_\pm[a,b] = \ln \int_{\mathbb{R}_\pm} dx\, e^{-\frac{1}{2} ax^2 + bx}
  = A[a,b] + \ln p_\pm[a,b], \\
  &r_\pm[a,b] = \pm \frac{1}{\sqrt{a}} \left\{
  z_\pm +  \frac{\mathcal{N}(z_\pm)}{\Phi(z_\pm)}
  \right\}, \\
  &v_\pm[a,b] = \frac{1}{a} \left\{
    1
    - \frac{z_\pm \mathcal{N}(z_\pm)}{\Phi(z_\pm)}
    -  \frac{\mathcal{N}(z_\pm)^2}{\Phi(z_\pm)^2}
  \right\}.
\end{align}
The corresponding exponential family is the half Normal:
\begin{equation}
  p_\pm(x\mid a,b) = \frac{1}{p_\pm[a,b]} \mathcal{N}(x\mid r,v) \mathbf{1}_{ \mathbb{R}_\pm }(x)
  \quad\text{with}\quad 
  r=\frac{b}{a}, \quad  v=\frac{1}{a}.
\end{equation}

\subsubsection{Phase (circular) variable \label{module:phase_variable}} 
For $x = e^{i\theta_x} \in \mathbb{S}^1$, sufficient statistics $\phi(x) = x$, natural parameter $\lambda = b$.
Generally the von Mises distribution on the circle
is defined over the angle $\theta_x \in [0,2\pi[$
but we find it more convenient to define it over the phase
$x = e^{i\theta_x} \in \mathbb{S}^1$. Then the natural parameter $b =  |b| e^{i \theta_b} \in \mathbb{C}$ and: 
\begin{align}
&A_{\mathbb{S}^1}[b] = \ln \int_{\mathbb{S}^1} dx\, e^{b^\intercal  x} = \ln 2\pi I_0(|b|) \\
&r_{\mathbb{S}^1}[b] = \frac{b}{|b|} \frac{I_1(|b|)}{I_0(|b|)}, \qquad
v_{\mathbb{S}^1}[b] = \frac{1}{2} \left[ 1 -\frac{I_1(|b|)^2}{I_0(|b|)^2} \right],
\end{align}
where $I_0$ is the modified Bessel function of the first kind.
For the natural parameter $b$, its modulus $|b|$ is called
the concentration parameter and is analogous to the precision for a Gaussian,
and its angle $\theta_b =\theta_r$ is the circular mean.
The corresponding exponential family is the von Mises:
\begin{equation}
  p(x \mid  b) = \frac{e^{\kappa \cos(\theta_x - \mu)}}{2\pi I_0(\kappa)}
  \quad\text{with}\quad 
  \kappa=|b|, \quad  \mu = \theta_b.
\end{equation}

\subsection{Linear channels}

\subsubsection{Generic linear channel \label{module:linear}}

\begin{figure}[H]
  \centering
  \begin{tikzpicture}
    \node[latent, label=$\mathbb{R}^{N}$] (z) {$z$} ;
    \node[latent, right=2 of z, label=$\mathbb{R}^{M}$] (x) {$x$} ;
    \factor[left=1 of x] {f} {$W$} {z} {x} ;
  \end{tikzpicture}
\end{figure}

The factor $f(x, z) = p(x\mid z) = \delta(x - Wz)$ is the
deterministic channel $x = Wz$. Unless explicitly specified, we will always consider isotropic Gaussian beliefs on both $x$ and $z$. The log-partition is given by:
\begin{align}
    &A_f[a_{z\to f},b_{z\to f},a_{x\to f},b_{x\to f}]
    = \frac{1}{2} b^\intercal \Sigma b + \frac{1}{2} \ln \det 2\pi\Sigma \\
    &b = b_{z\to f} + W^\intercal  b_{x\to f}, \quad
    a = a_{z\to f} + a_{x\to f} W^\intercal  W, \quad
    \Sigma = a^{-1}.
\end{align}
The posterior means and variances are given by:
\begin{align}
  \label{linear_rv_vz}
  &r_z^{f} = \Sigma b, \qquad
  &\;v_z^{f} = \mathbb{E}_\lambda
  \frac{1}{a_{z\to f} + a_{x\to f} \lambda}, \\
    \label{linear_rx_vx}
  &r_x^{f} = W r_z^{f}, \qquad
  &\alpha v_x^{f} = \mathbb{E}_\lambda
  \frac{\lambda}{a_{z\to f} + a_{x\to f} \lambda},
\end{align}
where $\lambda = \text{Spec } W^\intercal  W$ denotes the spectrum of $W^\intercal W$ and
$\alpha = \frac{M}{N}$ the aspect ratio of $W$. 
There is actually no need to explicitly compute the
matrix inverse $\Sigma = a^{-1}$ at each update; it is more numerically
efficient to use
the SVD decomposition (see Section~\ref{module:svd}). The variances satisfy:
\begin{align}
  &a_{z \to f} v_z^{f}
  + \alpha a_{x \to f} v_x^{f} = 1, \\
  &\alpha a_{x \to f} v_x^{f}
  = 1 - a_{z \to f} v_z^{f}
  = n_\text{eff},
\end{align}
where
$n_\text{eff} = \mathbb{E}_\lambda \frac{a_{x\to f}\lambda}{a_{z\to f}+a_{x\to f}\lambda}$
is known as the effective number of parameters in Bayesian linear
regression \citep{Bishop2006}. 

\subsubsection{Teacher prior second moment}
The log-partition for the teacher prior second moment is equal to:
\begin{equation}
\label{linear_teacher_second_moment}
A_f^{(0)}[\hat{\tau}_{z \to f}^{(0)}, \hat{\tau}_{x \to f}^{(0)}] = \frac{1}{2} \EE_\lambda \ln \frac{2\pi}{\hat{\tau}_\lambda^{(0)}}
\quad \text{with} \quad \hat{\tau}_\lambda^{(0)} = \hat{\tau}_{z \to f}^{(0)} + \lambda \hat{\tau}_{x \to f}^{(0)} 
\end{equation}
which yields the dual mapping:
\begin{equation}
    \tau_z^{(0)} = \EE_\lambda \tau_\lambda^{(0)} , \quad \alpha \tau_x^{(0)} = \EE_\lambda \lambda \tau_\lambda^{(0)} 
    \quad \text{with} \quad  \tau_\lambda^{(0)} = \frac{1}{\hat{\tau}_\lambda^{(0)} }.
\end{equation}
When the teacher factor graph is a Bayesian network (Section~\ref{sec:teacher_second_moments}) we have
\begin{equation}
    \hat{\tau}^{(0)}_{z \to f} = \frac{1}{\tau_z^{(0)}}, \quad   
    \hat{\tau}^{(0)}_{x \to f} = 0, \quad
    \tau_\lambda^{(0)} = \tau_z^{(0)}, \quad A_f^{(0)} = \frac{1}{2} \ln 2\pi \tau_z^{(0)}.
\end{equation}

\subsubsection{Replica symmetric \label{module:linear_rs}}

The RS potential is given by:
 \begin{align}
    \notag
     &\bar{A}_f[
     \hat{m}_{z \to f},\hat{q}_{z \to f},\hat{\tau}_{z \to f},
     \hat{m}_{x \to f},\hat{q}_{x \to f},\hat{\tau}_{x \to f}
     ] = \EE_{\lambda} \bar{A}_\lambda[\hat{m}_\lambda,\hat{q}_\lambda,\hat{\tau}_\lambda] , \\
    \notag
     &\bar{A}_\lambda[\hat{m}_\lambda,\hat{q}_\lambda,\hat{\tau}_\lambda] = 
     \frac{\hat{m}_\lambda^2 \tau_\lambda^{(0)} + \hat{q}_\lambda}{2a_\lambda} + \frac{1}{2} \ln \frac{2\pi}{a_\lambda} ,\\
     &\text{with} \quad
     a_\lambda = \hat{\tau}_\lambda + \hat{q}_\lambda
     \quad \text{and} \quad
     \hat{m}_\lambda = \hat{m}_{z \to f} + \lambda \hat{m}_{x \to f} 
     \quad (\text{idem } \hat{q}, \hat{\tau}, a, \hat{\tau}^{(0)})
\end{align}
We recognize $\bar{A}_\lambda$ as the variable RS potential Eq.~\eqref{RS_potential_variable} which yields 
the dual mapping:
\begin{align}
    \notag
     &m_z^f  = \EE_{\lambda} m_\lambda, \quad \alpha m_x^f = \EE_{\lambda} \lambda m_\lambda
     \quad (\text{idem } q, \tau, v) \\
     \text{with} \quad
     &m_\lambda = \frac{\hat{m}_\lambda \tau_\lambda^{(0)}}{a_\lambda}, \quad
     q_\lambda = \frac{\hat{m}_\lambda^2 \tau_\lambda^{(0)} + \hat{q}_\lambda}{a_\lambda^2}, \quad
     \tau_\lambda = q_\lambda + v_\lambda, \quad
     v_\lambda = \frac{1}{a_\lambda}.
 \end{align}
 In particular we recover Eqs~\eqref{linear_rv_vz}-\eqref{linear_rx_vx} for the variances.

\subsubsection{Bayes-optimal \label{module:linear_bo}}

The BO potential is given by:
 \begin{align}
    \notag
     &\bar{A}_f^{(0)}[\hat{m}_{z \to f},\hat{m}_{x \to f}] = 
     \EE_{\lambda} \bar{A}_\lambda[\hat{m}_\lambda] , \quad 
     \bar{A}_\lambda^{(0)}[\hat{m}_\lambda] =
     \frac{\hat{m}_\lambda \tau_\lambda^{(0)}}{2} + \frac{1}{2} \ln \frac{2\pi}{a_\lambda}, \\
     &\text{with} \quad
     a_\lambda = \hat{\tau}^{(0)}_\lambda + \hat{m}_\lambda
     \quad \text{and} \quad
     \hat{m}_\lambda = \hat{m}_{z \to f} + \lambda \hat{m}_{x \to f} 
     \quad (\text{idem } a, \hat{\tau}^{(0)})
\end{align}
We recognize $\bar{A}_\lambda^{(0)}$ as the variable BO potential Eq.~\eqref{BO_potential_variable}
which yields 
the dual mapping:
\begin{align}
     m_z^f  = \EE_{\lambda} m_\lambda, \quad \alpha m_x^f = \EE_{\lambda} \lambda m_\lambda
     \quad (\text{idem } \tau^{(0)}, v) \quad
     \text{with} \quad
     m_\lambda = \tau_\lambda^{(0)} - v_\lambda, \quad v_\lambda = \frac{1}{a_\lambda}.
 \end{align}
In particular we recover Eqs~\eqref{linear_rv_vz}-\eqref{linear_rx_vx} for the variances. 
The decomposition Eq.~\eqref{BO_potential_decomposition} for the BO potential reads:
\begin{equation}
\bar{A}_f^{(0)}[\hat{m}_{z \to f}, \hat{m}_{x \to f}] = \frac{\hat{m}_{z \to f} \tau_z^{(0)} + \alpha \hat{m}_{x \to f} \tau_x^{(0)}}{2} 
- I_f[\hat{m}_{z \to f}, \hat{m}_{x \to f}] + A_f^{(0)}[\hat{\tau}_{z \to f}^{(0)},\hat{\tau}_{z \to f}^{(0)}]
\end{equation}
where $A_f^{(0)}$ is given by Eq.~\eqref{linear_teacher_second_moment} and the mutual information by:
\begin{equation}
    I_f[\hat{m}_{z \to f}, \hat{m}_{x \to f}] = \frac{1}{2} \mathbb{E}_\lambda \ln a_\lambda \tau_\lambda^{(0)} .
\end{equation}
From these expressions, it is straightforward to check the dual mapping to the variances:
\begin{equation}
  \tfrac{1}{2} v_z^{f}
  = \partial_{\hat{m}_{z \to f}} I_f, \qquad
  \tfrac{1}{2} \alpha v_x^{f}
  = \partial_{\hat{m}_{x \to f}} I_f,
\end{equation}
as well as the dual mapping to the overlaps:
\begin{equation}
  \tfrac{1}{2} m_z^{f}
  = \partial_{\hat{m}_{z \to f}} \bar{A}_f^{(0)}, \qquad
  \tfrac{1}{2} \alpha m_x^{f}
  = \partial_{\hat{m}_{x \to f}} \bar{A}_f^{(0)}.
\end{equation}

\subsubsection{Random matrix theory expressions \label{module:rmt}} 
The posterior variances and the mutual information are closely 
related to the following
transforms in random matrix theory \citep{Tulino2004}:
\begin{align}
  \text{Shannon transform} \quad
  &\mathcal{V}(\gamma) = \mathbb{E}_\lambda \ln(1+\gamma\lambda) \\
  \eta\text{ transform} \quad
  &\eta(\gamma) = \mathbb{E}_\lambda \frac{1}{1+\gamma\lambda} \\
  \text{Stieltjes transform} \quad
  &\mathcal{S}(z) = \mathbb{E}_\lambda \frac{1}{\lambda - z} \\
  \text{R transform} \quad
  &R(s) = \mathcal{S}^{-1}(-s) - \frac{1}{s}
\end{align}
where $\mathcal{S}^{-1}$ denotes the functional inverse of $\mathcal{S}$.
Following \cite{Reeves2017} let's introduce
the integrated R-transform and its Legendre transform
\begin{equation}
    \label{integrated_R_transform}
  J(t) = \frac{1}{2} \int_0^t dz R(-z), \qquad
  J^*(u) = \sup_{t} J(t) -\frac{1}{2} ut.
\end{equation}
Using the identities \citep{Tulino2004}
\begin{equation}
  \gamma \frac{d}{d\gamma}\mathcal{V}(\gamma) = 1 - \eta(\gamma) = -\phi R(\phi )
  \quad \text{with }
  \phi =  - \gamma \eta(\gamma),
\end{equation}
it can be shown that:
\begin{equation}
 a_{z \to f} v_z^{f} = \eta(\gamma) \quad \text{with }
  \gamma = \frac{a_{x \to f}}{a_{z \to f}}, \quad
  u = \frac{\alpha v_x^{f}}{v_z^{f}} = R(\phi), \quad
  J^*(u) = \frac{1}{2} \EE_\lambda \ln a_\lambda v_z  .
\end{equation}
The Legendre transforms of the BO potential and mutual information
\begin{align}
    \bar{A}_f^{(0*}[m_z,m_x] &= \sup_{\hat{m}_{z \to f},\hat{m}_{x \to f}} \frac{\hat{m}_{z \to f}m_z + \alpha\hat{m}_{x \to f}m_x }{2} -  \bar{A}_f^{(0)}[\hat{m}_{z \to f},\hat{m}_{x \to f}] , \\
     I_f^*[v_z,v_x] &= \sup_{\hat{m}_{z \to f},\hat{m}_{x \to f}}  I_f[\hat{m}_{z \to f},\hat{m}_{x \to f}]  - \frac{\hat{m}_{z \to f}v_z + \alpha\hat{m}_{x \to f}v_x }{2} ,
\end{align}
are equal to
\begin{align}
    &\bar{A}_f^{(0)*}[m_z,m_x] = J^*(u) - \frac{1}{2} \ln 2\pi e v_z + 
    \frac{\hat{\tau}_{z \to f}^{(0)} v_z + \alpha \hat{\tau}_{x \to f}^{(0)} v_x}{2}, \\
     &I_f^*[v_z,v_x] = \bar{A}_f^{(0)*}[m_z,m_x] + 
     A_f^{(0)}[\hat{\tau}_{z \to f}^{(0)} , \hat{\tau}_{x \to f}^{(0)} ].
\end{align}
In particular when the teacher factor graph is a Bayesian network we recover \citep{Reeves2017}:
\begin{equation}
    I_f^*[v_z,v_x] = J^*(u) + I_z^*[v_z], \quad 
    I_z^*[v_z] = \frac{1}{2} \left( \ln\frac{\tau_z^{(0)}}{v_z} + \frac{v_z}{\tau_z^{(0)}} -1\right).
\end{equation}
When the matrix $W$ belongs to an ensemble for which the limiting spectral density
of $W^\intercal  W$ is known (for example Section~\ref{module:iid_entries}) the transforms above can be derived analytically, leading to the S-AMP approach \citep{Cakmak2014, Cakmak2016}.

\subsubsection{Gibbs free energy \label{module:gibbs_linear}}

The Gibbs free energy for the linear channel can be expressed as a function of the posterior variances:
\begin{equation}
    G_f[v_z, v_x] =  G_z[v_z] + N J^*(u) , 
    \quad u = \frac{\alpha v_x}{v_z},
\end{equation}
where $G_z[v_z] = - \frac{N}{2} \ln 2\pi e v_z$ is the variable negative entropy and 
 $J^*(u)$ the dual integrated R-transform Eq.~\eqref{integrated_R_transform}.
Viewed as a function of the posterior variances, $G_f$ gives the dual mapping to the incoming precisions:
\begin{equation}
    - \frac{N}{2} a_{z \to f} = \partial_{v_z} G_f[v_z, v_x] , \quad 
    - \frac{M}{2} a_{x \to f} = \partial_{v_x} G_f[v_z, v_x] ,
\end{equation}
Similarly the variable negative entropy $G_z$ gives the dual mapping to the precision
\begin{equation}
    - \frac{N}{2} a_z = \partial_{v_z} G_z[v_z] \implies a_z = \frac{1}{v_z} \quad 
    \text{(idem x)}.
\end{equation}
Then the potential 
\begin{equation}
    \label{linear_G_tilde}
   \tilde{G}_f[v_x, v_z] = G_x[v_x] + G_z[v_z] - G_f[v_z, v_x] = G_x[v_x] - N J^*(u) 
\end{equation}
gives the dual mapping to the \emph{outgoing} precisions:
\begin{equation}
    \label{dual_mapping_linear}
    -\frac{N}{2} a_{f \to z} = \partial_{v_z} \tilde{G}_f[v_z,v_z] , \quad 
    -\frac{M}{2} a_{f \to x} = \partial_{v_x} \tilde{G}_f[v_z, v_x] .
\end{equation}

\subsubsection{Matrix with iid entries \label{module:iid_entries}}

Let $W$ be a random matrix with iid entries of mean $0$ and variance $\frac{1}{N}$. Then the limiting spectral density $\rho(\lambda)$ of $\lambda = \text{Spec } W^\intercal  W$ as $N \to \infty$ with $\alpha=\frac{M}{N} = O(1)$ follows the Marchenko-Pastur law:
\begin{equation}
    \rho(\lambda) = \max(0, 1 - \alpha) \delta(\lambda) + 
    \frac{1}{2\pi\lambda} \sqrt{(\lambda_+ - \lambda)(\lambda-\lambda_-)} \mathbf{1}_{[\lambda_-, \lambda_+]}(\lambda)
    \quad
    \lambda_\pm = (1 \pm \sqrt{\alpha})^2
\end{equation}
Then the dual integrated R-transform is \citep{Reeves2017}:
\begin{equation}
    J^*(u)  = \frac{\alpha}{2} \left( \ln \frac{\alpha}{u} + \frac{u}{\alpha} - 1 \right).
\end{equation}
From Eq.~\eqref{linear_G_tilde} the potential $\tilde{G}_f$ is then equal to:
\begin{equation}
    \tilde{G}_f[v_z, v_x] = - \frac{M}{2} \left( \ln 2\pi v_z + \frac{v_x}{v_z} \right)
\end{equation}
which gives the dual mapping Eq.~\eqref{dual_mapping_linear} to the \emph{outgoing} precisions:
\begin{equation}
    a_{f \to z} = \frac{\alpha}{v_z} \left( 1 - \frac{v_x}{v_z}\right)  , \quad
    a_{f \to x} = \frac{1}{v_z}.
\end{equation}
The fact that the outgoing precision $a_{f \to x}$ towards $x$ is equal to the precision $a_z = \frac{1}{v_z}$ of $z$ is only true for the iid entries case. For a generic linear channel $a_{f \to x}$ will depend on both $v_x$ and $v_z$. 

\subsubsection{Rotation channel} If $W = R$ is a rotation, then
$v_z^{f} = v_x^{f} = \frac{1}{a}$ with
$a = a_{z \to f} + a_{x \to f}$.
Besides the forward $f\to x$
and backward $f \to z$  updates are simple rotations in parameter space:
\begin{align}
  &a_{f\to x}^\text{new} = a_{z\to f}, \qquad
  &b_{f\to x}^\text{new} = R b_{z\to f}, \\
  &a_{f\to z}^\text{new} = a_{x\to f}, \qquad
  &b_{f\to z}^\text{new} = R^\intercal  b_{x\to f}.
\end{align}

\subsubsection{Scaling channel} 
When the weight matrix $W=S$ is a diagonal $M \times N$
matrix, the eigenvalue distribution is equal to $\lambda = S^\intercal  S$ and
the posterior mean and variances are given by:
\begin{align}
  &r_z^{f} =
  \frac{b_{z\to f}+S^\intercal  b_{x\to f}}{a_{z\to f}+a_{x\to f}\lambda},
  \qquad
  &\; v_z^{f} = \mathbb{E}_\lambda
  \frac{1}{a_{z\to f}+a_{x\to f}\lambda}, \\
  &r_x^{f} = S r_z^{f},
  &\alpha v_x^{f} = \mathbb{E}_\lambda
  \frac{\lambda}{a_{z\to f}+a_{x\to f}\lambda}.
\end{align}
In particular, for out-of-rank components, the posterior
mean $r_z^{f}$ is set to the prior and the posterior
mean $r_x^{f}$ is set to zero:
\begin{equation}
  r_z^{f(n)} = \frac{b^{(n)}_{z\to f}}{a_{z \to f}}
  \quad \text{for} \quad R < n \leq N, 
  \quad
  r_x^{f(m)} = 0 
  \quad \text{for} \quad
   R < m \leq M .
\end{equation}

\subsubsection{SVD decomposition \label{module:svd}} 
As proposed by \cite{Rangan2018} for VAMP,
it is more efficient
to precompute the SVD decomposition:
\begin{equation}
  W=U_x S V_z^\intercal , \; U_x \in O(M), \; V_z \in O(N), \;
  S\in\mathbb{R}^{M\times N}\text{ diagonal}.
\end{equation}
The eigenvalue distribution of $W^\intercal W$ is equal to $\lambda = S^\intercal  S$.
Then the EP updates for $W$ are equivalent to
the composition of a rotation $V_z^\intercal $ in $z$-space, a scaling $S$ that projects
$z$ into the $x$ space and a rotation $U_x$ in $x$-space.
\begin{figure}[H]
  \centering
  \begin{tikzpicture}
    \node[latent, label=$\mathbb{R}^{N}$] (z) {$z$} ;
    \node[latent, right=2 of z, label=$\mathbb{R}^{N}$] (z1) {$\tilde{z}$} ;
    \node[latent, right=2 of z1, label=$\mathbb{R}^{M}$] (x1) {$\tilde{x}$} ;
    \node[latent, right=2 of x1, label=$\mathbb{R}^{M}$] (x) {$x$} ;
    \factor[left=1 of z1] {f} {$V_z^\intercal $} {z} {z1} ;
    \factor[left=1 of x1] {f} {$S$} {z1} {x1} ;
    \factor[left=1 of x] {f} {$U_x$} {x1} {x} ;
  \end{tikzpicture}
\end{figure}
These updates are only rotations or element-wise scaling
and are thus considerably faster than
solving $ a r_z^{f} = b$ or even worse computing
the inverse $\Sigma = a^{-1}$ at each update.
It comes at the expense of computing
the SVD decomposition of $W$, but this only needs to be done once.

\subsubsection{Complex linear channel}

The real linear channel can be easily extended to the complex
linear channel $x = Wz$ with  $x \in \mathbb{C}^{M}$, $z \in \mathbb{C}^{N}$
and $W \in \mathbb{C}^{M\times N}$ and $\lambda = \text{Spec } W^\dagger W$.

\subsubsection{Unitary channel} 
When $W=U$ is unitary, for instance
when $W=\mathcal{F}$ is the discrete Fourier transform (DFT), then
$v_z^{f} = v_x^{f} = \frac{1}{a}$ with
$a = a_{z \to f} + a_{x \to f}$.
Besides the forward $f\to x$
and backward $f \to z$ updates are simple unitary transforms
in parameter space:
\begin{align}
  &a_{f\to x}^\text{new} = a_{z\to f}, \qquad
  &b_{f\to x}^\text{new} = U b_{z\to f}, \\
  &a_{f\to z}^\text{new} = a_{x\to f}, \qquad
  &b_{f\to z}^\text{new} = U^\dagger b_{x\to f}.
\end{align}

\subsubsection{Convolution channel (complex)}

\begin{figure}[H]
  \centering
  \begin{tikzpicture}
    \node[latent, label=$\mathbb{C}^{N}$] (z) {$z$} ;
    \node[latent, right=2 of z, label=$\mathbb{C}^{N}$] (x) {$x$} ;
    \factor[left=1 of x] {f} {$\ast w$} {z} {x} ;
  \end{tikzpicture}
\end{figure}
The convolution channel $x = w \ast z$ with convolution weights
$w \in \mathbb{C}^{N}$ is a complex linear channel $x = Wz$ with $M=N$.
It is equivalent to the composition of a discrete Fourier transform (DFT) $\mathcal{F}$ for $z$,
a multiplication by $\hat{w} = \mathcal{F} w \in \mathbb{C}^N$, and
an inverse DFT $\mathcal{F}^{-1}$ for $x$.
The eigenvalue distribution of $W^\dagger W$ is equal to
$\lambda = \hat{w}^\dagger \hat{w} = \vert \hat{w} \vert^2$.

\begin{figure}[H]
  \centering
  \begin{tikzpicture}
    \node[latent, label=$\mathbb{C}^{N}$] (z) {$z$} ;
    \node[latent, right=2 of z, label=$\mathbb{C}^{N}$] (z1) {$\hat{z}$} ;
    \node[latent, right=2 of z1, label=$\mathbb{C}^{N}$] (x1) {$\hat{x}$} ;
    \node[latent, right=2 of x1, label=$\mathbb{C}^{N}$] (x) {$x$} ;
    \factor[left=1 of z1] {f} {$\mathcal{F}$} {z} {z1} ;
    \factor[left=1 of x1] {f} {$\hat{w}$} {z1} {x1} ;
    \factor[left=1 of x] {f} {$\mathcal{F}^{-1}$} {x1} {x} ;
  \end{tikzpicture}
\end{figure}

\subsubsection{Convolution channel (real)}

\begin{figure}[H]
  \centering
  \begin{tikzpicture}
    \node[latent, label=$\mathbb{R}^{N}$] (z) {$z$} ;
    \node[latent, right=2 of z, label=$\mathbb{R}^{N}$] (x) {$x$} ;
    \factor[left=1 of x] {f} {$\ast w$} {z} {x} ;
  \end{tikzpicture}
\end{figure}
The convolution channel $x = w \ast z$ with convolution weights
$w \in \mathbb{R}^{N}$ is a real linear channel $x = Wz$ with $M=N$.
It is equivalent to the composition of a discrete Fourier transform (DFT) $\mathcal{F}$ for $z$,
a multiplication by $\hat{w} = \mathcal{F} w \in \mathbb{C}^N$, and
an inverse DFT $\mathcal{F}^{-1}$ for $x$.
The eigenvalue distribution of $W^\intercal  W$ is equal to
$\lambda = \hat{w}^\dagger \hat{w} = \vert \hat{w} \vert^2$.

\begin{figure}[H]
  \centering
  \begin{tikzpicture}
    \node[latent, label=$\mathbb{R}^{N}$] (z) {$z$} ;
    \node[latent, right=2 of z, label=$\mathbb{C}^{N}$] (z1) {$\hat{z}$} ;
    \node[latent, right=2 of z1, label=$\mathbb{C}^{N}$] (x1) {$\hat{x}$} ;
    \node[latent, right=2 of x1, label=$\mathbb{R}^{N}$] (x) {$x$} ;
    \factor[left=1 of z1] {f} {$\mathcal{F}$} {z} {z1} ;
    \factor[left=1 of x1] {f} {$\hat{w}$} {z1} {x1} ;
    \factor[left=1 of x] {f} {$\mathcal{F}^{-1}$} {x1} {x} ;
  \end{tikzpicture}
\end{figure}

\subsubsection{Full covariance beliefs \label{module:linear_full_cov}}

In this subsection we will consider the linear channel $x=Wz$ with full covariance beliefs on $x$ and $z$, meaning that the precisions $a_{z\to f}$ and $a_{x\to f}$ are $N \times N$ and $M \times M$ matrices. The log-partition is given by:
\begin{align}
    &A_f[a_{z\to f}, b_{z\to f}, a_{x\to f}, b_{x\to f}]
    = \frac{1}{2} b^\intercal \Sigma b + \frac{1}{2} \ln \det 2\pi\Sigma \\
    &b = b_{z\to f} + W^\intercal  b_{x\to f}, \quad
    a = a_{z\to f} +  W^\intercal  a_{x\to f} W, \quad
    \Sigma = a^{-1}.
\end{align}
The posterior mean and covariance are given by:
\begin{align}
  r_z^{f} &= \Sigma b,  \quad
  &\Sigma_z^{f} =& \Sigma, \\
  r_x^{f} &= W r_z^{f}, \quad
  &\Sigma_x^{f} =& W \Sigma W^\intercal ,
\end{align}
We have $a_z^f = a$ and $b_z^f = b$ so the backward $f \to z$ update is simply given by:
\begin{equation}
    a_{f \to z}^\text{new} = W^\intercal  a_{x \to f} W, \qquad 
    b_{f \to z}^\text{new} = W^\intercal  b_{x \to f}
\end{equation}
We have:
\begin{align}
 \label{linearop_ax_new}
    a_x^f &= (W \Sigma W^\intercal )^{-1}
            = a_{x\to f} + (W a_{z\to f}^{-1} W^\intercal )^{-1}
            \\
  \label{linearop_bx_new}
  b_x^f &= (W \Sigma W^\intercal )^{-1} W \Sigma b
            =  b_{x\to f} +
            (W \Sigma W^\intercal )^{-1} W\Sigma b_{z\to f}
\end{align}
We can obtain the RHS of
Eq~\eqref{linearop_ax_new} by two applications of the Woodbury identity. The RHS of Eq~\eqref{linearop_bx_new} follows directly from the definition
of $b$. Using the mean and covariance of the $z\to f$
and $f\to x$ messages:
\begin{align}
  \Sigma_{z \to f} = a_{z \to f}^{-1},  \qquad
  &r_{z \to f} = a_{z \to f}^{-1} b_{z \to f} \\
  \Sigma_{f \to x} = a_{f \to x}^{-1}, \qquad
  &r_{f \to x} = a_{f \to x}^{-1} b_{f \to x} 
\end{align}
we can write the forward $f \to x$ update as:
\begin{equation}
  \Sigma_{f\to x}^\text{new} = W \Sigma_{z\to f} W^\intercal , \qquad
  r_{f\to x}^\text{new} = W r_{z\to f}.
\end{equation}

\subsection{Separable priors}

\subsubsection{Generic separable prior}

\begin{figure}[H]
  \centering
  \begin{tikzpicture}
    \node[latent, label=$\mathbb{R}^N$] (x) {$x$} ;
    \factor[left=1 of x] {p0} {$p_0$} {} {x} ;
  \end{tikzpicture}
\end{figure}

Let $f(x) = p_0(x) = \prod_{n=1}^N p_0(x^{(n)})$ be a separable prior over $x \in \mathbb{R}^N$. The log-partition, posterior mean and variance are given by:
\begin{align}
    \label{separable_Af_prior}
    &A_f[a_{x \to f}, b_{x \to f}] = \sum_{n=1}^{N} A_f[a_{x \to f}, b_{x \to f}^{(n)}], \\
    \label{separable_vf_prior}
    &r_x^{f}[a_{x \to f}, b_{x \to f}]^{(n)} = r_x^{f}[a_{x \to f}, b_{x \to f}^{(n)}], \quad
    v_x^{f}[a_{x \to f}, b_{x \to f}] = \frac{1}{N} \sum_{n=1}^{N}
    v_x^{f}[a_{x \to f}, b_{x \to f}^{(n)}].
\end{align}
where on the RHS the same quantities are defined over scalar
$b_{x \to f}^{(n)}$ in $\mathbb{R}$:
\begin{align}
    \label{scalar_Af_prior}
    &A_f[a_{x \to f}, b_{x \to f}] = \ln \int_\mathbb{R} dx \, p_0(x) \, e^{-\frac{1}{2} a_{x \to f} x^2 + b_{x \to f} x}, \\ 
    \label{scalar_vf_prior}
    &r_x^f[a_{x \to f}, b_{x \to f}]= \partial_{b_{x \to f}} 
    A_f[a_{x \to f}, b_{x \to f}], \quad 
    v_x^f[a_{x \to f}, b_{x \to f}]= \partial^2_{b_{x \to f}} 
    A_f[a_{x \to f}, b_{x \to f}].
\end{align}
In the remainder we will only derive the scalar case, as
it can be straightforwardly extended to the
high dimensional counterpart through Eqs~\eqref{separable_Af_prior}-\eqref{separable_vf_prior}. 
For a large class of priors that we call natural priors we can derive closed-form expressions for the log-partition, mean and variance as shown in Section~\ref{module:natural_prior}; familiar examples include the Gaussian, binary, Gauss-Bernoulli, and positive priors.

\subsubsection{Teacher prior second moment}
The approximate teacher marginal (Proposition~\ref{prop:teacher_second_moments}) is:
\begin{equation}
    \label{teacher_second_moments_prior}
    p_f^{(0)}(x^{(0)} \mid \hat{\tau}_{x \to f}^{(0)}) = 
    p_0^{(0)}(x^{(0)}) \,
    e^{-\frac{1}{2} \hat{\tau}_{x \to f}^{(0)} x^{(0)2} - A_f^{(0)}[\hat{\tau}_{x \to f}^{(0)}]}, 
\end{equation}
with log-partition:
\begin{equation}
    \label{teacher_second_moments_prior_A}
    A_f^{(0)}[\hat{\tau}_{x \to f}^{(0)}] = \ln \int_\mathbb{R} dx \, p_0^{(0)}(x) \, 
    e^{-\frac{1}{2} \hat{\tau}_{x \to f}^{(0)} x^2}
\end{equation}
which yields the dual mapping:
\begin{equation}
    \tau_x^{(0)} = -2 \partial_{\hat{\tau}_{x \to f}^{(0)}} A_f^{(0)}[\hat{\tau}_{x \to f}^{(0)}]. 
\end{equation}
When the teacher factor graph is a Bayesian network (Section~\ref{sec:teacher_second_moments}) we have
\begin{equation}
    \hat{\tau}^{(0)}_{x \to f} = 0, \quad
     p_f^{(0)}(x^{(0)} \mid \hat{\tau}_{x \to f}^{(0)}) = p_0^{(0)}(x^{(0)}).
\end{equation}

\subsubsection{Replica symmetric \label{module:prior_rs}}
The RS potential and overlaps are given by a low-dimensional integration of the corresponding scalar EP quantities Eqs~\eqref{scalar_Af_prior}-\eqref{scalar_vf_prior}:
\begin{align}
    &\bar{A}_f[\hat{m}_{x \to f},\hat{q}_{x \to f},\hat{\tau}_{x \to f}] = 
    \int_\mathbb{R} db_{x \to f}  \, p_f^{(0)}(b_{x \to f}) \, A_f[a_{x \to f}, b_{x \to f}], \\
    &m_x^f[\hat{m}_{x \to f},\hat{q}_{x \to f},\hat{\tau}_{x \to f}] = 
    \int_{\mathbb{R}^2} db_{x \to f} dx^{(0)}  \, p_f^{(0)}(b_{x \to f}, x^{(0)}) \, x^{(0)} r_x^f[a_{x \to f}, b_{x \to f}] \\
    &q_x^f[\hat{m}_{x \to f},\hat{q}_{x \to f},\hat{\tau}_{x \to f}] = 
    \int_\mathbb{R} db_{x \to f}  \, p_f^{(0)}(b_{x \to f}) \, r_x^f[a_{x \to f}, b_{x \to f}]^2, \\
    &v_x^f[\hat{m}_{x \to f},\hat{q}_{x \to f},\hat{\tau}_{x \to f}] = 
    \int_\mathbb{R} db_{x \to f}  \, p_f^{(0)}(b_{x \to f}) \, v_x^f[a_{x \to f}, b_{x \to f}], \\
    &\tau_x^f[\hat{m}_{x \to f},\hat{q}_{x \to f},\hat{\tau}_{x \to f}] 
    = q_x^f[\hat{m}_{x \to f},\hat{q}_{x \to f},\hat{\tau}_{x \to f}] 
    + v_x^f[\hat{m}_{x \to f},\hat{q}_{x \to f},\hat{\tau}_{x \to f}], 
\end{align}
where the ensemble average Eq.~\eqref{RS_ensemble_avg_f} is now over scalar $b_{x \to f}$ and $x^{(0)}$ in $\mathbb{R}$:
\begin{equation}
    p_f^{(0)}(b_{x \to f}, x^{(0)}) = \mathcal{N}(b_{x \to f} \mid \hat{m}_{x \to f} x^{(0)}, \hat{q}_{x \to f}) \, p_f^{(0)}(x^{(0)} \mid \hat{\tau}_{x \to f}^{(0)}) 
\end{equation}
with $a_{x \to f} = \hat{\tau}_{x \to f} + \hat{q}_{x \to f}$ and 
$p_f^{(0)}(x^{(0)} \mid \hat{\tau}_{x \to f}^{(0)})$ given by Eq.~\eqref{teacher_second_moments_prior}.
The dual mapping to the overlaps now simply reads:
\begin{equation}
    m_x^{f}
  = \partial_{\hat{m}_{x \to f}} \bar{A}_f, \quad 
  -\tfrac{1}{2} q_x^{f}
  = \partial_{\hat{q}_{x \to f}} \bar{A}_f, \quad
  -\tfrac{1}{2} \tau_x^{f}
  = \partial_{\hat{\tau}_{x \to f}} \bar{A}_f.
\end{equation}

\subsubsection{Bayes-optimal \label{module:prior_bo}}
The BO potential and overlap are given by a low-dimensional integration of the corresponding scalar EP quantities:
\begin{align}
    &\bar{A}_f^{(0)}[\hat{m}_{x \to f}] = 
    \int_\mathbb{R} db_{x \to f}  \, p_f^{(0)}(b_{x \to f}) \, A_f^{(0)}[a_{x \to f}, b_{x \to f}], \\
    &v_x^f[\hat{m}_{x \to f}] = 
    \int_\mathbb{R} db_{x \to f}  \, p_f^{(0)}(b_{x \to f}) \, v_x^f[a_{x \to f}, b_{x \to f}], \\
    &m_x^f[\hat{m}_{x \to f}] = \tau_x^{(0)} - v_x^f[\hat{m}_{x \to f}], 
\end{align}
where the ensemble average Eq.~\eqref{BO_ensemble_avg_f} is now over scalar $b_{x \to f}$ and $x^{(0)}$ in $\mathbb{R}$:
\begin{equation}
    p_f^{(0)}(b_{x \to f}, x^{(0)}) = \mathcal{N}(b_{x \to f} \mid \hat{m}_{x \to f} x^{(0)}, \hat{m}_{x \to f}) \, p_f^{(0)}(x^{(0)} \mid \hat{\tau}_{x \to f}^{(0)}) 
\end{equation}
with $a_{x \to f} =  \hat{\tau}_{x \to f}^{(0)}+ \hat{m}_{x \to f}$ and 
$p_f^{(0)}(x^{(0)} \mid \hat{\tau}_{x \to f}^{(0)})$ given by Eq.~\eqref{teacher_second_moments_prior}.
The relationship Eq.~\eqref{BO_potential_decomposition}
between the mutual information and the BO potential now reads:
\begin{equation}
    I_f[\hat{m}_{x \to f}] = \tfrac{1}{2} \hat{m}_{x \to f}\tau_x^{(0)} - \bar{A}_f^{(0)}[\hat{m}_{x \to f}] + A_f^{(0)}[\hat{\tau}_{x \to f}^{(0)}].
\end{equation}
The dual mapping to the variance and overlap simply reads:
\begin{equation}
  \tfrac{1}{2} v_x^{f}
  = \partial_{\hat{m}_{x \to f}} I_f, \qquad
   \tfrac{1}{2} m_x^{f}
  = \partial_{\hat{m}_{x \to f}} \bar{A}_f^{(0)}.
\end{equation}

\subsubsection{Natural prior \label{module:natural_prior}}

Let $z$ be a variable of base space $Z$, with sufficient statistics $\phi(z)$ and associated natural parameters $\lambda_z$. Several examples of variable types are presented in Section~\ref{module:variable} such as the real, binary, sparse and positive variable. A natural prior over the variable $z$ is an exponential family distribution  $p_0(z) = p(z \mid  \lambda_0) = e^{\lambda_0^\intercal  \phi(z) - A_z[\lambda_0]}$ with a given natural parameter $\lambda_0$. 

\paragraph{Variable $z$ belief} Let the factor $f(z) = p_0(z) = p(z \mid  \lambda_0)$ be a natural prior. 
Let us first consider the corresponding module with variable $z$ belief.

\begin{figure}[H]
  \centering
  \begin{tikzpicture}
    \node[const] (lambda) {$\lambda_0$} ;
    \node[latent, right=2 of lambda, label=$Z$] (z) {$z$} ;
    \factor[right=1 of lambda, left=1 of z]  {f} {$p_0$} {lambda} {z} ;
  \end{tikzpicture}
\end{figure}

The log-partition and moment function are given by:
\begin{align}
  &A_f[\lambda_{z \to f}] =
  A_z[\lambda_{z \to f} + \lambda_0] - A_z[\lambda_0], \\
  &\mu_z^f[\lambda_{z \to f}] = \mu_z[\lambda_{z  \to f} + \lambda_0],
\end{align}
where $A_z[\lambda]$ and $\mu_z[\lambda]$ denotes the log-partition and moment function of the variable $z$. The $f \to z$ update is the constant message:
\begin{equation}
  \lambda_{f \to z}^\text{new} = \lambda_0.
\end{equation}

\paragraph{Variable $x$ belief} Let $x$ be a variable of different type than $z$, with base space $X$ and sufficient statistics $\phi(x)$ and associated natural parameters $\lambda_x$. We still consider the  same factor $f(x) = p_0(x) = p(x \mid  \lambda_0)$ 
which is a natural prior in the $z$ variable, but we derive the corresponding module using a variable $x$ belief.

\begin{figure}[H]
  \centering
  \begin{tikzpicture}
    \node[const] (lambda) {$\lambda_0$} ;
    \node[latent, right=2 of lambda, label=$X$] (x) {$x$} ;
    \factor[right=1 of lambda, left=1 of x]  {f} {$p_0$} {lambda} {x} ;
  \end{tikzpicture}
\end{figure}

This is meaningful only if we can inject $Z \hookrightarrow X$.  We denote by $\phi^{(0)}$, $\phi^{(1)}$ and $\phi^{(2)}$ the set of sufficient statistics common to $x$ and $z$,
specific to $z$ and specific to $x$ respectively. We will assume that the sufficient statistics $\phi^{(2)}$ specific to $x$ are constant on $Z$:
\begin{equation}
    \phi^{(2)}(z) = \mu_Z^{(2)} \quad \text{for all} \quad z\in Z. 
\end{equation}
Then the log-partition and moment function are given by:
\begin{align}
  &A_f[\lambda_{x \to f}] = 
  A_z[\lambda_{z \to f} + \lambda_0] - A_z[\lambda_0] + 
  \lambda_{x \to f}^{(2)T} \mu_Z^{(2)}, \\
  &\mu_x^{f(0)}[\lambda_{x \to f}] = 
  \mu_z^{(0)}[\lambda_{z \to f} + \lambda_0], \qquad
  \mu_x^{f(2)}[\lambda_{x \to f}] = \mu_Z^{(2)}
\end{align}
with $\lambda_{z \to f}^{(0)} = \lambda_{x \to f}^{(0)}$ and 
$\lambda_{z \to f}^{(1)} = 0$.

\paragraph{Isotropic Gaussian belief}
To derive the isotropic Gaussian belief modules, we take $x$ to be a real variable that is $X=\mathbb{R}$ and $\phi(x) = ( x, -\frac{1}{2}x^2 )$ and associated natural parameters $\lambda_x = (b_x, a_x)$. When $z$ is a real, binary, sparse, and interval variable we obtain respectively the Gaussian, binary, Gauss-Bernoulli and truncated Normal prior as detailed in the next subsections.

\subsubsection{Gaussian prior \label{module:gaussian_prior}}

\begin{figure}[H]
  \centering
  \begin{tikzpicture}
    \node[latent, label=$\mathbb{R}$] (x) {$x$} ;
    \factor[left=1 of x]  {f} {$\mathcal{N}$} {} {x} ;
  \end{tikzpicture}
\end{figure}
The factor $f(x)=p(x \mid a_0, b_0) = \mathcal{N}(x\mid r_0, v_0)$ is the Normal prior with natural
parameters $a_0 = \frac{1}{v_0}$ and 
$b_0 = \frac{r_0}{v_0}$. The Normal prior corresponds to the natural prior for the real variable $x \in \mathbb{R}$.
According to Section~\ref{module:natural_prior} the log-partition is given by:
\begin{equation}
    A_f[a_{x\to f}, b_{x\to f}] = A[a,b] - A[a_0, b_0]
    \quad \text{with} \quad
    a=a_{x\to f} + a_0, \quad
    b = b_{x\to f} + b_0,
\end{equation}
where $A[a,b]$ is the log-partition of a real variable, see Section~\ref{module:real_variable}.
The posterior mean and variance are given by:
\begin{equation}
    r_x^{f} = \frac{b}{a}, \qquad v_x^{f} = \frac{1}{a},
\end{equation}{}
leading to the constant $f \to x$ update:
\begin{equation}
  a_{f\to x}^\text{new} = a_0,
  \qquad
  b_{f\to x}^\text{new} = b_0.
\end{equation}
The ensemble average variance is directly given by $v_x^f = \frac{1}{a}$.

\subsubsection{Binary prior}

\begin{figure}[H]
  \centering
  \begin{tikzpicture}
    \node[latent, label=$\mathbb{R}$] (x) {$x$} ;
    \factor[left=1 of x]  {f} {$p_{\pm}$} {} {x} ;
  \end{tikzpicture}
\end{figure}
The factor $f(x)=p(x \mid b_0)=p_+ \delta_{+1}(x) + p_- \delta_{-1}(x)$ is the
binary prior with natural
parameter $b_0 = \frac{1}{2} \ln \frac{p_+}{p_-}$. The binary prior corresponds to the natural prior for the binary variable $z \in \pm$.
According to Section~\ref{module:natural_prior} the log-partition, posterior mean and variance are given by:
\begin{align}
    &A_f[a_{x\to f}, b_{x\to f}] =
    A[b] - A[b_0] - \frac{a_{x\to f}}{2} \quad \text{with} \quad
    b = b_0 +  b_{x\to f}, \\
    &r_x^{f} = r[b], \qquad v_x^{f} = v[b],
\end{align}
where $A[b]$, $r[b]$ and $v[b]$ denote the log-partition, mean
and variance of a binary variable, see Section~\ref{module:binary_variable}.

\subsubsection{Sparse prior}

\begin{figure}[H]
  \centering
  \begin{tikzpicture}
    \node[latent, label=$\mathbb{R}$] (x) {$x$} ;
    \factor[left=1 of x]  {f} {$\mathcal{N}_\rho$} {} {x} ;
  \end{tikzpicture}
\end{figure}
The factor $f(x)=p(x \mid a_0, b_0 \eta_0)= [1 - \rho_0] \delta_0(x) + \rho_0 \mathcal{N}(x \mid  r_0, v_0)$
is the
Gauss-Bernoulli prior with natural
parameters $a_0 =  \frac{1}{v_0}$, $b_0 = \frac{r_0}{v_0}$ and $\eta_0 = A[a_0, b_0] - \ln \frac{\rho_0}{1 -\rho_0}$ where $A[a,b] = \frac{b^2}{2a} + \frac{1}{2} \ln \frac{2\pi}{a}$ is the log-partition of a real variable. 
The Gauss-Bernoulli prior corresponds to the natural prior 
for the sparse variable $z \in \mathbb{R}\cup\{0\}$.
According to Section~\ref{module:natural_prior} the log-partition, posterior mean and variance are given by:
\begin{align}
    &A_f[a_{x\to f}, b_{x\to f}] =
    A[a,b,\eta_0] - A[a_0, b_0, \eta_0] \quad \text{with} \quad
    a = a_0 +  a_{x\to f} , \quad 
    b = b_0 +  b_{x\to f}, \\
    &r_x^{f} = r[a,b,\eta_0], \qquad v_x^{f} = v[a,b,\eta_0],
\end{align}
where $A[a,b,\eta]$, $r[a,b,\eta]$ and $v[a,b,\eta]$ denote the log-partition, mean
and variance of a sparse variable, see Section~\ref{module:sparse_variable}. 
Also the sparsity of $x$ is equal to $\rho_x^{f} = \rho[a,b,\eta_0]$.

\subsubsection{Interval prior}

Let  $X \subset \mathbb{R}$ denotes any real interval, for example $X = \mathbb{R}_+$
for a positive prior.
\begin{figure}[H]
  \centering
  \begin{tikzpicture}
    \node[latent, label=$\mathbb{R}$] (x) {$x$} ;
    \factor[left=1 of x]  {f} {$p_X$} {} {x} ;
  \end{tikzpicture}
\end{figure}
The factor
$f(x)=p_X(x \mid a_0, b_0)= \frac{1}{p_X[a_0, b_0]} \mathcal{N}(x \mid  r_0, v_0) \delta_X(x)$
is the truncated Normal prior with natural
parameters $a_0 = \frac{1}{v_0}$ and $b_0 = \frac{r_0}{v_0}$. The truncated Normal corresponds to the natural prior for the interval variable $z \in X$.
According to Section~\ref{module:natural_prior} the log-partition, posterior mean and variance are given by:
\begin{align}
    &A_f[a_{x\to f}, b_{x\to f}] =
    A_X[a,b] - A_X[a_0, b_0] \quad \text{with} \quad
    a = a_0 +  a_{x\to f} , \quad 
    b = b_0 +  b_{x\to f}, \\
    &r_x^{f} = r_X[a,b], \qquad v_x^{f} = v_X[a,b]
\end{align}
where $A_X[a,b]$, $r_X[a,b]$ and $v_X[a,b]$ denote the log-partition, mean
and variance of a interval $X$ variable, see Section~\ref{module:interval_variable}.

\subsection{Separable likelihoods}

\subsubsection{Generic separable likelihood} 

\begin{figure}[H]
  \centering
  \begin{tikzpicture}
    \node[latent, label=$\mathbb{R}^N$] (z) {$z$} ;
    \node[obs, right=2 of z, label=$Y^N$] (y) {$y$} ;
    \factor[left=1 of y] {f} {$p_\text{out}$} {z} {y} ;
  \end{tikzpicture}
\end{figure}

Let $f(z) = p_\text{out}(y \mid z) = \prod_{n=1}^N p_\text{out}(y^{(n)} \mid  z^{(n)})$ be a separable likelihood over $z \in \mathbb{R}^N$ with observed $y \in Y^N$. The log-partition, posterior means and variances  are given by:
\begin{align}
    \label{separable_Af_likelihood}
    &A_f[a_{z \to f}, b_{z \to f}; y] = \sum_{n=1}^{N} A_f[a_{z \to f}, b_{z \to f}^{(n)}; y^{(n)}], \\
    &r_z^f[a_{z \to f}, b_{z \to f}; y]^{(n)} = r_z^{f}[a_{z \to f}, b_{z \to f}^{(n)}; y^{(n)}], \\
    \label{separable_vf_likelihood}
    &v_z^f[a_{z \to f}, b_{z \to f}; y] = \frac{1}{N} \sum_{n=1}^{N}
    v_z^f[a_{z \to f}, b_{z \to f}^{(n)}; y^{(n)}].
\end{align}
where on the RHS the same quantities are defined over scalar
$b_{z \to f}^{(n)}$ and $y^{(n)}$  in $\mathbb{R}$:
\begin{align}
    \label{scalar_Af_likelihood}
    &A_f[a_{z \to f}, b_{z \to f}; y] = \ln \int_\mathbb{R} dz \, p_\text{out}(y \mid z) \, e^{-\frac{1}{2} a_{z \to f} z^2 + b_{z \to f} z}, \\ 
    &r_z^f[a_{z \to f}, b_{z \to f}; y] = \partial_{b_{z \to f}} 
    A_f[a_{z \to f}, b_{z \to f}; y],  \\
    \label{scalar_vf_likelihood}
    &v_z^f[a_{z \to f}, b_{z \to f}; y] = \partial^2_{b_{z \to f}} 
    A_f[a_{z \to f}, b_{z \to f}; y].
\end{align}
In the remainder we will only derive the scalar case, as
it can be straightforwardly extended to the
high dimensional counterpart through Eqs~\eqref{separable_Af_likelihood}-\eqref{separable_vf_likelihood}.

\subsubsection{Teacher prior second moment}
The approximate teacher marginal (Proposition~\ref{prop:teacher_second_moments}) is:
\begin{equation}
    \label{teacher_second_moments_likelihood}
    p_f^{(0)}(y, z^{(0)} \mid \hat{\tau}_{z \to f}^{(0)}) = 
    p_\text{out}^{(0)}(y \mid z^{(0)}) \,
    e^{-\frac{1}{2} \hat{\tau}_{z \to f}^{(0)} z^{(0)2} - A_z^{(0)}[\hat{\tau}_{z \to f}^{(0)}]}, 
\end{equation}
with log-partition:
\begin{equation}
    \label{teacher_second_moments_likelihood_A}
    A_f^{(0)}[\hat{\tau}_{z \to f}^{(0)}] = \ln \int_\mathbb{R} dy dz \, p_\text{out}^{(0)}(y \mid z) \, 
    e^{-\frac{1}{2} \hat{\tau}_{z \to f}^{(0)} z^2} = \frac{1}{2} \ln \frac{2\pi}{\hat{\tau}_{z \to f}^{(0)}}
\end{equation}
which yields the dual mapping:
\begin{equation}
    \tau_z^{(0)} = -2 \partial_{\hat{\tau}_{z \to f}^{(0)}} A_f^{(0)}[\hat{\tau}_{z \to f}^{(0)}] = \frac{1}{\hat{\tau}_{z \to f}^{(0)}}
\end{equation}
The approximate teacher marginal is therefore equal to:
\begin{equation}
     p_f^{(0)}(y, z^{(0)} \mid \hat{\tau}_{z \to f}^{(0)})  =  p_\text{out}^{(0)}(y \mid z^{(0)}) \, \mathcal{N}(z^{(0)} \mid 0, \tau_z^{(0)}).
\end{equation}

\subsubsection{Replica symmetric \label{module:likelihood_rs}}
The RS potential and overlaps are given by a low-dimensional integration of the corresponding scalar EP quantities Eqs~\eqref{scalar_Af_likelihood}-\eqref{scalar_vf_likelihood}:
\begin{align}
    &\bar{A}_f[\hat{m}_{z \to f},\hat{q}_{z \to f},\hat{\tau}_{z \to f}] = 
    \int_{\mathbb{R}^2} db_{z \to f} dy \, p_f^{(0)}(b_{z \to f}, y) \, A_f[a_{z \to f}, b_{z \to f};y], \\
    &m_z^f[\hat{m}_{z \to f},\hat{q}_{z \to f},\hat{\tau}_{z \to f}] = 
    \int_{\mathbb{R}^3} db_{z \to f} dy dz^{(0)}  \, p_f^{(0)}(b_{z \to f}, y,  z^{(0)}) \, z^{(0)} r_z^f[a_{z \to f}, b_{z \to f};y] \\
    &q_z^f[\hat{m}_{z \to f},\hat{q}_{z \to f},\hat{\tau}_{z \to f}] = 
    \int_{\mathbb{R}^2} db_{z \to f} dy \, p_f^{(0)}(b_{z \to f}, y) \, r_z^f[a_{z \to f}, b_{z \to f};y]^2, \\
    &v_z^f[\hat{m}_{z \to f},\hat{q}_{z \to f},\hat{\tau}_{z \to f}] = 
    \int_{\mathbb{R}^2} db_{z \to f} dy  \, p_f^{(0)}(b_{z \to f}, y) \, v_z^f[a_{z \to f}, b_{z \to f};y], \\
    &\tau_z^f[\hat{m}_{z \to f},\hat{q}_{z \to f},\hat{\tau}_{z \to f}] 
    = q_z^f[\hat{m}_{z \to f},\hat{q}_{z \to f},\hat{\tau}_{z \to f}] 
    + v_z^f[\hat{m}_{z \to f},\hat{q}_{z \to f},\hat{\tau}_{z \to f}], 
\end{align}
where the ensemble average Eq.~\eqref{RS_ensemble_avg_f} is now over scalar $b_{z \to f}$, $y$ and $z^{(0)}$ in $\mathbb{R}$:
\begin{equation}
    p_f^{(0)}(b_{z \to f}, y, z^{(0)}) = \mathcal{N}(b_{z \to f} \mid \hat{m}_{z \to f} z^{(0)}, \hat{q}_{z \to f}) \, p_f^{(0)}(y, z^{(0)} \mid \hat{\tau}_{z \to f}^{(0)}) 
\end{equation}
with $a_{z \to f} = \hat{\tau}_{z \to f} + \hat{q}_{z \to f}$ and 
$p_f^{(0)}(y, z^{(0)} \mid \hat{\tau}_{z \to f}^{(0)})$ given by Eq.~\eqref{teacher_second_moments_likelihood}.
The dual mapping to the overlaps now simply reads:
\begin{equation}
    m_z^{f}
  = \partial_{\hat{m}_{z \to f}} \bar{A}_f, \quad 
  -\tfrac{1}{2} q_z^{f}
  = \partial_{\hat{q}_{z \to f}} \bar{A}_f, \quad
  -\tfrac{1}{2} \tau_z^{f}
  = \partial_{\hat{\tau}_{z \to f}} \bar{A}_f.
\end{equation}

\subsubsection{Bayes-optimal \label{module:likelihood_bo}}
The BO potential and overlap are given by a low-dimensional integration of the corresponding scalar EP quantities:
\begin{align}
    &\bar{A}_f^{(0)}[\hat{m}_{z \to f}] = 
    \int_{\mathbb{R}^2} db_{z \to f} dy  \, p_f^{(0)}(b_{z \to f}, y) \, A_f^{(0)}[a_{z \to f}, b_{z \to f};y], \\
    &v_z^f[\hat{m}_{z \to f}] = 
    \int_{\mathbb{R}^2} db_{z \to f} dy \, p_f^{(0)}(b_{z \to f}, y) \, v_z^f[a_{z \to f}, b_{z \to f};y], \\
    &m_z^f[\hat{m}_{z \to f}] = \tau_z^{(0)} - v_z^f[\hat{m}_{z \to f}], 
\end{align}
where the ensemble average Eq.~\eqref{BO_ensemble_avg_f} is now over scalar $b_{z \to f}$, $y$ and $z^{(0)}$ in $\mathbb{R}$:
\begin{equation}
    p_f^{(0)}(b_{z \to f}, y, z^{(0)}) = \mathcal{N}(b_{z \to f} \mid \hat{m}_{z \to f} z^{(0)}, \hat{m}_{z \to f}) \, p_f^{(0)}(y, z^{(0)} \mid \hat{\tau}_{z \to f}^{(0)}) 
\end{equation}
with $a_{z \to f} =  \hat{\tau}_{z \to f}^{(0)}+ \hat{m}_{z \to f}$ and 
$p_f^{(0)}(z^{(0)} \mid \hat{\tau}_{z \to f}^{(0)})$ given by Eq.~\eqref{teacher_second_moments_likelihood}.
The relationship Eq.~\eqref{BO_potential_decomposition}
between the mutual information and the BO potential now reads:
\begin{equation}
    I_f[\hat{m}_{z \to f}] + E_f = \tfrac{1}{2} \hat{m}_{z \to f}\tau_z^{(0)} - \bar{A}_f^{(0)}[\hat{m}_{z \to f}] + A_f^{(0)}[\hat{\tau}_{z \to f}^{(0)}].
\end{equation}
The dual mapping to the variance and overlap simply reads:
\begin{equation}
  \tfrac{1}{2} v_z^{f}
  = \partial_{\hat{m}_{z \to f}} I_f, \qquad
   \tfrac{1}{2} m_z^{f}
  = \partial_{\hat{m}_{z \to f}} \bar{A}_f^{(0)}.
\end{equation}

\subsubsection{Gaussian likelihood \label{module:gaussian_likelihood}}

\begin{figure}[H]
  \centering
  \begin{tikzpicture}
    \node[latent, label=$\mathbb{R}$] (z) {$z$} ;
    \node[obs, right=2 of z, label=$\mathbb{R}$] (y) {$y$} ;
    \factor[left=1 of y] {f} {$\Delta$} {z} {y} ;
  \end{tikzpicture}
\end{figure}
The factor $f(z)=p_\text{out}(y \mid z)=\mathcal{N}(y\mid z, \Delta)$ is the Gaussian likelihood
with noise variance $\Delta$ and observed $y$.
The log-partition is given by:
\begin{align}
    &A_f[a_{z\to f}, b_{z\to f}; y] = A[a, b] - A[a_y, b_y], \\
    &a_y = \frac{1}{\Delta}, \quad
    b_y=\frac{y}{\Delta}, \quad
    a = a_{z\to f} + a_y, \quad
    b = b_{z\to f} + b_y,
\end{align}
where $A[a,b]$ denotes the
log-partition of a real variable, see Section~\ref{module:real_variable}.
The posterior mean and variance are given by:
\begin{equation}
r_z^{f} = \frac{b}{a}, \qquad v_z^{f} = \frac{1}{a},
\end{equation}
leading to the constant $f \to z$
update:
\begin{equation}
  a_{f\to z}^\text{new} = a_y,
  \qquad
  b_{f\to z}^\text{new} = b_y.
\end{equation}
The ensemble average variance is directly given by $v_z^f = \frac{1}{a}$.

\subsubsection{Sgn likelihood}

\begin{figure}[H]
  \centering
  \begin{tikzpicture}
    \node[latent, label=$\mathbb{R}$] (z) {$z$} ;
    \node[obs, right=2 of z, label=$\pm$] (y) {$y$} ;
    \factor[left=1 of y] {f} {sgn} {z} {y} ;
  \end{tikzpicture}
\end{figure}
The factor $f(z)=p_\text{out}(y \mid z)=\delta(y-\text{sgn}(z))$ is the deterministic likelihood
$y = \text{sgn}(z) \in \pm$.
The log-partition, posterior mean and variance are given by:
\begin{align}
    &A_f[a_{z\to f}, b_{z\to f}; y] =
    A_y[a_{z\to f}, b_{z\to f}], \\
    &r_z^{f} = r_y[a_{z\to f}, b_{z\to f}], \qquad
    v_z^{f} = v_y[a_{z\to f}, b_{z\to f}],
\end{align}
where $A_\pm[a,b]$, $r_\pm[a,b]$ and $v_\pm[a,b]$ denote the log-partition, mean
and variance of a positive/negative variable, see Section~\ref{module:positive_variable}.

\subsubsection{Abs likelihood}

\begin{figure}[H]
  \centering
  \begin{tikzpicture}
    \node[latent, label=$\mathbb{R}$] (z) {$z$} ;
    \node[obs, right=2 of z, label=$\mathbb{R}_+$] (y) {$y$} ;
    \factor[left=1 of y] {f} {abs} {z} {y} ;
  \end{tikzpicture}
\end{figure}
The factor $f(z)=p_\text{out}(y \mid z)=\delta(y-\text{abs}(z))$ is the deterministic likelihood
$y = \text{abs}(z) \in \mathbb{R}_+$.
The log-partition, posterior mean and variance are given by:
\begin{align}
    &A_f[a_{z\to f}, b_{z\to f}; y] =
    -\frac{a_{z\to f}y^2}{2} + A[b]
    \quad \text{with} \quad
    b = y b_{z\to f}, \\
    &r_z^{f} = y \, r[b], \qquad
    v_z^{f} = y^2 v[b],
\end{align}
where $A[b]$, $r[b]$ and $v[b]$ denote the log-partition, mean
and variance of a binary variable, see Section~\ref{module:binary_variable}.

\subsubsection{Phase likelihood}

\begin{figure}[H]
  \centering
  \begin{tikzpicture}
    \node[latent, label=$\mathbb{C}$] (z) {$z$} ;
    \node[obs, right=2 of z, label=$\mathbb{S}^1$] (y) {$y$} ;
    \factor[left=1 of y] {f} {$e^{i\theta(.)}$} {z} {y} ;
  \end{tikzpicture}
\end{figure}
The factor $f(z)=p_\text{out}(y \mid z)=\delta(y-e^{i\theta(z)})$ is the deterministic likelihood
$y = e^{i\theta(z)} \in \mathbb{S}^1$.
The log-partition, posterior mean and variance are given by:
\begin{align}
    &A_f[a_{z\to f}, b_{z\to f}; y] =
    A_+[a_{z\to f}, b] \quad \text{with} \quad
    b = y^\intercal  b_{z\to f}, \\
    &r_z^{f} = y \, r_+[a_{z\to f}, b], \qquad
    v_z^{f} = \tfrac{1}{2} v_+[a_{z\to f}, b],
\end{align}
where $A_+[a,b]$, $r_+[a,b]$ and $v_+[a,b]$ denote the log-partition, mean
and variance of a positive variable, see Section~\ref{module:positive_variable}. 
The $\frac{1}{2}$ factor in the variance comes from the average over the real 
and imaginary parts.

\subsubsection{Modulus likelihood}

\begin{figure}[H]
  \centering
  \begin{tikzpicture}
    \node[latent, label=$\mathbb{C}$] (z) {$z$} ;
    \node[obs, right=2 of z, label=$\mathbb{R}_+$] (y) {$y$} ;
    \factor[left=1 of y] {f} {$| \cdot| $} {z} {y} ;
  \end{tikzpicture}
\end{figure}
The factor $f(z)=p_\text{out}(y \mid z)=\delta(y-|z| )$ is the deterministic likelihood
$y = |z|  \in \mathbb{R}_+$.
The log-partition, posterior mean and variance are given by:
\begin{align}
    &A_f[a_{z\to f}, b_{z\to f}; y] =
    -\frac{a_{z\to f}y^2}{2} + \ln y + A_{\mathbb{S}^1}[b]
    \quad \text{with} \quad
    b = y b_{z\to f}, \\
    &r_z^{f} =  y \, r_{\mathbb{S}^1}[b], \qquad
    v_z^{f} = y^2 v_{\mathbb{S}^1}[b],
\end{align}
where $A_{\mathbb{S}^1}[b]$, $r_{\mathbb{S}^1}[b]$ and $v_{\mathbb{S}^1}[b]$ denote the log-partition,
mean and variance of a phase variable, see Section~\ref{module:phase_variable}.

\subsection{Separable channels}

\subsubsection{Generic separable channel}

\begin{figure}[H]
  \centering
  \begin{tikzpicture}
    \node[latent, label=$\mathbb{R}^N$] (z) {$z$} ;
    \node[latent, right=2 of z, label=$\mathbb{R}^N$] (x) {$x$} ;
    \factor[left=1 of x] {not} {$p$} {z} {x} ;
  \end{tikzpicture}
\end{figure}

Let $f(x, z) = p(x \mid  z) = \prod_{n=1}^N p(x^{(n)} \mid  z^{(n)})$ be a separable channel with input $z \in \mathbb{R}^N$ and output $x \in \mathbb{R}^N$. The log-partition, posterior means and variances  are given by:
\begin{align}
    \label{separable_Af_channel}
    &A_f[a_f, b_f] = \sum_{n=1}^{N} 
    A_f[a_f, b_f^{(n)}], \\
    &r_x^{f(n)}[a_f, b_f] = r_x^{f}[a_f, b_f^{(n)}], \quad
    v_x^{f}[a_f, b_f] = \frac{1}{N} \sum_{n=1}^{N} v_x^{f}[a_f, b_f^{(n)}], \\
    \label{separable_vf_channel}
    &r_z^{f(n)}[a_f, b_f] = r_z^{f}[a_f, b_f^{(n)}], \quad
    v_z^{f}[a_f, b_f] = \frac{1}{N} \sum_{n=1}^{N} v_z^{f}[a_f, b_f^{(n)}].
\end{align}
where on the RHS the same quantities are defined over scalar
$b_{x \to f}^{(n)}$ and $b_{z \to f}^{(n)}$ in $\mathbb{R}$:
\begin{align}
    \label{scalar_Af_channel}
    &A_f[a_f, b_f] = \ln \int_{\mathbb{R}^2} dxdz \, p(x \mid z) \, e^{
    -\frac{1}{2} a_{x \to f} x^2 + b_{x \to f} x
    -\frac{1}{2} a_{z \to f} z^2 + b_{z \to f} z
    }, \\ 
    &r_x^f[a_f, b_f]= \partial_{b_{x \to f}} 
    A_f[a_{x \to f}, b_{x \to f}], \quad 
    v_x^f[a_f, b_f]= \partial^2_{b_{x \to f}} 
    A_f[a_f, b_f], \\
     \label{scalar_vf_channel}
     &r_z^f[a_f, b_f]= \partial_{b_{z \to f}} 
    A_f[a_{x \to f}, b_{x \to f}], \quad 
    v_z^f[a_f, b_f]= \partial^2_{b_{z \to f}} 
    A_f[a_f, b_f].
\end{align}
In the remainder we will only derive the scalar case, as
it can be straightforwardly extended to the
high dimensional counterpart through Eqs~\eqref{separable_Af_channel}-\eqref{separable_vf_channel}.

\subsubsection{Teacher prior second moment}
The approximate teacher marginal (Proposition~\ref{prop:teacher_second_moments}) is:
\begin{equation}
    \label{teacher_second_moments_channel}
    p_f^{(0)}(x^{(0)}, z^{(0)} \mid \hat{\tau}_f^{(0)}) = 
    p^{(0)}(x^{(0)} \mid z^{(0)}) \,
    e^{
    -\frac{1}{2} \hat{\tau}_{x \to f}^{(0)} x^{(0)2} 
    -\frac{1}{2} \hat{\tau}_{z \to f}^{(0)} z^{(0)2} 
    - A_f^{(0)}[\hat{\tau}_f^{(0)}]
    }, 
\end{equation}
with log-partition:
\begin{equation}
    \label{teacher_second_moments_channel_A}
    A_f^{(0)}[\hat{\tau}_f^{(0)}] = \ln \int_{\mathbb{R}^2} dxdz \, p^{(0)}(x \mid z) \, 
    e^{
    -\frac{1}{2} \hat{\tau}_{x \to f}^{(0)} x^2
    -\frac{1}{2} \hat{\tau}_{z \to f}^{(0)} z^2
    }
\end{equation}
which yields the dual mapping:
\begin{equation}
    \tau_x^{(0)} = -2 \partial_{\hat{\tau}_{x \to f}^{(0)}} A_f^{(0)}[\hat{\tau}_f^{(0)}], \quad
    \tau_z^{(0)} = -2 \partial_{\hat{\tau}_{z \to f}^{(0)}} A_f^{(0)}[\hat{\tau}_f^{(0)}],
\end{equation}
When the teacher factor graph is a Bayesian network (Section~\ref{sec:teacher_second_moments}) we have
\begin{equation}
    \hat{\tau}^{(0)}_{x \to f} = 0,  \quad
    \hat{\tau}^{(0)}_{z \to f} = \frac{1}{\tau_z^{(0)}}, \quad
     p_f^{(0)}(x^{(0)}, z^{(0)} \mid \hat{\tau}_f^{(0)}) = p^{(0)}(x^{(0)} \mid z^{(0)} ) \,
     \mathcal{N}(z^{(0)} \mid 0, \tau_z^{(0)}).
\end{equation}

\subsubsection{Replica symmetric \label{module:channel_rs}}
The RS potential and overlaps are given by a low-dimensional integration of the corresponding scalar EP quantities Eqs~\eqref{scalar_Af_channel}-\eqref{scalar_vf_channel}:
\begin{align}
    &\bar{A}_f[\hat{m}_f,\hat{q}_f,\hat{\tau}_f] = 
    \int_{\mathbb{R}^2} db_{x \to f} db_{z \to f}  \, p_f^{(0)}(b_{x \to f}, b_{z \to f}) \, A_f[a_f, b_f], \\
    &m_x^f[\hat{m}_f,\hat{q}_f,\hat{\tau}_f] = 
    \int_{\mathbb{R}^3} db_{x \to f} db_{z \to f} dx^{(0)} \, p_f^{(0)}(b_{x \to f}, b_{z \to f}, x^{(0)}) \, x^{(0)} r_x^f[a_f, b_f] , \\
    &q_x^f[\hat{m}_f,\hat{q}_f,\hat{\tau}_f] = 
    \int_{\mathbb{R}^2} db_{x \to f} db_{z \to f}  \, p_f^{(0)}(b_{x \to f}, b_{z \to f}) \, r_x^f[a_f, b_f]^2, \\
    &v_x^f[\hat{m}_f,\hat{q}_f,\hat{\tau}_f] = 
    \int_{\mathbb{R}^2} db_{x \to f} db_{z \to f} \, p_f^{(0)}(b_{x \to f}, b_{z \to f}) \, v_x^f[a_f, b_f], \\
    &\tau_x^f[\hat{m}_f,\hat{q}_f,\hat{\tau}_f] 
    = q_x^f[\hat{m}_f,\hat{q}_f,\hat{\tau}_f] 
    + v_x^f[\hat{m}_f,\hat{q}_f,\hat{\tau}_f], \quad (\text{idem } z)
\end{align}
where the ensemble average Eq.~\eqref{RS_ensemble_avg_f} is now over scalar $b_{x \to f}$, $b_{z \to f}$, $x^{(0)}$ and $z^{(0)}$ in $\mathbb{R}$:
\begin{align}
    \notag
    &p_f^{(0)}(b_{x \to f}, b_{z \to f}, x^{(0)}, z^{(0)}) = \\
    &\mathcal{N}(b_{x \to f} \mid \hat{m}_{x \to f} x^{(0)}, \hat{q}_{x \to f}) \, 
    \mathcal{N}(b_{z \to f} \mid \hat{m}_{z \to f} x^{(0)}, \hat{q}_{z \to f}) \,
    p_f^{(0)}(x^{(0)}, z^{(0)} \mid \hat{\tau}_f^{(0)}) 
\end{align}
with $a_{x \to f} = \hat{\tau}_{x \to f} + \hat{q}_{x \to f}$, 
$a_{z \to f} = \hat{\tau}_{z \to f} + \hat{q}_{z \to f}$ and
$p_f^{(0)}(x^{(0)}, z^{(0)} \mid \hat{\tau}_f^{(0)})$ given by Eq.~\eqref{teacher_second_moments_channel}.
The dual mapping to the overlaps now simply reads:
\begin{align}
   & m_x^{f}
  = \partial_{\hat{m}_{x \to f}} \bar{A}_f, \quad 
  -\tfrac{1}{2} q_x^{f}
  = \partial_{\hat{q}_{x \to f}} \bar{A}_f, \quad
  -\tfrac{1}{2} \tau_x^{f}
  = \partial_{\hat{\tau}_{x \to f}} \bar{A}_f, \\
   & m_z^{f}
  = \partial_{\hat{m}_{z \to f}} \bar{A}_f, \quad 
  -\tfrac{1}{2} q_z^{f}
  = \partial_{\hat{q}_{z \to f}} \bar{A}_f, \quad
  -\tfrac{1}{2} \tau_z^{f}
  = \partial_{\hat{\tau}_{z \to f}} \bar{A}_f.
\end{align}

\subsubsection{Bayes-optimal \label{module:channel_bo}}
The BO potential and overlap are given by a low-dimensional integration of the corresponding scalar EP quantities:
\begin{align}
    &\bar{A}_f^{(0)}[\hat{m}_f] = 
    \int_{\mathbb{R}^2} db_{x \to f} db_{z \to f}  \, p_f^{(0)}(b_{x \to f}, b_{z \to f}) \, A_f^{(0)}[a_f, b_f], \\
    &v_x^f[\hat{m}_{x \to f}] = 
     \int_{\mathbb{R}^2} db_{x \to f} db_{z \to f}  \, p_f^{(0)}(b_{x \to f}, b_{z \to f}) \, v_x^f[a_f, b_f], \\
    &m_x^f[\hat{m}_{x \to f}] = \tau_x^{(0)} - v_x^f[\hat{m}_{x \to f}], \quad (\text{idem } z)
\end{align}
where the ensemble average Eq.~\eqref{BO_ensemble_avg_f} is now over scalar $b_{x \to f}$, $b_{z \to f}$, $x^{(0)}$ and $z^{(0)}$ in $\mathbb{R}$:
\begin{align}
    &p_f^{(0)}(b_{x \to f}, b_{z \to f}, x^{(0)}, z^{(0)}) = \\ 
    &\mathcal{N}(b_{x \to f} \mid \hat{m}_{x \to f} x^{(0)}, \hat{m}_{x \to f}) \, 
    \mathcal{N}(b_{z \to f} \mid \hat{m}_{z \to f} x^{(0)}, \hat{m}_{z \to f}) \,
    p_f^{(0)}(x^{(0)}, z^{(0)} \mid \hat{\tau}_f^{(0)}) 
\end{align}
with $a_{x \to f} = \hat{\tau}_{x \to f}^{(0)} + \hat{m}_{x \to f}$, 
$a_{z \to f} = \hat{\tau}_{z \to f}^{(0)} + \hat{m}_{z \to f}$ and
$p_f^{(0)}(x^{(0)}, z^{(0)} \mid \hat{\tau}_f^{(0)})$ given by Eq.~\eqref{teacher_second_moments_channel}.
The relationship Eq.~\eqref{BO_potential_decomposition}
between the mutual information and the BO potential now reads:
\begin{equation}
    I_f[\hat{m}_f] = 
    \tfrac{1}{2} \hat{m}_{x \to f}\tau_x^{(0)} +  \tfrac{1}{2} \hat{m}_{z \to f}\tau_z^{(0)} 
    - \bar{A}_f^{(0)}[\hat{m}_f] + A_f^{(0)}[\hat{\tau}_f^{(0)}].
\end{equation}
The dual mapping to the variances and overlaps simply reads:
\begin{align}
  &\tfrac{1}{2} v_x^{f}
  = \partial_{\hat{m}_{x \to f}} I_f, \qquad
   \tfrac{1}{2} m_x^{f}
  = \partial_{\hat{m}_{x \to f}} \bar{A}_f^{(0)}, \\
  &\tfrac{1}{2} v_z^{f}
  = \partial_{\hat{m}_{z \to f}} I_f, \qquad
   \tfrac{1}{2} m_z^{f}
  = \partial_{\hat{m}_{z \to f}} \bar{A}_f^{(0)}.
\end{align}

\subsubsection{Gaussian noise channel \label{module:gaussian_channel}}

\begin{figure}[H]
  \centering
  \begin{tikzpicture}
    \node[latent, label=$\mathbb{R}$] (z) {$z$} ;
    \node[latent, right=2 of z, label=$\mathbb{R}$] (x) {$x$} ;
    \factor[left=1 of x] {not} {$\Delta$} {z} {x} ;
  \end{tikzpicture}
\end{figure}

The factor $f(x,z)=p(x\mid z)=\mathcal{N}(x\mid z, \Delta)$ is the additive 
Gaussian noise channel $x = z + \sqrt{\Delta} \xi$
with variance $\Delta$ and precision $a_\Delta = \Delta^{-1}$.
The log-partition is given by:
\begin{align}
    &A_f[a_f, b_f]
    = \frac{1}{2} b^\intercal \Sigma b + \frac{1}{2} \ln \det 2\pi\Sigma
    - \frac{1}{2}\ln 2\pi \Delta ,
    \\
    &b = \begin{bmatrix}
          b_{z\to f} \\ b_{x\to f}
        \end{bmatrix},
    \quad
    A = \begin{bmatrix}
          a_\Delta + a_{z\to f} & -a_\Delta  \\
          -a_\Delta& a_\Delta + a_{x\to f}
        \end{bmatrix},
    \quad
    \Sigma = A^{-1}.
\end{align}
The posterior mean and variance are given by:
\begin{align}
\label{noise_rz_vz}
&v_z^f = \frac{a_\Delta + a_{x \to \Delta}}{a_\Delta a},
\qquad
r_z^{f} = v_z^{f} \left[
b_{z \to f} +
\frac{a_\Delta b_{x \to f}}{a_\Delta + a_{x \to f}}
\right], \\
\label{noise_rx_vx}
&v_x^f = \frac{a_\Delta + a_{z \to \Delta}}{a_\Delta a},
\qquad
r_x^{f} = v_x^{f} \left[
b_{x \to f} +
\frac{a_\Delta b_{z \to f}}{a_\Delta + a_{z \to f}}
\right], \\
\text{with} \quad &a = a_{x \to f} + a_{z \to f}
  + \frac{a_{x \to f}a_{z \to f}}{a_\Delta}.
\end{align}
We therefore have:
\begin{align}
&a_z^{f} = a_{z \to f}
+ \frac{a_\Delta a_{x \to f}}{a_\Delta + a_{x \to f}},
\qquad
b_z^{f} = \left[
b_{z \to f} +
\frac{a_\Delta b_{x \to f}}{a_\Delta + a_{x \to f}}
\right], \\
&a_x^f = a_{x \to f}
+ \frac{a_\Delta a_{z \to f}}{a_\Delta + a_{z \to f}},
\qquad
b_x^{f} =  \left[
b_{x \to f} +
\frac{a_\Delta b_{z \to f}}{a_\Delta + a_{z \to f}}
\right].
\end{align}
leading to the backward $f \to z$ and forward $f \to x$
updates:
\begin{align}
  &a_{f\to z}^\text{new} =
  \frac{a_\Delta}{a_\Delta + a_{x\to f}} a_{x\to f}, \qquad
  b_{f\to z}^\text{new} =
  \frac{a_\Delta}{a_\Delta + a_{x\to f}} b_{x\to f}, \\
  &a_{f\to x}^\text{new} =
  \frac{a_\Delta}{a_\Delta + a_{z\to f}} a_{z\to f}, \qquad
  b_{f\to x}^\text{new} =
  \frac{a_\Delta}{a_\Delta + a_{z\to f}} b_{z\to f}.
\end{align}
The ensemble average variances are still given by
Eqs~\eqref{noise_rz_vz}-\eqref{noise_rx_vx}. 

\subsubsection{Piecewise linear activation}

\begin{figure}[H]
  \centering
  \begin{tikzpicture}
    \node[latent, label=$\mathbb{R}$] (z) {$z$} ;
    \node[latent, right=2 of z, label=$\mathbb{R}$] (x) {$x$} ;
    \factor[left=1 of x] {f} {$\sigma$} {z} {x} ;
  \end{tikzpicture}
\end{figure}

Factor $f(x,z)=p(x \mid z)=\delta(x-\sigma(z))$ is the deterministic channel
$x = \sigma(z)$, where we assume the activation to be piecewise linear\footnote{
  ReLU, leaky ReLU, hard tanh, hard sigmoid, sgn and abs are popular examples.
}
\begin{equation}
  \sigma(z) = \sum_{i \in I} \mathbf{1}_{R_i}(z) [x_i + \gamma_i z], \quad \mathbb{R} = \bigsqcup_{i \in I} R_i .
\end{equation}
As the real line $\mathbb{R}$ is the disjoint union of the regions $R_i$ the 
factor partition function is simply the sum of the region partition functions:
\begin{align}
    &Z_f[a_f, b_f] = \sum_{i \in I} Z_f^i[a_f, b_f], \\
    &Z_f^i[a_f, b_f] = \int_{R_i} dz\, e^{
    -\frac{1}{2} a_{x\to f} [x_i + \gamma_i z] ^2 
    + b_{x \to f} [x_i + \gamma_i z]
    -\frac{1}{2} a_{z\to f} z^2 
    + b_{z \to f} z
    }.
\end{align}
Thus at the factor level, the log-partition, posterior means and variances are given by:
\begin{align}
  &A_f[a_f, b_f] = \ln \sum_{i \in I} e^{A_f^i[a_f, b_f]}, \\
  &r_z^{f} = \sum_{i \in I} p_i \, r_z^i, \quad
  v_z^{f} = \sum_{i \in I} p_i \, v_z^i + 
  \sum_{i<j \in I} p_i \, p_j [ r_z^i - r_z^j]^2 , \\
  &r_x^{f} = \sum_{i \in I} p_i \, r_x^i, \quad
  v_x^{f} = \sum_{i \in I} p_i \, v_x^i
  + \sum_{i<j \in I} p_i \, p_j [
  r_x^i - r_x^j
  ]^2 .
\end{align}
where the region probabilities are given by
\begin{equation}
 p_i = \frac{\exp(A_f^i)}{ \sum_{j \in I} \exp(A_f^j) } \quad ie \quad
 p = \mathrm{softmax}\,[(A_f^i)_{i \in I}].
\end{equation}
The corresponding quantities at the linear region level are given by:
\begin{align}
  \notag
  &A_f^i[a_f, b_f] =
  -\frac{a_{x \to f}x_i^2}{2} + b_{x \to f}x_i + A_{R_i}[a_i,b_i] \\
   &\text{with} \quad 
   a_i = a_{z \to f} + \gamma_i^2 a_{x \to f}, \quad
   b_i = b_{z \to f} + \gamma_i [ b_{x \to f}  - a_{x \to f} x_i ] , \\
   &r_z^i = r_{R_i}[a_i,b_i] , \qquad
   r_x^i = x_i + \gamma_i r_z^i, \\
   &v_z^i = v_{R_i}[a_i,b_i], \qquad
   v_x^i = \gamma_i^2 v_z^i, 
\end{align}
where $A_R[a,b]$, $r_R[a,b]$ and $v_R[a,b]$ denote the log-partition, mean
and variance of an interval $R$ variable, see Section~\ref{module:interval_variable}.

\newpage
\bibliographystyle{plainnat}
\bibliography{JMLRtramp}

\end{document}